\documentclass[letterpaper,journal]{IEEEtran}
\usepackage{lineno}
\modulolinenumbers[5]
\usepackage{xpatch}
\usepackage{threeparttable}
\usepackage{tabularx}
\usepackage{bm}
\usepackage{array}
\usepackage{graphicx}
\usepackage{amsmath,amssymb,amsthm}
\usepackage{graphicx}
\usepackage{caption}
\usepackage{multirow}
\usepackage{float}
\usepackage[table]{xcolor} 
\usepackage{enumitem} 
\usepackage{makecell}
\usepackage{subcaption}
\usepackage{algorithm}
\usepackage{algorithmic}

\usepackage{xcolor}
\usepackage{tabularx}    
\usepackage{ragged2e}  
\usepackage{textcomp}
\usepackage{stfloats}
\usepackage{url}
\usepackage{verbatim}
\usepackage{booktabs}
\usepackage{cite}
\usepackage{dsfont}
\usepackage{bm}
\usepackage{multirow}
\usepackage{amssymb}
\usepackage{siunitx}
\usepackage{hyperref}
\usepackage{listings}
\usepackage{multirow}
\usepackage{array}
\usepackage{graphicx}

\newcommand{\figAutoAlgo}{%
\begin{figure*}[htbp]
    \centering
    \includegraphics[width=\textwidth]{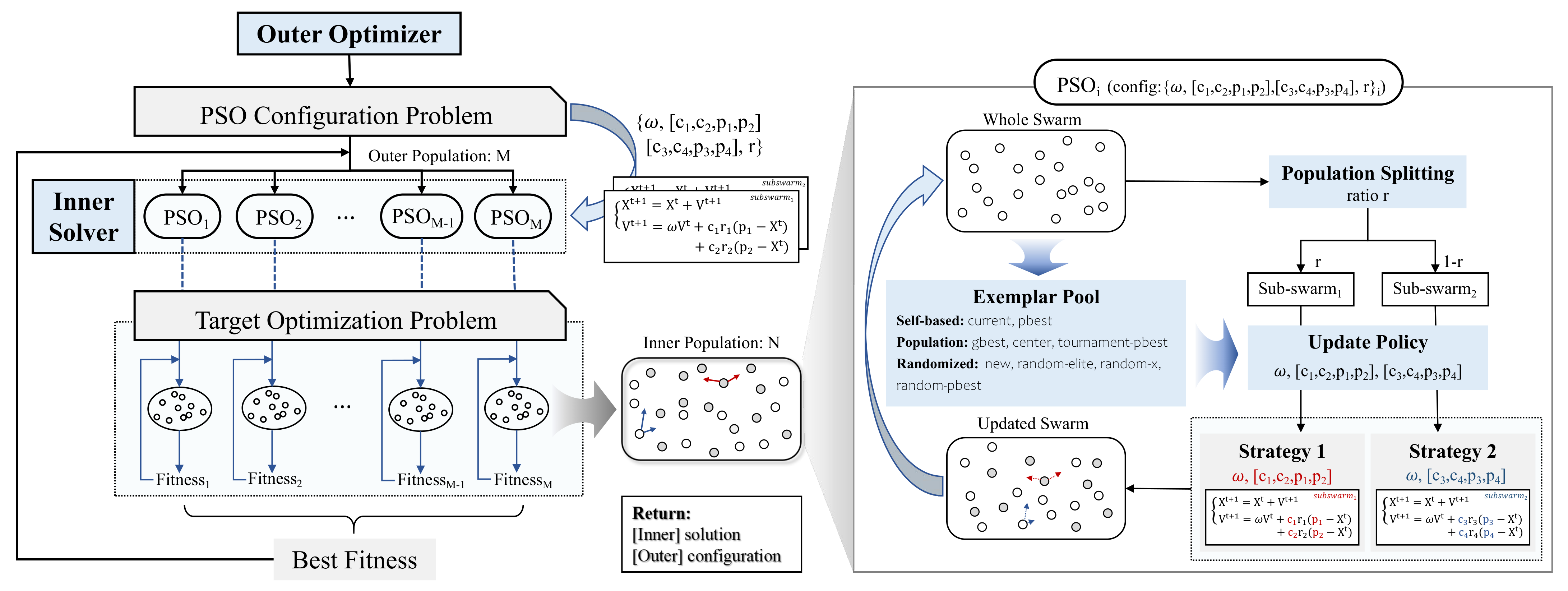}
    \caption{The meta-framework of proposed  AutoPSO. \textbf{Left:} the outer-level optimizer searches the configuration space of PSO variants, while the inner-level solver instantiates each candidate configuration as a generalized PSO and evaluates its optimization performance on the target problem. \textbf{Right:} the inner generalized PSO evolves the population by partitioning the swarm into two sub-swarms, assigning sub-swarm-specific update policies, selecting exemplars from a shared exemplar pool, and updating particle velocities and positions to produce the next-generation swarm.}
    \label{fig:autoAlgo}
\end{figure*}
}

\newcommand{\figCECfixTime}{%
\begin{figure}[htbp]
   
    \centering
    \includegraphics[width=0.22\textwidth]{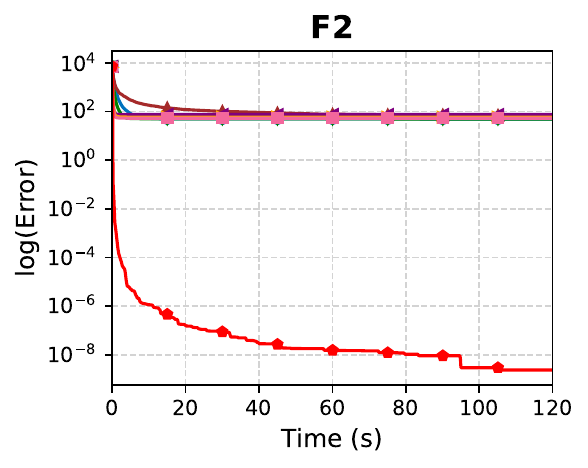}
    \hspace{-0.1cm}
    \includegraphics[width=0.22\textwidth]{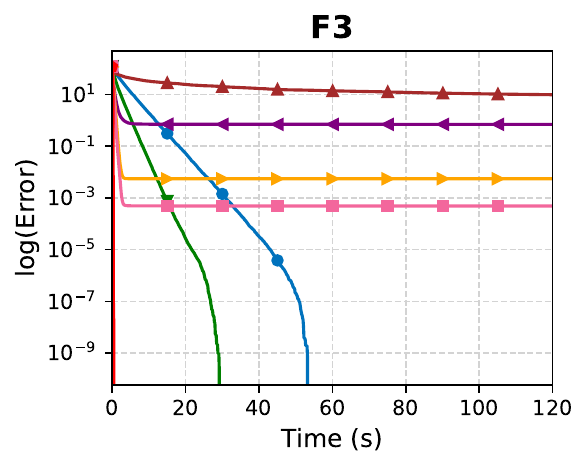}
    
    \vspace{0.05cm}
    
    \includegraphics[width=0.22\textwidth]{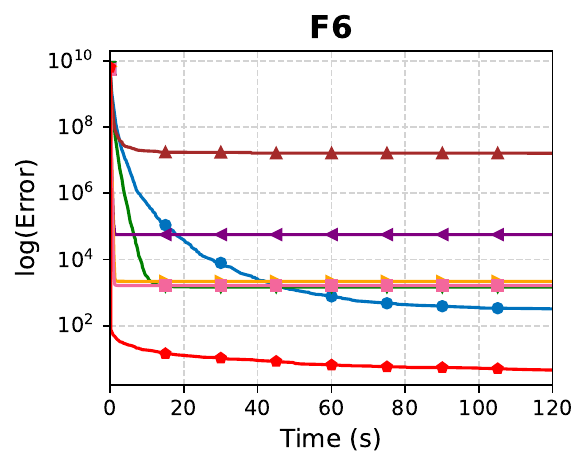}
    \hspace{-0.1cm}
    \includegraphics[width=0.22\textwidth]{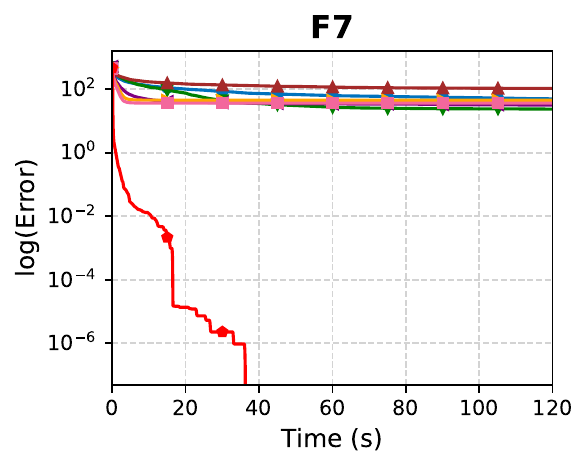}

    \vspace{-0.3cm}

    \includegraphics[width=0.22\textwidth]{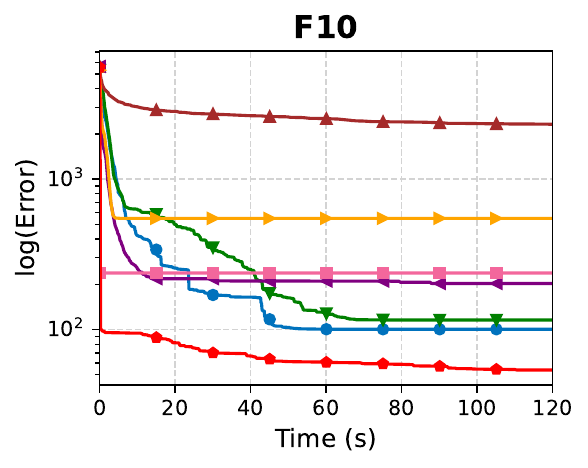}
    \hspace{-0.3cm}
    \includegraphics[width=0.22\textwidth]{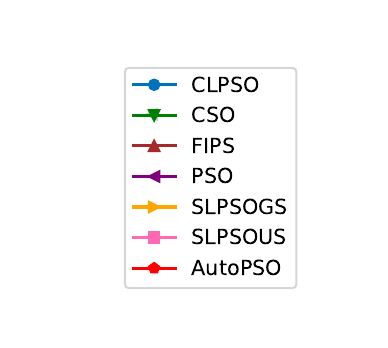}
    
    \caption{Convergence curves on 20D problems in CEC2022 benchmark suite.}
    \label{fig:cec022-20d}
\end{figure}
}

\newcommand{\tabCECFE}{%
\begin{table*}[htbp]
\centering
\caption{Comparison between AutoPSO with other PSO variants under equal FEs.}
\label{tab:CEC-FE}
\renewcommand{\arraystretch}{1.2}
\scalebox{0.78}{
\begin{threeparttable}

\begin{tabular}{p{0.4cm}cccccccc}
\toprule
Dim & Func &  \textbf{AutoPSO} & \textbf{PSO}     & \textbf{CSO}     & \textbf{CLPSO}   & \textbf{FIPS}    & \textbf{SLPSOGS} & \textbf{SLPSOUS} \\ \midrule
\multirow{6}{*}{10D}
& $F_2$  & \cellcolor{lightgray!30}\textbf{2.36E-09 (6.30E-09)} & 9.39E+00 (3.21E+00)$-$ & 6.76E+00 (3.03E+00)$-$ & 1.47E+00 (2.96E+00)$\approx$ & 6.01E+01 (9.89E+01)$-$ & 1.51E+01 (1.58E+01)$-$ & 8.57E+00 (6.36E-01)$-$ \\
& $F_4$  & \cellcolor{lightgray!30}\textbf{0.00E+00 (0.00E+00)} & 6.15E+00 (1.89E+00)$-$ & 1.72E+00 (7.46E-01)$-$ & 1.63E+00 (8.77E-01)$-$ & 9.30E+00 (1.76E+00)$-$ & 3.26E+00 (1.48E+00)$-$ & 3.53E+00 (1.23E+00)$-$ \\
& $F_5$  & \cellcolor{lightgray!30}\textbf{0.00E+00 (0.00E+00)} & 1.31E-01 (3.05E-01)$-$ & \cellcolor{lightgray!30}\textbf{0.00E+00 (0.00E+00)}$\approx$ & \cellcolor{lightgray!30}\textbf{0.00E+00 (0.00E+00)}$\approx$ & \cellcolor{lightgray!30}\textbf{0.00E+00 (0.00E+00)}$\approx$ & 8.85E-02 (2.80E-01)$-$ & 1.05E-08 (2.28E-08)$\approx$ \\
& $F_6$  & \cellcolor{lightgray!30}\textbf{4.29E+00 (4.74E+00)} & 3.50E+03 (2.23E+03)$-$ & 6.14E+02 (7.40E+02)$-$ & 3.71E+01 (1.88E+01)$-$ & 2.48E+03 (2.56E+03)$-$ & 1.06E+03 (7.95E+02)$-$ & 1.86E+03 (1.31E+03)$-$ \\
& $F_{10}$ & \cellcolor{lightgray!30}\textbf{5.38E+01 (4.13E+01)} & 1.20E+02 (4.27E+01)$-$ & 1.20E+02 (4.10E+01)$-$ & 1.00E+02 (3.33E-02)$\approx$ & 2.07E+02 (2.23E+02)$-$ & 1.42E+02 (4.19E+01)$-$ & 1.20E+02 (3.76E+01)$-$ \\
& $F_{12}$ & \cellcolor{lightgray!30}\textbf{1.59E+02 (5.95E-04)} & 1.64E+02 (1.85E+00)$-$ & 1.65E+02 (1.26E+00)$-$ & 1.61E+02 (1.28E+00)$-$ & 1.71E+02 (7.44E+00)$-$ & 1.66E+02 (7.66E-01)$-$ & 1.65E+02 (6.97E-01)$-$ \\

\midrule
\multirow{6}{*}{20D}
& $F_2$  & \cellcolor{lightgray!30}\textbf{2.36E-09 (6.30E-09)} & 7.76E+01 (2.87E+01)$-$ & 4.91E+01 (6.95E-05)$-$ & 4.23E+01 (1.52E+01)$-$ & 5.63E+01 (1.61E+01)$-$ & 6.06E+01 (1.16E+01)$-$ & 4.94E+01 (3.84E-01)$-$ \\
& $F_4$  & \cellcolor{lightgray!30}\textbf{0.00E+00 (0.00E+00)} & 3.58E+01 (1.03E+01)$-$ & 5.70E+00 (1.21E+00)$-$ & 1.55E+01 (3.39E+00)$-$ & 7.54E+01 (5.75E+00)$-$ & 8.05E+00 (3.59E+00)$-$ & 8.59E+00 (1.91E+00)$-$ \\
& $F_5$  & \cellcolor{lightgray!30}\textbf{0.00E+00 (0.00E+00)} & 1.83E+01 (3.51E+01)$-$ & \cellcolor{lightgray!30}\textbf{0.00E+00 (0.00E+00)}$\approx$ & 8.14E-03 (2.57E-02)$\approx$ & 6.31E+01 (1.08E+02)$-$ & 1.79E+00 (3.70E+00)$-$ & 2.27E-01 (6.62E-01)$-$ \\
& $F_6$  & \cellcolor{lightgray!30}\textbf{4.57E+00 (5.27E+00)} & 7.10E+03 (7.99E+03)$-$ & 1.43E+03 (1.97E+03)$-$ & 1.85E+02 (1.32E+02)$-$ & 1.15E+07 (1.70E+07)$-$ & 8.56E+02 (8.33E+02)$-$ & 1.01E+03 (9.57E+02)$-$ \\
& $F_{10}$ & \cellcolor{lightgray!30}\textbf{5.38E+01 (4.13E+01)} & 1.74E+02 (1.52E+02)$-$ & 1.10E+02 (3.25E+01)$\approx$ & 1.00E+02 (3.33E-02)$-$ & 1.25E+03 (6.06E+02)$-$ & 4.54E+02 (4.03E+02)$-$ & 2.96E+02 (2.43E+02)$-$ \\
& $F_{12}$ & \cellcolor{lightgray!30}\textbf{1.59E+02 (5.93E-04)} & 2.61E+02 (1.58E+01)$-$ & 2.42E+02 (4.25E+00)$-$ & 2.37E+02 (2.78E+00)$-$ & 2.82E+02 (3.46E+01)$-$ & 2.58E+02 (7.12E+00)$-$ & 2.59E+02 (8.17E+00)$-$ \\

\bottomrule
\end{tabular}
\begin{tablenotes}[flushleft]
\footnotesize
\item[*] The Wilcoxon rank-sum tests (significance level $\alpha=0.05$) were conducted between AutoPSO and each algorithm individually.
\end{tablenotes}

\end{threeparttable}
}
\end{table*}
}

\newcommand{\tabBasic}{%
\begin{table}[t]
\centering
\caption{Performance comparison of algorithms on large-scale benchmark functions}
\label{tab:basic}
\renewcommand{\arraystretch}{1.8}
\scalebox{0.63}{
\begin{threeparttable}
\begin{tabular}{c c c c c c c c c}
\hline
\textbf{Func} & \textbf{Dim} & \textbf{AutoPSO} & \textbf{PSO} & \textbf{CSO} & \textbf{CLPSO} & \textbf{FIPS} & \textbf{SLPSOGS} & \textbf{SLPSOUS} \\
\hline
\multirow{6}{*}{\rotatebox{90}{Ackley}}
 & 50  & \cellcolor{lightgray!30}\textbf{\makecell{0.00E+00\\(0.00E+00)}} 
 & \makecell{7.49E+00\\(3.50E+00)}$-$
 & \makecell{2.52E-01\\(1.12E-02)}$-$
 & \makecell{5.52E-01\\(2.41E-02)}$-$
 & \makecell{2.00E+01\\(7.32E-03)}$-$
 & \makecell{1.51E+00\\(4.93E-01)}$-$
 & \makecell{1.33E+00\\(6.79E-01)}$-$ \\
 
 & 100 & \cellcolor{lightgray!30}\textbf{\makecell{3.94E-06\\(6.74E-07)}} 
 & \makecell{1.37E+01\\(2.96E+00)}$-$
 & \makecell{4.50E-01\\(1.81E-02)}$-$
 & \makecell{1.17E+00\\(4.20E-01)}$-$
 & \makecell{2.02E+01\\(7.81E-02)}$-$
 & \makecell{5.38E+00\\(9.13E-01)}$-$
 & \makecell{4.92E+00\\(7.16E-01)}$-$ \\ 

 & 200  & \cellcolor{lightgray!30}\textbf{\makecell{4.98E-02\\(7.16E-02)}} 
 & \makecell{1.87E+01\\(1.37E+00)}$-$
 & \makecell{8.67E-01\\(2.76E-02)}$-$
 & \makecell{3.55E+00\\(3.17E-01)}$-$
 & \makecell{2.04E+01\\(4.51E-02)}$-$
 & \makecell{1.03E+01\\(6.75E-01)}$-$
 & \makecell{1.10E+01\\(6.15E-01)}$-$ \\  
 \cline{2-9}
& 500  & \cellcolor{lightgray!30}\textbf{\makecell{4.87E-01\\(8.43E-02)}}
 & \makecell{1.98E+01\\(4.25E-01)}$-$
 & \makecell{1.07E+00\\(3.61E-01)}$-$
 & \makecell{6.86E+00\\(4.96E-01)}$-$
 & \makecell{2.05E+01\\(5.54E-02)}$-$
 & \makecell{1.46E+01\\(4.30E-01)}$-$
 & \makecell{1.54E+01\\(4.84E-01)}$-$ \\
 
 & 1000 & \cellcolor{lightgray!30}\textbf{\makecell{8.16E-01\\(5.13E-02)}}
 & \makecell{2.00E+01\\(5.87E-03)}$-$
 & \makecell{2.51E+00\\(1.99E-01)}$-$
 & \makecell{1.05E+01\\(5.54E-01)}$-$
 & \makecell{2.05E+01\\(5.12E-02)}$-$
 & \makecell{1.65E+01\\(1.59E-01)}$-$
 & \makecell{1.70E+01\\(1.89E-01)}$-$ \\
 
 & 2000 & \cellcolor{lightgray!30}\textbf{\makecell{1.04E+00\\(9.12E-02)}}
 & \makecell{2.00E+01\\(2.68E-01)}$-$
 & \makecell{5.78E+00\\(4.88E-01)}$-$
 & \makecell{1.84E+01\\(1.93E+00)}$-$
 & \makecell{2.05E+01\\(4.53E-02)}$-$
 & \makecell{1.75E+01\\(1.01E-01)}$-$
 & \makecell{1.78E+01\\(1.78E-01)}$-$ \\
 
\hline

\multirow{6}{*}{\rotatebox{90}{Rastrigin}}
 & 50   & \cellcolor{lightgray!30}\textbf{\makecell{5.98E-03\\(6.68E-03)}} 
 & \makecell{1.03E+02\\(2.42E+01)}$-$
 & \makecell{5.59E+01\\(6.65E+00)}$-$
 & \makecell{1.20E+02\\(1.16E+01)}$-$
 & \makecell{6.01E+02\\(1.74E+01)}$-$
 & \makecell{4.09E+01\\(7.29E+00)}$-$
 & \makecell{3.83E+01\\(5.97E+00)}$-$ \\

 & 100  & \cellcolor{lightgray!30}\textbf{\makecell{6.22E-02\\(5.42E-02)}} 
 & \makecell{4.13E+02\\(6.10E+01)}$-$
 & \makecell{2.13E+02\\(2.01E+01)}$-$
 & \makecell{3.71E+02\\(1.53E+01)}$-$
 & \makecell{1.51E+03\\(2.01E+01)}$-$
 & \makecell{1.28E+02\\(1.40E+01)}$-$
 & \makecell{1.34E+02\\(1.61E+01)}$-$ \\

 & 200  & \cellcolor{lightgray!30}\textbf{\makecell{4.85E-01\\(2.01E-01)}} 
 & \makecell{1.30E+03\\(1.01E+02)}$-$
 & \makecell{7.13E+02\\(4.60E+01)}$-$
 & \makecell{1.06E+03\\(4.25E+01)}$-$
 & \makecell{3.30E+03\\(4.12E+01)}$-$
 & \makecell{5.17E+02\\(4.79E+01)}$-$
 & \makecell{4.94E+02\\(4.28E+01)}$-$ \\
 \cline{2-9}
  & 500  & \cellcolor{lightgray!30}\textbf{\makecell{1.08E+01\\(3.08E+00)}}
 & \makecell{4.92E+03\\(2.97E+02)}$-$
 & \makecell{8.39E+02\\(5.36E+01)}$-$
 & \makecell{1.02E+03\\(7.21E+01)}$-$
 & \makecell{8.64E+03\\(6.45E+01)}$-$
 & \makecell{2.48E+03\\(1.01E+02)}$-$
 & \makecell{2.44E+03\\(1.18E+02)}$-$ \\

 & 1000 & \cellcolor{lightgray!30}\textbf{\makecell{4.79E+01\\(9.23E+00)}}
 & \makecell{1.12E+04\\(2.51E+02)}$-$
 & \makecell{2.76E+03\\(1.77E+02)}$-$
 & \makecell{2.70E+03\\(1.48E+02)}$-$
 & \makecell{1.76E+04\\(1.17E+02)}$-$
 & \makecell{6.80E+03\\(1.43E+02)}$-$
 & \makecell{6.90E+03\\(2.34E+02)}$-$ \\

 & 2000 & \cellcolor{lightgray!30}\textbf{\makecell{1.49E+02\\(2.99E+01)}}
 & \makecell{2.48E+04\\(4.22E+02)}$-$
 & \makecell{7.53E+03\\(2.02E+02)}$-$
 & \makecell{8.43E+03\\(3.51E+02)}$-$
 & \makecell{3.59E+04\\(2.02E+02)}$-$
 & \makecell{1.66E+04\\(2.73E+02)}$-$
 & \makecell{1.70E+04\\(2.25E+02)}$-$ \\
\bottomrule
\end{tabular}
\begin{tablenotes}[flushleft]
\footnotesize
\item[*] The Wilcoxon rank-sum tests (significance level $\alpha=0.05$) were conducted between AutoPSO and each algorithm individually.
\end{tablenotes}

\end{threeparttable}
}
\end{table}
}

\newcommand{\figAckley}{%
\begin{figure}[htbp]
    \centering
    \includegraphics[width=0.22\textwidth]{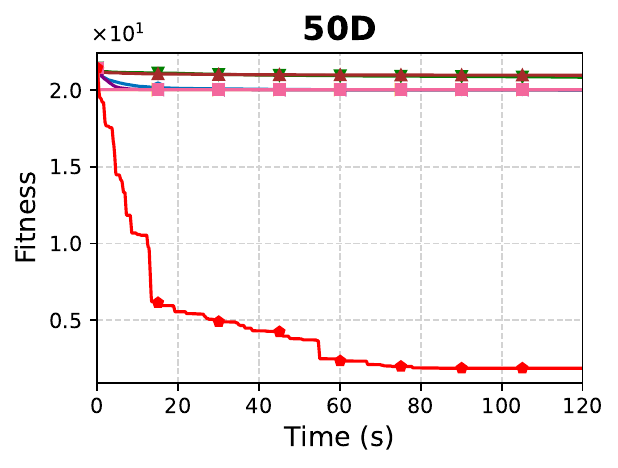}
    \hspace{-0.1cm}
    \includegraphics[width=0.22\textwidth]{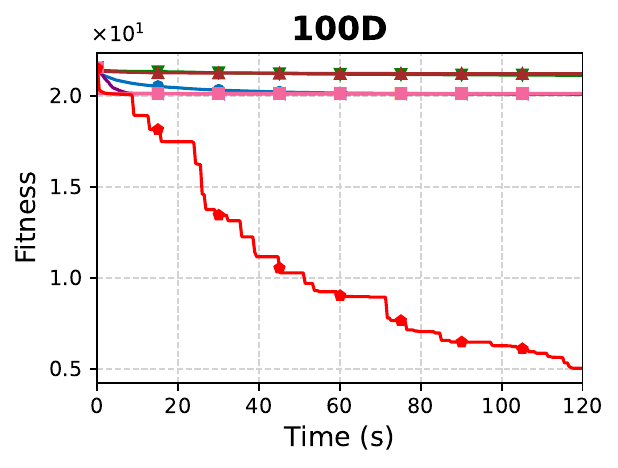}

    \vspace{-0.5cm}
    
    \includegraphics[width=0.22\textwidth]{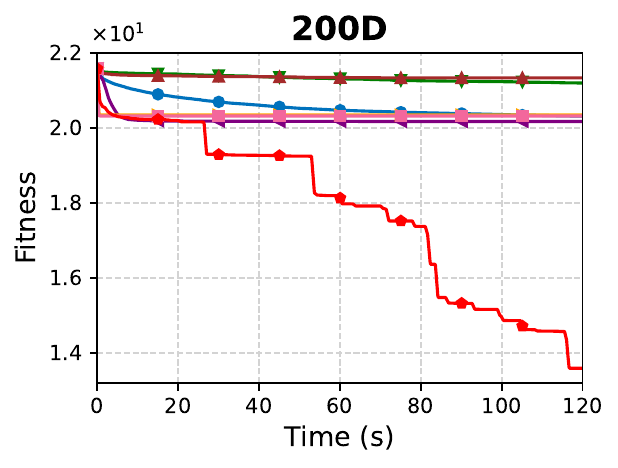}
    \hspace{-0.3cm}
    \includegraphics[width=0.22\textwidth]{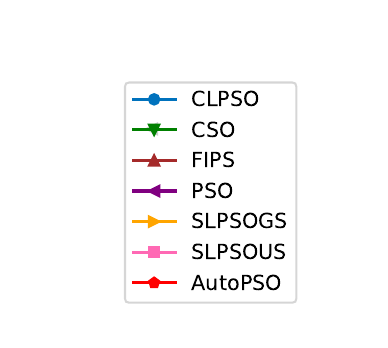}   
    \caption{Convergence curves on shifted and rotated Ackley function.}
    \label{fig:ackley}
\end{figure}
}

\newcommand{\figBrax}{%
\begin{figure}[htbp]
   
    \centering
    \includegraphics[width=0.22\textwidth]{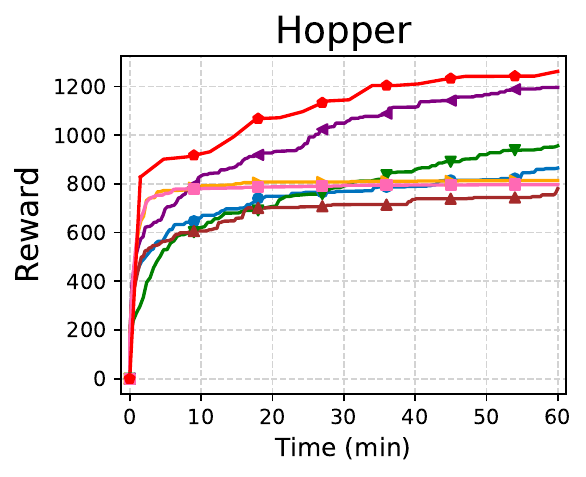}
    \hspace{-0.1cm}
    \includegraphics[width=0.22\textwidth]{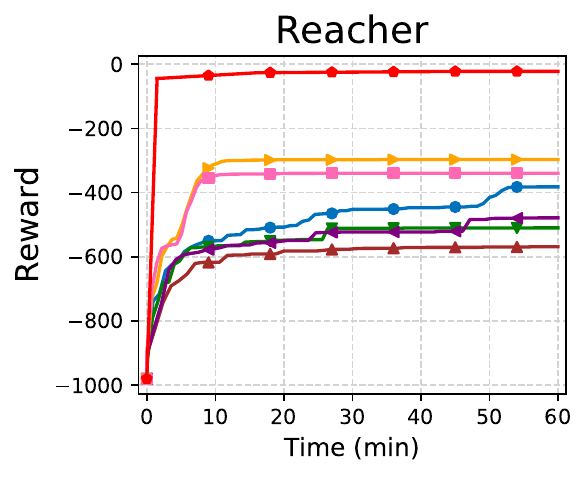}

    \vspace{-0.3cm}

    \includegraphics[width=0.22\textwidth]{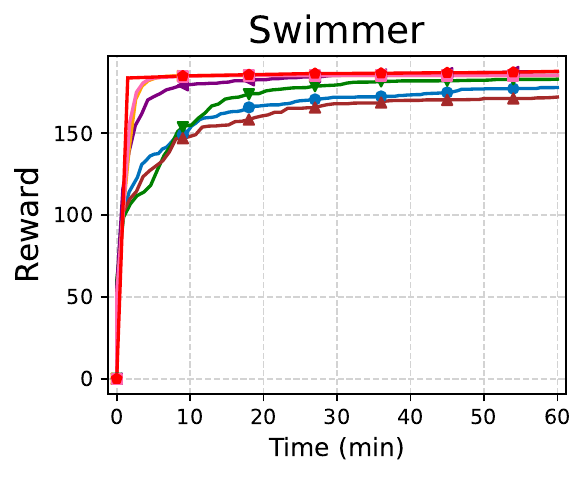}
    \hspace{-0.3cm}
    \includegraphics[width=0.22\textwidth]{fig/data/brax/label.pdf}
    
    \caption{The reward curves achieved by AutoPSO and baseline PSO variants when applied to each robot control task.}
    \label{fig:brax}
\end{figure}
}

\newcommand{\tabAad}{%
\begin{table}[htpb]
\centering
\caption{Comparison with recent MetaBBO methods on CEC2022-20D.}
\label{tab:aad}
\renewcommand{\arraystretch}{1.8}
\scalebox{0.75}{
\begin{threeparttable}
\begin{tabular}{ccccccc}
\toprule
\textbf{Func} & \textbf{AutoPSO} & \textbf{DEDQN} & \textbf{SYMBOL} & \textbf{GLEET} & \textbf{RLHPSDE} & \textbf{RLEPSO} \\
\midrule
$F_{1}$  & \makecell{5.48e-09\\(1.73e-08)} & \makecell{4.94e+03\\(1.57e+03) $-$} & \makecell{1.02e+00\\(6.59e-01) $-$} & \makecell{6.73e+01\\(2.13e+02) $-$} & \cellcolor{lightgray!30}\textbf{\makecell{0.00e+00\\(0.00e+00) $+$}} & \cellcolor{lightgray!30}\textbf{\makecell{0.00e+00\\(0.00e+00) $+$}} \\

$F_{2}$  & \cellcolor{lightgray!30}\textbf{\makecell{1.62e+01\\(1.14e+01)}} & \makecell{4.91e+01\\(6.49e-02) $-$} & \makecell{6.07e+01\\(1.19e+01) $-$} & \makecell{2.98e+01\\(2.44e+01) $-$} & \makecell{4.76e+01\\(2.02e+00) $-$} & \makecell{3.09e+01\\(2.34e+01) $-$} \\

$F_{3}$  & \cellcolor{lightgray!30}\textbf{\makecell{0.00e+00\\(0.00e+00)}} & \cellcolor{lightgray!30}\textbf{\makecell{0.00e+00\\(0.00e+00) $\approx$}} & \makecell{3.62e+01\\(9.58e+00) $-$} & \makecell{5.13e-01\\(1.13e+00) $-$} & \cellcolor{lightgray!30}\textbf{\makecell{0.00e+00\\(0.00e+00) $\approx$}} & \makecell{9.03e-01\\(1.21e+00) $-$} \\

$F_{4}$  & \cellcolor{lightgray!30}\textbf{\makecell{4.34e+00\\(2.26e+00)}} & \makecell{9.08e+01\\(5.80e+00) $-$} & \makecell{6.50e+01\\(8.43e+00) $-$} & \makecell{5.76e+01\\(1.64e+01) $-$} & \makecell{3.35e+01\\(6.04e+00) $-$} & \makecell{5.74e+01\\(1.41e+01) $-$} \\

$F_{5}$  & \cellcolor{lightgray!30}\textbf{\makecell{0.00e+00\\(0.00e+00)}} & \makecell{3.30e-08\\(7.66e-08) $\approx$} & \makecell{8.20e+02\\(5.22e+02) $-$} & \makecell{1.96e+00\\(1.39e+00) $-$} & \cellcolor{lightgray!30}\textbf{\makecell{0.00e+00\\(0.00e+00) $\approx$}} & \makecell{1.91e+00\\(1.11e+00) $-$} \\

$F_{6}$  & \makecell{9.53e+01\\(2.17e+01)} & \makecell{1.54e+05\\(3.75e+04) $-$} & \makecell{5.26e+03\\(2.88e+03) $-$} & \makecell{5.13e+01\\(2.06e+01) $+$} & \cellcolor{lightgray!30}\textbf{\makecell{2.99e-01\\(5.45e-02) $+$}} & \makecell{2.23e+03\\(2.46e+03) $-$} \\

$F_{7}$  & {\makecell{1.57e+01\\(8.01e+00)}} & \makecell{4.63e+01\\(5.29e+00) $-$} & \makecell{9.04e+01\\(2.11e+01) $-$} & \makecell{7.65e+01\\(5.97e+01) $-$} & \cellcolor{lightgray!30}\textbf{\makecell{5.53e+00\\(7.76e+00) $+$}} & \makecell{4.74e+01\\(1.21e+01) $-$} \\

$F_{8}$  & {\makecell{2.12e+01\\(5.18e-01)}} & \makecell{3.17e+01\\(1.39e+00) $-$} & \makecell{3.42e+01\\(7.54e+00) $-$} & \makecell{6.86e+01\\(5.81e+01) $-$} & \cellcolor{lightgray!30}\textbf{\makecell{1.04e+01\\(9.00e+00) $+$}} & \makecell{2.11e+01\\(3.44e-01) $\approx$} \\

$F_{9}$  & \cellcolor{lightgray!30}\textbf{\makecell{1.81e+02\\(3.01e-04)}} & \cellcolor{lightgray!30}\textbf{\makecell{1.81e+02\\(1.27e-14) $\approx$}} & \makecell{1.82e+02\\(3.25e-01) $-$} & \makecell{2.00e+02\\(1.54e+01) $\approx$} & \cellcolor{lightgray!30}\textbf{\makecell{1.81e+02\\(2.54e-14) $\approx$}} & \cellcolor{lightgray!30}\textbf{\makecell{1.81e+02\\(3.11e-14) $\approx$}} \\

$F_{10}$ & \cellcolor{lightgray!30}\textbf{\makecell{1.00e+02\\(2.63e-02)}} & \makecell{1.37e+02\\(7.21e+01) $-$} & \cellcolor{lightgray!30}\textbf{\makecell{1.00e+02\\(2.74e-02) $\approx$}} & \makecell{3.39e+02\\(1.46e+02) $-$} & \cellcolor{lightgray!30}\textbf{\makecell{1.00e+02\\(2.53e-02) $\approx$}} & \makecell{2.90e+02\\(1.31e+02) $-$} \\

$F_{11}$ & \cellcolor{lightgray!30}\textbf{\makecell{2.46e+02\\(1.15e+02)}} & \makecell{3.00e+02\\(3.27e-05) $\approx$} & \makecell{3.30e+02\\(3.62e+01) $-$} & \makecell{3.86e+02\\(1.71e+02) $-$} & \makecell{3.40e+02\\(4.90e+01) $-$} & \makecell{3.20e+02\\(4.00e+01) $-$} \\

$F_{12}$ & \cellcolor{lightgray!30}\textbf{\makecell{2.32e+02\\(7.18e-01)}} & \makecell{2.38e+02\\(4.60e+00) $-$} & \makecell{3.96e+02\\(3.43e+01) $-$} & \makecell{2.82e+02\\(3.29e+01) $-$} & \makecell{2.37e+02\\(2.59e+00) $\approx$} & \makecell{2.60e+02\\(1.39e+01) $-$} \\

\bottomrule
\end{tabular}
\begin{tablenotes}[flushleft]
\footnotesize
\item[*] The Wilcoxon rank-sum tests (significance level $\alpha=0.05$) were conducted between AutoPSO and each algorithm individually.
\end{tablenotes}

\end{threeparttable}
}
\end{table}
}

\newcommand{\figGeneralize}{%
\begin{figure}[htbp]
   
    \centering
    \includegraphics[width=0.22\textwidth]{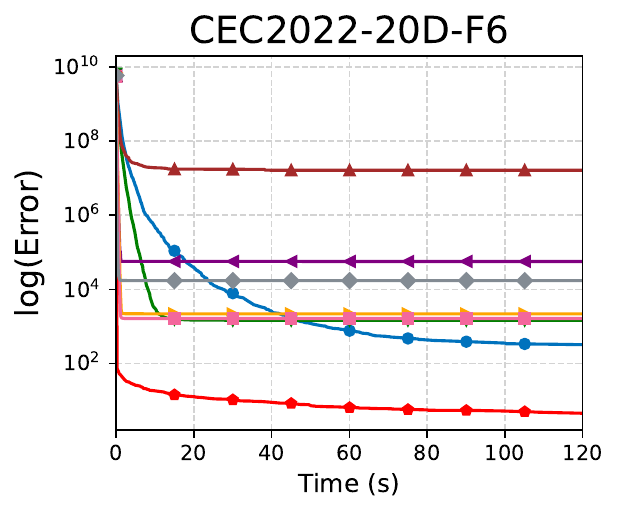}   
    \hspace{-0.2cm}  \includegraphics[width=0.22\textwidth]{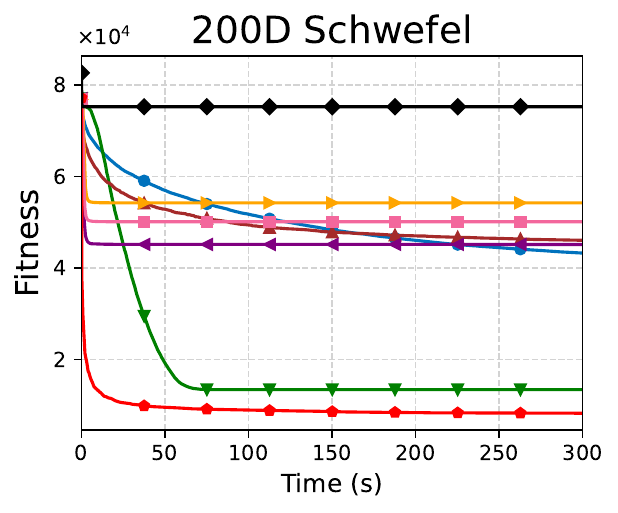}

    \vspace{-0.1cm}
    
    \includegraphics[width=0.23\textwidth]{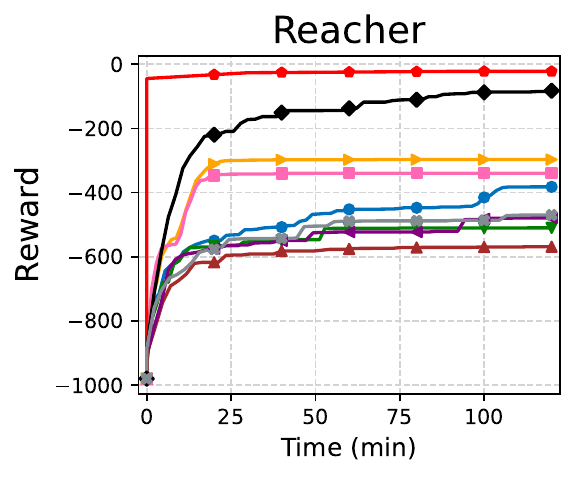}
    \hspace{0.9cm}  \includegraphics[width=0.16\textwidth]{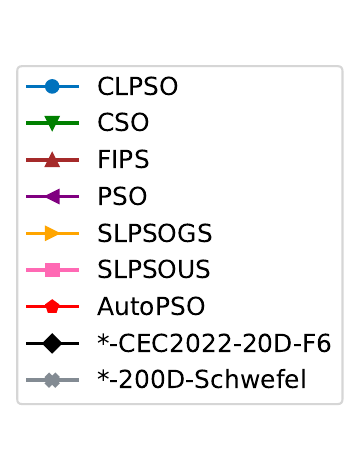}
    
    \caption{Generalization of discovered PSO variants across different functions.}
    \label{fig:generalize}
\end{figure}
}

\newcommand{\tabGeneralize}{%
\begin{table}[htbp]
\caption{Generalization study of AutoPSO on CEC2022-20D.}
\label{tab:generalize}
\centering
\scalebox{0.55}{
\begin{threeparttable}
\begin{tabular}{cccccc}
\toprule
\multirow{2}{*}{\textbf{Func}} 
& \multirow{2}{*}{\textbf{AutoPSO}} 
& \multicolumn{4}{c}{\textbf{Variants find on}} \\
\cline{3-6}
&  & \textbf{-F6} & \textbf{-F8} & \textbf{-F10} & \textbf{-F12} \\
\midrule
$F_1$  & \cellcolor{lightgray!30}\textbf{0.00E+00 (0.00E+00)} & \cellcolor{lightgray!30}\textbf{0.00E+00 (0.00E+00)}$\approx$ & 2.40E+02 (3.66E+02)$-$ & \cellcolor{lightgray!30}\textbf{0.00E+00 (0.00E+00)}$\approx$ & \cellcolor{lightgray!30}\textbf{0.00E+00 (0.00E+00)}$\approx$ \\
$F_2$  & \cellcolor{lightgray!30}\textbf{2.36E-09 (6.30E-09)} & 3.98E+01 (1.88E+01)$-$ & 5.22E+01 (1.01E+01)$-$ & 6.88E+01 (2.78E+01)$-$ & 4.80E+01 (1.87E+00)$-$ \\
$F_3$  & \cellcolor{lightgray!30}\textbf{0.00E+00 (0.00E+00)} & 8.36E-03 (4.10E-03)$-$ & 3.42E-02 (1.08E-01)$-$ & 1.09E+00 (1.22E+00)$-$ & \cellcolor{lightgray!30}\textbf{0.00E+00 (0.00E+00)}$\approx$ \\
$F_4$  & \cellcolor{lightgray!30}\textbf{0.00E+00 (0.00E+00)} & 3.06E+01 (5.59E+00)$-$ & 1.93E+01 (6.66E+00)$-$ & 3.40E+01 (1.12E+01)$-$ & 6.02E+01 (6.67E+00)$-$ \\
$F_5$  & \cellcolor{lightgray!30}\textbf{0.00E+00 (0.00E+00)} & 9.78E+00 (1.31E+01)$-$ & {6.89E-01 (1.41E+00)}$\approx$ & 1.74E+01 (5.41E+01)$-$ & \cellcolor{lightgray!30}\textbf{0.00E+00 (0.00E+00)}$-$ \\
$F_7$  & \cellcolor{lightgray!30}\textbf{0.00E+00 (0.00E+00)} & 5.85E+01 (2.19E+01)$-$ & 3.25E+01 (5.35E+00)$-$ & 2.98E+01 (6.96E+00)$-$ & 2.28E+01 (1.76E+00)$-$ \\
$F_9$  & \cellcolor{lightgray!30}\textbf{1.72E+02 (9.16E+01)} & 1.81E+02 (1.44E-03)$-$ & 1.84E+02 (7.26E+00)$-$ & 1.87E+02 (1.96E+01)$\approx$ & 1.81E+02 (3.84E-04)$-$ \\
$F_{11}$ & \cellcolor{lightgray!30}\textbf{0.00E+00 (0.00E+00)} & 3.18E+02 (3.86E+01)$-$ & 4.20E+02 (1.68E+02)$-$ & 6.09E+02 (3.79E+02)$-$ & 3.36E+02 (4.81E+01)$-$ \\

\bottomrule
\end{tabular}

\begin{tablenotes}[flushleft]
\footnotesize
\item[*] The Wilcoxon rank-sum tests (significance level $\alpha=0.05$) were conducted between AutoPSO and each algorithm individually.
\end{tablenotes}

\end{threeparttable}
}
\end{table}
}

\newcommand{\tabComponent}{%
\begin{table*}[htbp]
\caption{Ablation study of AutoPSO on CEC2022-20D.}
\label{tab:component}
\centering
\scalebox{0.66}{
\begin{threeparttable}
\begin{tabular}{cc|ccc|cccc|c}
\toprule
\textbf{Func} 
& \textbf{AutoPSO} 
& \textbf{AutoPSO-c} 
& \textbf{w/o. acc.} 
& \textbf{w/o. weight} 
& \textbf{AutoPSO-e} 
& \textbf{w/o. random} 
& \textbf{w/o. self} 
& \textbf{w/o. social} 
& \textbf{AutoPSO-m} \\
\midrule

$F_2$  & \cellcolor{lightgray!30}\textbf{2.36E-09 (6.30E-09)} & 7.23E-01 (1.50E+00)$-$ & 4.79E-07 (1.07E-06)$\approx$ & 1.72E-04 (4.76E-04)$-$ & 6.14E-03 (3.03E-02)$-$ & 1.62E-05 (5.07E-05)$-$ & 1.33E-06 (2.00E-06)$-$ & 3.16E-02 (5.79E-02)$-$ & 2.26E-08 (3.58E-08)$-$ \\
$F_4$  & \cellcolor{lightgray!30}\textbf{0.00E+00 (0.00E+00)} & 8.01E-01 (5.65E-01)$-$ & \cellcolor{lightgray!30}\textbf{0.00E+00 (0.00E+00)}$\approx$ & \cellcolor{lightgray!30}\textbf{0.00E+00 (0.00E+00)}$\approx$ & 6.33E+00 (1.25E+00)$-$ & 1.04E+00 (9.34E-01)$-$ & \cellcolor{lightgray!30}\textbf{0.00E+00 (0.00E+00)}$\approx$ & \cellcolor{lightgray!30}\textbf{0.00E+00 (0.00E+00)}$\approx$ & \cellcolor{lightgray!30}\textbf{0.00E+00 (0.00E+00)}$\approx$ \\
$F_6$  & \cellcolor{lightgray!30}\textbf{4.57E+00 (5.27E+00)} & 5.61E+01 (2.42E+01)$-$ & 5.18E+00 (3.23E+00)$\approx$ & 1.59E+01 (1.07E+01)$-$ & 6.18E+01 (3.25E+01)$-$ & 2.60E+01 (1.19E+01)$-$ & 1.69E+01 (1.08E+01)$-$ & 1.40E+01 (8.76E+00)$-$ & 7.50E+00 (7.23E+00)$\approx$ \\
$F_9$  & \cellcolor{lightgray!30}\textbf{1.72E+02 (9.16E+01)} & 1.81E+02 (8.11E-05)$-$ & 1.81E+02 (4.68E-05)$-$ & 1.81E+02 (1.61E-05)$-$ & 1.81E+02 (8.94E-05)$-$ & 1.81E+02 (6.65E-05)$-$ & 1.81E+02 (3.40E-05)$-$ & 1.81E+02 (1.27E-05)$-$ & 1.92E+02 (7.15E+01)$\approx$ \\
$F_{10}$ & {5.38E+01 (4.13E+01)} & 2.13E+01 (2.78E+01)$+$ & 3.05E+01 (4.29E+01)$\approx$ & 5.60E-01 (1.01E+00)$+$ & 5.13E+01 (4.32E+01)$\approx$ & 1.00E+02 (1.44E-02)$-$ & 1.97E+00 (4.86E+00)$+$ & \cellcolor{lightgray!30}\textbf{4.89E-01 (6.41E-01)}$+$ & 4.23E+01 (4.33E+01)$\approx$ \\
$F_{12}$ & \cellcolor{lightgray!30}\textbf{1.59E+02 (5.93E-04)} & 2.30E+02 (1.04E+00)$-$ & 2.30E+02 (7.46E-01)$-$ & 2.28E+02 (7.13E-01)$-$ & 2.34E+02 (2.36E+00)$-$ & 2.29E+02 (1.79E+00)$-$ & 2.28E+02 (8.08E-01)$-$ & 2.28E+02 (4.90E-01)$-$ & \cellcolor{lightgray!30}\textbf{1.59E+02 (2.47E-03)}$-$ \\

\bottomrule
\end{tabular}
\begin{tablenotes}[flushleft]
\footnotesize
\item[*] The Wilcoxon rank-sum tests (significance level $\alpha=0.05$) were conducted between AutoPSO and each variant individually.
\end{tablenotes}

\end{threeparttable}
}
\end{table*}

}

\newcommand{\tabSubswarm}{%
\begin{table}[t]
\centering
\caption{Comparison of different swarm partition strategies on CEC2022-20D.}
\label{tab:subswarm}

\scalebox{0.56}{
\begin{threeparttable}
\begin{tabular}{cccccc}
\toprule
\textbf{Func} & \textbf{\makecell{AutoPSO\\(2-sub, searched)}} & \textbf{1-swarm} & \textbf{\makecell{2-subswarm \\(fixed, 0.5/0.5)}} & \textbf{\makecell{3-subswarm\\(searched)}} & \textbf{\makecell{4-subswarm\\(searched)}} \\
\midrule

$F_2$ &
\cellcolor{lightgray!30}\textbf{2.36E-09 (6.30E-09)} &
2.26E-08 (3.58E-08)$-$ &
5.62E-05 (1.10E-04)$-$ &
1.98E-05 (4.26E-05)$-$ &
1.55E-05 (3.59E-05)$-$ \\

$F_3$ &
\cellcolor{lightgray!30}\textbf{0.00E+00 (0.00E+00)} &
\cellcolor{lightgray!30}\textbf{0.00E+00 (0.00E+00)}$\approx$ &
1.28E-08 (3.71E-08)$\approx$ &
\cellcolor{lightgray!30}\textbf{0.00E+00 (0.00E+00)}$\approx$ &
\cellcolor{lightgray!30}\textbf{0.00E+00 (0.00E+00)}$\approx$ \\

$F_6$ &
\cellcolor{lightgray!30}\textbf{4.57E+00 (5.27E+00)} &
7.50E+00 (7.23E+00)$\approx$ &
3.85E+01 (1.14E+01)$-$ &
1.90E+01 (1.29E+01)$-$ &
1.79E+01 (1.27E+01)$-$ \\

$F_7$ &
\cellcolor{lightgray!30}\textbf{0.00E+00 (0.00E+00)} &
\cellcolor{lightgray!30}\textbf{0.00E+00 (0.00E+00)}$\approx$ &
3.12E+00 (2.39E+00)$-$ &
1.30E-02 (1.09E-02)$-$ &
1.91E-02 (2.01E-02)$-$ \\

$F_{10}$ &
{5.38E+01 (4.13E+01)} &
4.23E+01 (4.33E+01)$\approx$ &
7.58E+01 (3.99E+01)$-$ &
1.07E+01 (2.84E+01)$+$ &
\cellcolor{lightgray!30}\textbf{1.04E+01 (2.84E+01)}$+$ \\

$F_{12}$ &
\cellcolor{lightgray!30}\textbf{1.59E+02 (5.93E-04)} &
1.59E+02 (2.47E-03)$-$ &
2.30E+02 (2.11E+00)$-$ &
2.28E+02 (8.60E-01)$-$ &
2.29E+02 (9.11E-01)$-$ \\

\bottomrule
\end{tabular}
\begin{tablenotes}[flushleft]
\footnotesize
\item[*] The Wilcoxon rank-sum tests (significance level $\alpha=0.05$) were conducted between AutoPSO and each algorithm individually.
\end{tablenotes}

\end{threeparttable}
}
\end{table}
}

\newcommand{\figOuter}{%
\begin{figure}[htbp]
   
    \centering
    \includegraphics[width=0.22\textwidth]{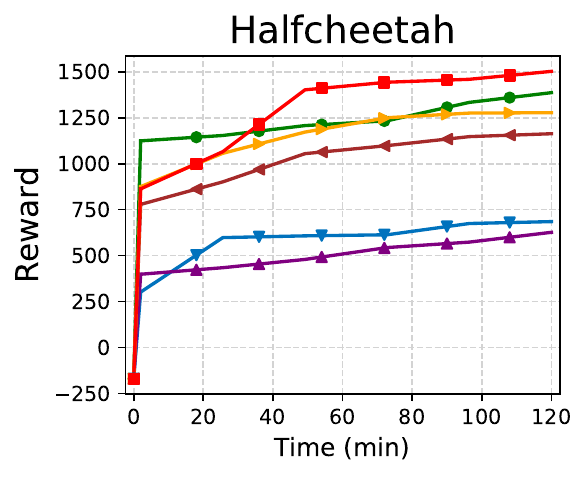}
    \hspace{-0.1cm}
    \includegraphics[width=0.22\textwidth]{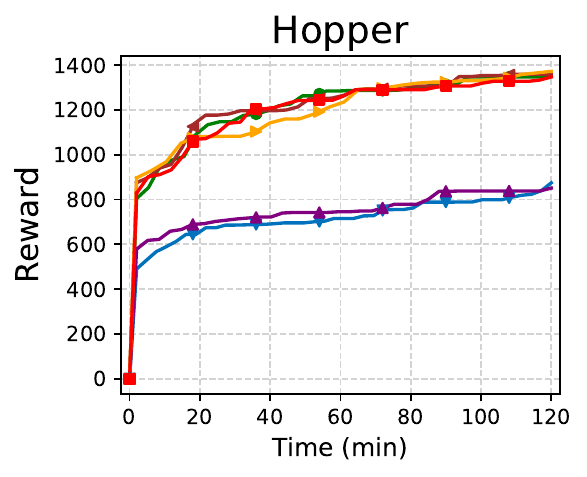}
    
    \vspace{0.05cm}
    
    \includegraphics[width=0.22\textwidth]{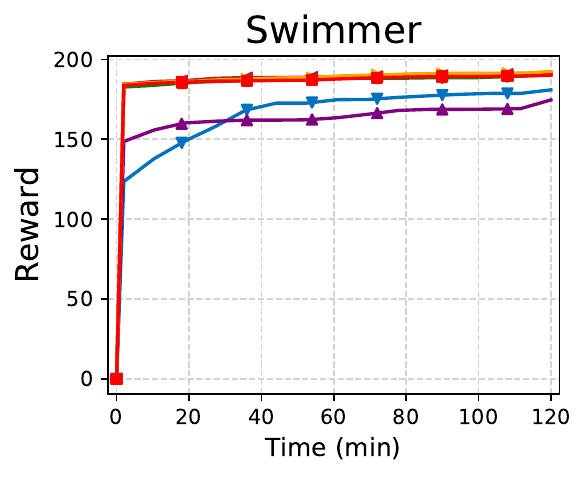}
    \hspace{-0.1cm}
    \includegraphics[width=0.22\textwidth]{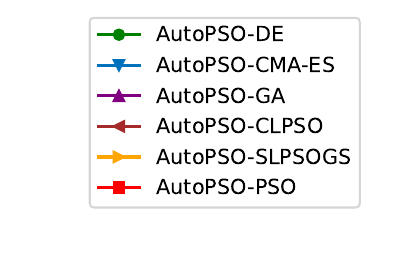}
    
    \caption{The reward curves achieved by AutoPSO with different outer optimizers when applied to each robot control task.}
    \label{fig:outer}
\end{figure}
}

\newcommand{\figScalingPopsize}{%
\begin{figure}[htbp]
    \centering
    \includegraphics[width=0.23\textwidth]{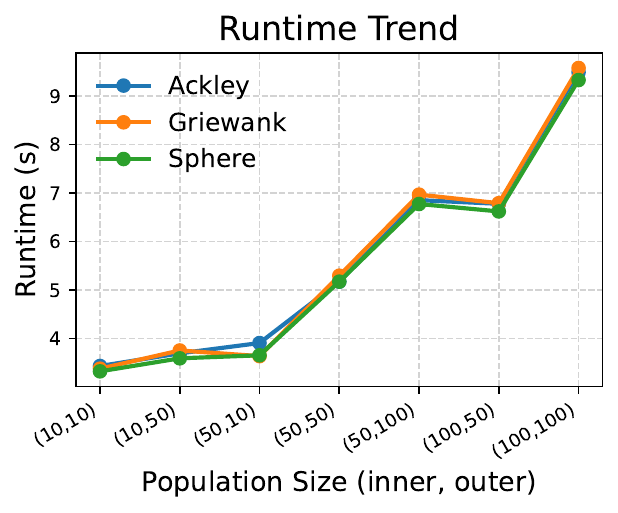}
    \hspace{-0.1cm}
    \includegraphics[width=0.23\textwidth]{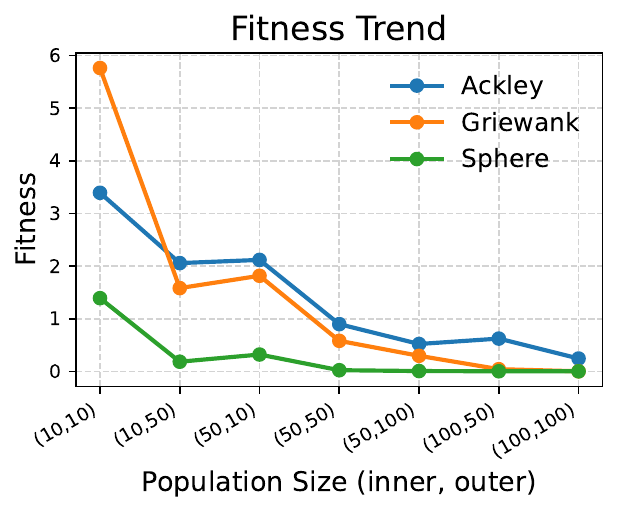}
    
    \caption{Runtime and fitness trends of AutoPSO with varying population sizes on 200D functions over 10 iterations, where lower fitness indicates better performance.}
    \label{fig:scaling-popsize}
\end{figure}
}

\newcommand{\figPopsize}{%
\begin{figure}[htbp]
   
    \centering
    \includegraphics[width=0.22\textwidth]{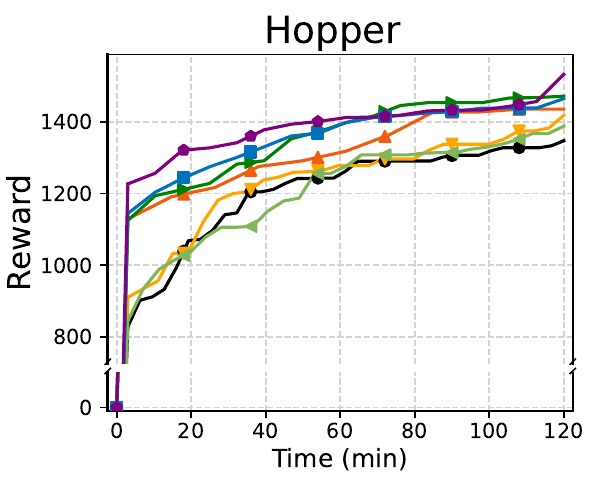}
    \hspace{-0.1cm}
    \includegraphics[width=0.22\textwidth]{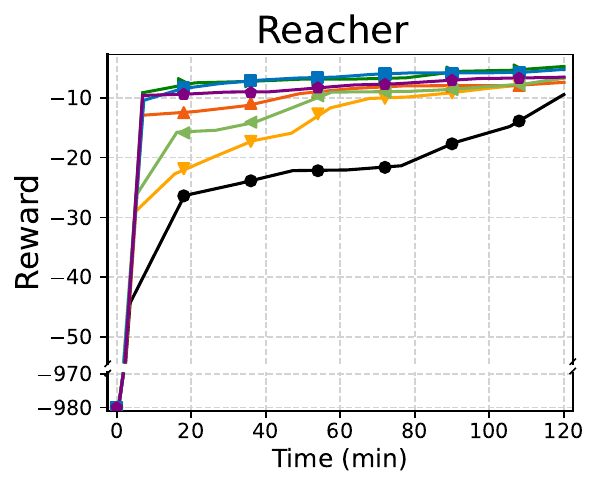}
    
    \vspace{0.05cm}
    
    \includegraphics[width=0.22\textwidth]{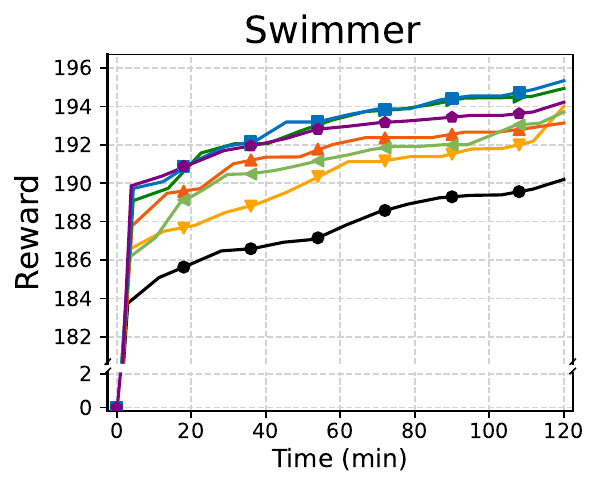}
    \hspace{-0.1cm}
    \includegraphics[width=0.22\textwidth]{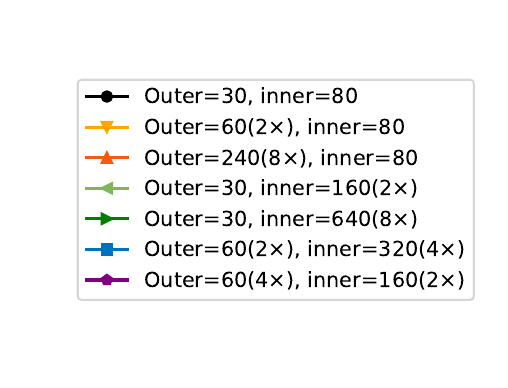}
    
    \caption{Reward curves of AutoPSO with varying population sizes on robot control tasks, where higher reward indicates better performance.}
    \label{fig:popsize}
\end{figure}
}

\newcommand{\figScaling}{%
\begin{figure}[htbp]
    \centering
    \includegraphics[width=0.22\textwidth]{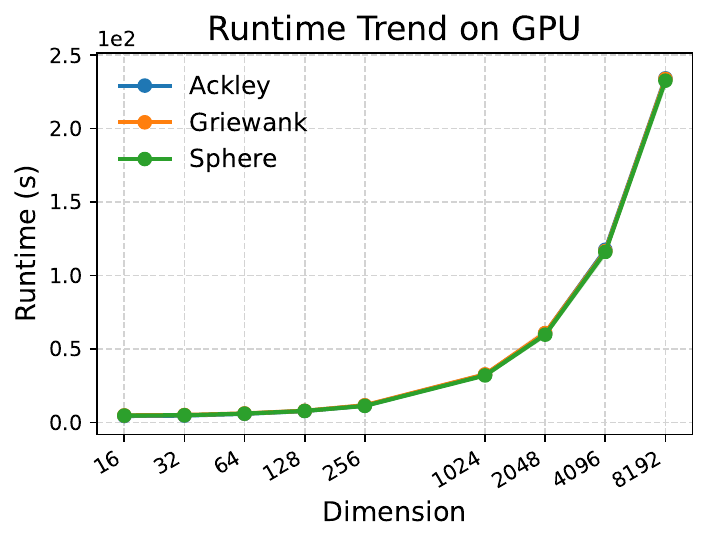}   
    \hspace{-0.1cm}
    \includegraphics[width=0.22\textwidth]{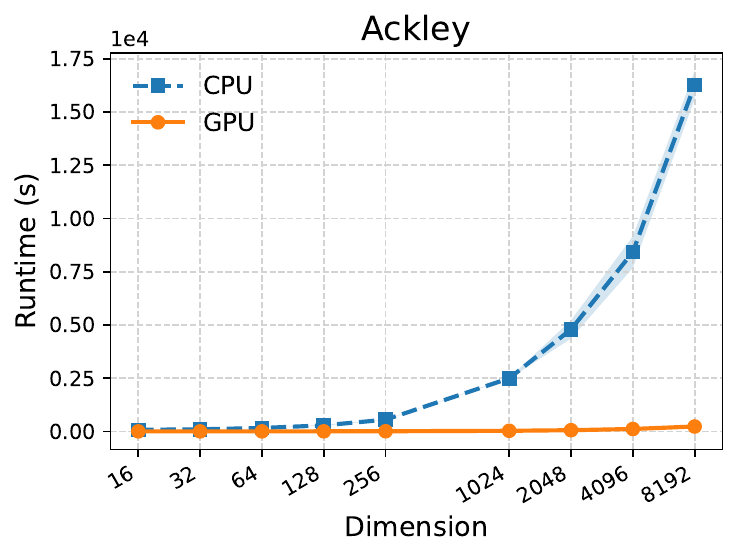}
     
    \caption{Runtime trends of AutoPSO tested under varying problem dimension.}
    \label{fig:scaling}
\end{figure}
}

\newcommand{\algoGPSO}{%
\begin{algorithm}[htbp]
  \caption{Pseudo-code of the generalized PSO}
  \label{alg:gPSO}
  \begin{algorithmic}[1]
    \REQUIRE Population size $N$; objective function $f$;
             maximum generations $T_{\max}$; configuration repository $\theta(\cdot)$
    \ENSURE  Best solution $\mathbf{X}^\star$
    \STATE $\mathbf{X}^{0}=\{\mathbf{X}_1,\!\ldots,\!\mathbf{X}_N\}\gets\textsc{InitPosition}(N)$
    \STATE $\mathbf{V}^{0}=\{\mathbf{V}_1,\!\ldots,\!\mathbf{V}_N\}\gets\textsc{InitVelocity}(N)$
    \STATE $\mathbf{Fitness} \gets \textsc{Evaluate}(f,\mathbf{X}^{0}\bigr)$
    \STATE $t \gets 0$
    \WHILE{$t < T_{\max}$}
        \STATE $t \gets t+1$    
        \STATE /* \emph{\textbf{Component 1}:  partition ratio} $r$ */ \\ $r \gets \theta_{\text{ratio}}$ 
        \STATE $\{\mathbf{X}_{s_1}^{t},\mathbf{X}_{s_2}^{t}\}\gets\textsc{PopulationPartition}(\mathbf{X}^{t}, r)$
        \FOR{each sub-population $\mathbf{X}_{s_{i,i\in\{1,2\}}}^{t} $}
         \STATE /* \emph{\textbf{Component 2}: parameter $w, {c}_{A}$ and ${c}_{B}$} */ \\
          $(w,{c}_{A,s_i},{c}_{B,s_i}) \gets \theta_{\text{parameter}}$
    \STATE /*\emph{\textbf{ Component 3}: learning exemplar $\mathbf p_A$ and $\mathbf p_B$} */ \\
             $(\mathbf p_{A,s_i},\mathbf p_{B,s_i}) \gets \theta_{\text{exemplar}}$ 
            
            \STATE /* update velocity and position */ \\ 
             $\mathbf{V}_{s_i}^{t+1} \gets
                   w * \mathbf{V}_{s_i}^{t}
                   + {c}_{A,s_i} * (\mathbf p_{A,s_i}-\mathbf{X}_{s_i}^{t})
                   + {c}_{B,s_i} * (\mathbf p_{B,s_i}-\mathbf{X}_{s_i}^{t})$ \\
             $\mathbf{X}_{s_i}^{t+1} \gets \mathbf{X}_{s_i}^{t} + \mathbf{V}_{s_i}^{t+1}$
            
        \ENDFOR
        \STATE $\mathbf{Fitness} \gets \textsc{Evaluate}(f,\mathbf{X^{t+1}}\bigr)$
    \ENDWHILE
    \STATE $\mathbf{X}^\star, {\mathbf{Fitness}}^\star \gets
           \textsc{Best}(\mathbf{X}, \mathbf{Fitness}$)
    \RETURN $\mathbf{X}^\star, {\mathbf{Fitness}}^\star$
  \end{algorithmic}
\end{algorithm}
}

\newcommand{\algoAutoPSO}{%
\begin{algorithm}[htbp]
  \caption{Pseudo-code of the outer layer in AutoPSO}
  \label{alg:autopso}
  \begin{algorithmic}[1]
    \REQUIRE Population size $M$; configuration mapping function $\mathcal{F}_{\text{map}}$;
             objective function $f$; update operator $\mathcal{F}_{\text{upd}}$; configuration repository $\theta(\cdot)$
    \ENSURE  Best configuration $\theta^\star$
    \STATE $\mathbf{\theta}= \{\mathbf{\theta}_1,...,\mathbf{\theta_M}\} \gets \textsc{InitPopulation}(M)$ \\/* each individual encodes one PSO configuration */
    \WHILE{termination condition is not met}
        \STATE  $\{\text{PSO}_1,...,\text{PSO}_M\}
        \gets \mathcal{F}_{\text{map}}(\mathbf{\theta})$ \\/* decode and map to PSO instances */
        \STATE $\{{\text{fitness}}^\star_{1,...,\text{M}}\}$$ \gets \textsc{ProblemSolving}(f, \{\text{PSO}_{1,...,\text{M}}\})$
        \\/* problem solving by gPSO instances */
        \STATE $\theta' \gets
               \textsc{Optimization}(\theta, \{{\mathbf{fitness}}^\star_{1,...,\text{M}}\})$ \\/* update configuration by outer optimizer */
    \ENDWHILE
    \STATE $\theta^\star \gets
           \textsc{Best}(\theta', \{{\mathbf{fitness}}^\star_{1,...,\text{M}}\}$)
    \RETURN $\theta^\star$
  \end{algorithmic}
\end{algorithm}
}

\newcommand{\suppTabBenchmark}{
\begin{table*}[htbp]
\centering
\footnotesize
\caption{Summary of benchmarks used in this study.}
\label{supp:tab:benchmark}

\begin{tabular}{clcl}
\toprule
Type & Func. & Range & Description \\
\hline
\multirow{7}{*}{\makecell{Numerical\\function}} &
CEC2022 & [-100, 100] & benchmark suite with diverse complex functions \\
&Ackley & [-32, 32] & multimodal with many local minima \\
&Griewank & [-600, 600] & many widespread local minima \\
&Rastrigin & [-5.12, 5.12] & highly multimodal with a global minimum at origin \\
&Rosenbrock & [-5, 10] & narrow, curved valley; difficult to converge \\
&Schwefel & [-500, 500] & deceptive landscape with global optima far from origin \\
&Sphere & [-5.12, 5.12] & simple unimodal quadratic bowl-shaped function \\

\hline
Type & Env. & Dim  & Objectives \\
\hline
\multirow{5}{*}{\makecell{Robot \\control\\task}} & Halfcheetah & 1812 & fast forward locomotion \\
& Hopper & 1539 & balance and jump \\
& Pusher & 2131 & pushing object to target \\
& Reacher & 1506 & precise reaching \\
& Swimmer & 1410 & maximizing movement\\
\bottomrule                                                 
\end{tabular}

\end{table*}

}

\newcommand{\suppTabParams}{
\begin{table}[htbp]
\centering
\footnotesize
\caption{Key parameters of compared PSO variants.}
\label{supp:tab:params}
\scalebox{0.85}{
\begin{tabular}{cl}
\toprule
Algorithm & Key parameters \\
\hline
CSO       & popsize=100,phi=0  \\
CLPSO     & popsize=100,w=0.5,const coeff=1.49445, learning probability = 0.05 \\
FIPS      & popsize=100, max\_phi=4.1,topology="Square", weight\_type = "Distance" \\  
PSO       & popsize=100,w=0.6, cognitive coeff=2.5,social coeff= 0.8\\
SLPSOGS   & popsize=100, social\_coeff=0.7, demonstrator\_choice\_factor=0.4 \\
SLPSOUS   & popsize=100, social\_coeff=0.7, demonstrator\_choice\_factor=0.4 \\
\bottomrule                                                 
\end{tabular}
}
\end{table}
}

\newcommand{\suppTabExemplar}{
\begin{table}[htbp]
\centering
\caption{Exemplar candidates in AutoPSO and their corresponding implementation in code.}
\label{supp:tab:exemplar}
\scalebox{0.85}{
\begin{tabular}{ccc}
\hline
\textbf{Category} & \textbf{Exemplar} & \textbf{Implementation in Code} \\
\hline
\multirow{2}{*}{Self-based} & \texttt{current} & \texttt{population[i,:]} \\
& \texttt{pbest}   & \texttt{local\_best\_location[i,:]} \\
\hline
\multirow{3}{*}{Population-based} & \texttt{gbest} & \texttt{global\_best\_location} \\
& \texttt{center} & \texttt{mean(population)} \\
& \texttt{lbest}  & tournament between two random \texttt{pbest} \\
\hline
\multirow{4}{*}{Randomized} & \texttt{random-x} & \texttt{population[rand\_choice]} \\
& \texttt{random-pbest} & \texttt{local\_best\_location[rand\_k,:]} \\
& \texttt{random-elite} & \texttt{select\_rand\_pbest(pbest, top-p\%)} \\
& \texttt{new} & \texttt{uniform(lb, ub)} \\
\bottomrule
\end{tabular}
}
\end{table}
}

\newcommand{\suppTabCEC}{
\begin{table*}[htbp]
\centering
\caption{Comparison between AutoPSO with other PSO variants under equal Runtime.}
\label{supp:tab:CEC2022}
\scalebox{0.75}{
\begin{threeparttable}

\begin{tabular}{p{0.4cm}cccccccc}
\toprule
Dim & Func &  \textbf{AutoPSO} & \textbf{PSO}     & \textbf{CSO}     & \textbf{CLPSO}   & \textbf{FIPS}    & \textbf{SLPSOGS} & \textbf{SLPSOUS} \\ \midrule
\multirow{12}{*}{10D} 
& $F_{1}$ & \cellcolor{lightgray!30}\textbf{0.00E+00 (0.00E+00)} & \cellcolor{lightgray!30}\textbf{0.00E+00 (0.00E+00) $\approx$} & \cellcolor{lightgray!30}\textbf{0.00E+00 (0.00E+00) $\approx$} & \cellcolor{lightgray!30}\textbf{0.00E+00 (0.00E+00) $\approx$} & 1.41E+03 (4.67E+03) - & 5.63E+03 (2.81E+03) - & 2.10E+03 (1.67E+03) - \\
 & $F_{2}$ & \cellcolor{lightgray!30}\textbf{1.51E-08 (4.06E-08)} & 1.38E+01 (1.70E+01) - & 8.15E+00 (1.88E+00) - & 3.44E+00 (3.32E+00) - & 7.19E+01 (1.27E+02) - & 1.70E+01 (1.91E+01) - & 1.25E+01 (1.50E+01) - \\
 & $F_{3}$ & \cellcolor{lightgray!30}\textbf{0.00E+00 (0.00E+00)} & 9.67E-02 (2.99E-01) - & \cellcolor{lightgray!30}\textbf{0.00E+00 (0.00E+00) $\approx$} & \cellcolor{lightgray!30}\textbf{0.00E+00 (0.00E+00) $\approx$} & 1.03E+00 (1.86E+00) - & 2.79E-06 (4.23E-06) - & 7.91E-07 (1.34E-06) - \\
 & $F_{4}$ & \cellcolor{lightgray!30}\textbf{0.00E+00 (0.00E+00)} & 1.32E+01 (4.91E+00) - & 1.60E+00 (1.33E+00) - & 5.71E+00 (1.83E+00) - & 3.11E+01 (3.45E+00) - & 3.08E+00 (1.27E+00) - & 3.27E+00 (1.48E+00) - \\
 & $F_{5}$ & \cellcolor{lightgray!30}\textbf{0.00E+00 (0.00E+00)} & 9.92E-02 (2.18E-01) - & \cellcolor{lightgray!30}\textbf{0.00E+00 (0.00E+00) $\approx$} & \cellcolor{lightgray!30}\textbf{0.00E+00 (0.00E+00) $\approx$} & 2.62E-03 (6.39E-03) - & 4.75E-06 (1.26E-05) - & 1.34E-08 (5.18E-08) $\approx$ \\
 & $F_{6}$ & \cellcolor{lightgray!30}\textbf{6.47E+00 (6.33E+00)} & 4.05E+03 (2.23E+03) - & 1.44E+03 (1.36E+03) - & 1.05E+02 (4.12E+01) - & 5.39E+03 (1.77E+03) - & 1.36E+03 (1.35E+03) - & 1.43E+03 (1.51E+03) - \\
 & $F_{7}$ & \cellcolor{lightgray!30}\textbf{0.00E+00 (0.00E+00)} & 1.60E+01 (8.14E+00) - & 4.74E+00 (7.94E+00) - & 8.19E+00 (4.88E+00) - & 4.02E+01 (4.49E+01) - & 2.18E+01 (2.28E+01) - & 2.49E+01 (2.79E+01) - \\
 & $F_{8}$ & \cellcolor{lightgray!30}\textbf{8.51E-02 (7.75E-02)} & 1.67E+01 (7.98E+00) - & 2.22E+01 (6.26E+00) - & 1.77E+01 (5.57E+00) - & 2.52E+01 (2.02E+00) - & 2.12E+01 (1.32E+00) - & 2.05E+01 (2.58E+00) - \\
 & $F_{9}$ & \cellcolor{lightgray!30}\textbf{1.87E+02 (8.08E+01)} & 2.29E+02 (1.47E-05) - & 2.29E+02 (1.08E-05) - & 2.29E+02 (1.54E-05) - & 2.32E+02 (1.73E+01) - & 2.33E+02 (6.55E+00) - & 2.29E+02 (2.98E-01) - \\
 & $F_{10}$ & \cellcolor{lightgray!30}\textbf{6.04E+01 (4.04E+01)} & 1.14E+02 (3.63E+01) - & 1.24E+02 (4.42E+01) - & 1.00E+02 (4.89E-02) - & 2.44E+02 (2.09E+02) - & 1.45E+02 (4.33E+01) - & 1.50E+02 (5.28E+01) - \\
 & $F_{11}$ & \cellcolor{lightgray!30}\textbf{0.00E+00 (0.00E+00)} & 1.47E+02 (1.14E+02) - & 2.90E+01 (5.93E+01) - & 9.68E+00 (3.69E+01) $\approx$ & 1.40E+02 (1.40E+02) - & 1.45E+02 (8.91E+01) - & 1.46E+02 (1.11E+02) - \\
 & $F_{12}$ & \cellcolor{lightgray!30}\textbf{1.59E+02 (1.69E-03)} & 1.63E+02 (1.79E+00) - & 1.65E+02 (9.85E-01) - & 1.62E+02 (1.28E+00) - & 1.67E+02 (5.56E+00) - & 1.66E+02 (1.32E+00) - & 1.65E+02 (1.08E+00) - \\ \hline
\multicolumn{2}{c}{$+/ \approx/ -$} & - & 0/1/11 & 0/3/9 & 0/4/8 & 0/0/12 & 0/0/12 & 0/1/11 \\
\hline

\multirow{12}{*}{20D} 
& $F_{1}$ & \cellcolor{lightgray!30}\textbf{0.00E+00 (0.00E+00)} & \cellcolor{lightgray!30}\textbf{0.00E+00 (0.00E+00) $\approx$} & \cellcolor{lightgray!30}\textbf{0.00E+00 (0.00E+00) $\approx$} & 7.02E+00 (8.60E+00) - & 1.75E+04 (8.95E+03) - & 3.55E+04 (9.99E+03) - & 2.54E+04 (8.83E+03) - \\
 & $F_{2}$ & \cellcolor{lightgray!30}\textbf{2.36E-09 (6.30E-09)} & 7.63E+01 (3.78E+01) - & 4.90E+01 (7.43E-01) - & 4.98E+01 (3.73E+00) - & 6.87E+01 (2.41E+01) - & 5.91E+01 (1.41E+01) - & 5.35E+01 (8.91E+00) - \\
 & $F_{3}$ & \cellcolor{lightgray!30}\textbf{0.00E+00 (0.00E+00)} & 7.02E-01 (9.25E-01) - & \cellcolor{lightgray!30}\textbf{0.00E+00 (0.00E+00) $\approx$} & \cellcolor{lightgray!30}\textbf{0.00E+00 (0.00E+00) $\approx$} & 9.83E+00 (5.78E+00) - & 5.50E-03 (2.73E-02) - & 4.87E-04 (1.54E-03) - \\
 & $F_{4}$ & \cellcolor{lightgray!30}\textbf{0.00E+00 (0.00E+00)} & 3.99E+01 (1.33E+01) - & 1.51E+01 (2.44E+01) - & 2.20E+01 (4.79E+00) - & 1.21E+02 (9.49E+00) - & 8.28E+00 (2.69E+00) - & 9.02E+00 (2.82E+00) - \\
 & $F_{5}$ & \cellcolor{lightgray!30}\textbf{0.00E+00 (0.00E+00)} & 2.88E+01 (4.71E+01) - & \cellcolor{lightgray!30}\textbf{0.00E+00 (0.00E+00) $\approx$} & \cellcolor{lightgray!30}\textbf{0.00E+00 (0.00E+00) $\approx$} & 6.07E+02 (2.66E+02) - & 1.10E+00 (3.15E+00) - & 6.31E-01 (1.42E+00) - \\
 & $F_{6}$ & \cellcolor{lightgray!30}\textbf{4.57E+00 (5.27E+00)} & 5.66E+04 (2.51E+05) - & 1.47E+03 (1.84E+03) - & 3.23E+02 (6.46E+02) - & 1.63E+07 (1.60E+07) - & 2.19E+03 (2.86E+03) - & 1.63E+03 (1.73E+03) - \\
 & $F_{7}$ & \cellcolor{lightgray!30}\textbf{0.00E+00 (0.00E+00)} & 3.16E+01 (8.16E+00) - & 2.35E+01 (8.35E+00) - & 4.94E+01 (8.52E+00) - & 1.05E+02 (4.64E+01) - & 4.48E+01 (1.03E+01) - & 3.62E+01 (7.69E+00) - \\
 & $F_{8}$ & \cellcolor{lightgray!30}\textbf{6.42E-02 (6.20E-02)} & 2.37E+01 (3.79E+00) - & 2.82E+01 (1.95E+00) - & 2.75E+01 (1.54E+00) - & 4.43E+01 (1.60E+01) - & 3.32E+01 (3.23E+01) - & 2.92E+01 (2.36E+01) - \\
 & $F_{9}$ & \cellcolor{lightgray!30}\textbf{1.72E+02 (9.16E+01)} & 1.91E+02 (1.64E+01) - & 1.81E+02 (4.52E-02) - & 1.81E+02 (4.91E-04) - & 1.94E+02 (2.66E+01) - & 1.86E+02 (1.69E+00) - & 1.83E+02 (1.81E+00) - \\
 & $F_{10}$ & \cellcolor{lightgray!30}\textbf{5.38E+01 (4.13E+01)} & 2.02E+02 (1.69E+02) - & 1.16E+02 (4.90E+01) - & 1.00E+02 (5.07E-02) - & 2.32E+03 (6.18E+02) - & 5.49E+02 (4.59E+02) - & 2.38E+02 (2.06E+02) - \\
 & $F_{11}$ & \cellcolor{lightgray!30}\textbf{0.00E+00 (0.00E+00)} & 8.23E+02 (3.44E+02) - & 3.10E+02 (2.96E+01) - & 3.00E+02 (6.09E-04) - & 2.30E+03 (9.95E+02) - & 3.05E+02 (2.93E+01) - & 3.09E+02 (4.80E+01) - \\
 & $F_{12}$ & \cellcolor{lightgray!30}\textbf{1.59E+02 (5.93E-04)} & 2.65E+02 (1.96E+01) - & 2.42E+02 (5.73E+00) - & 2.38E+02 (2.96E+00) - & 2.88E+02 (3.53E+01) - & 2.59E+02 (7.85E+00) - & 2.54E+02 (1.03E+01) - \\ \hline
\multicolumn{2}{c}{$+/ \approx/ -$} & - & 0/1/11 & 0/3/9 & 0/2/10 & 0/0/12 & 0/0/12 & 0/0/12 \\
\bottomrule
\end{tabular}

\begin{tablenotes}[flushleft]
\footnotesize
\item[*] Wilcoxon rank-sum tests (significance level $\alpha=0.05$) were conducted by comparing AutoPSO against each algorithm individually.
The final row quantifies the tasks where the corresponding algorithm performs statistically better ($+$), similarly ($\approx$), or worse ($-$) relative to AutoPSO.
\end{tablenotes}
\end{threeparttable}
}
\end{table*}
}

\newcommand{\suppFigBrax}{
\begin{figure}[H]
    \centering
    \begin{minipage}[b]{0.24\textwidth}
        \centering
        \includegraphics[width=\textwidth]{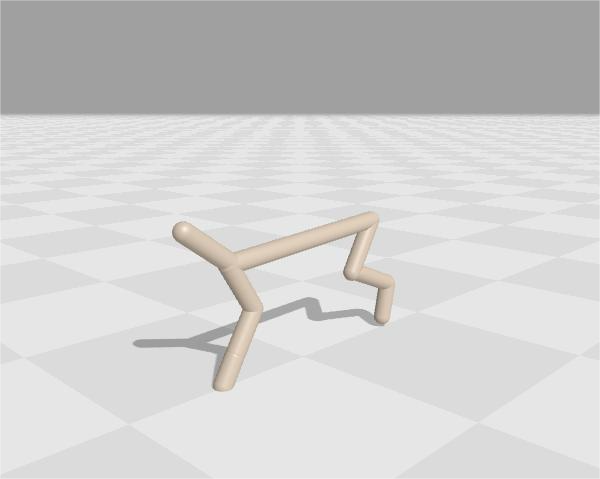}  
        \centering
        \subcaption{Halfcheetah}
    \end{minipage}
    \begin{minipage}[b]{0.24\textwidth}
        \centering
        \includegraphics[width=\textwidth]{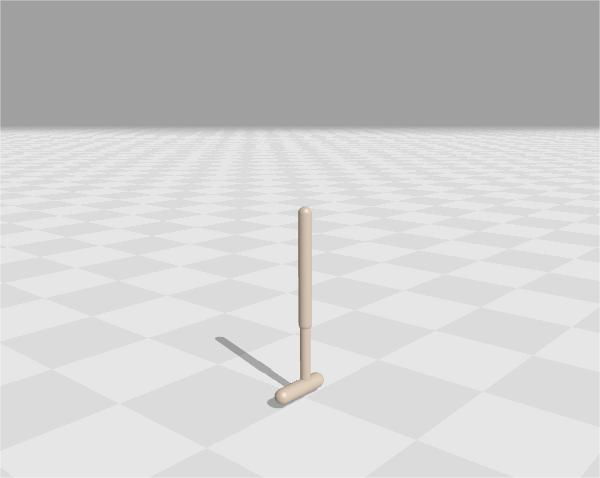}
        \centering
        \subcaption{Hopper}
    \end{minipage}

  \begin{minipage}[b]{0.24\textwidth}
        \centering
        \includegraphics[width=\textwidth]{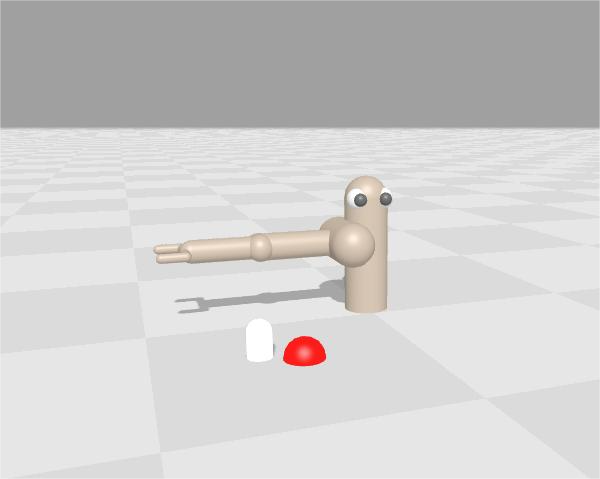} 
        \centering     
        \subcaption{Pusher}
    \end{minipage}
    \begin{minipage}[b]{0.24\textwidth}
        \centering
        \includegraphics[width=\textwidth]{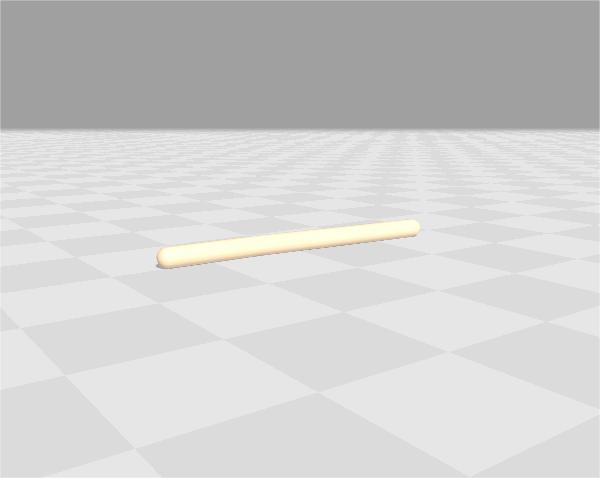} 
        \centering     
        \subcaption{Swimmer}
    \end{minipage}

    \caption{Four representative robotic control tasks based on Brax engine.}
    \label{supp:fig:brax-environments}
\end{figure}

}

\newcommand{\suppFigCECTenD}{
\begin{figure*}[htbp]
    \centering
    \includegraphics[width=\textwidth]{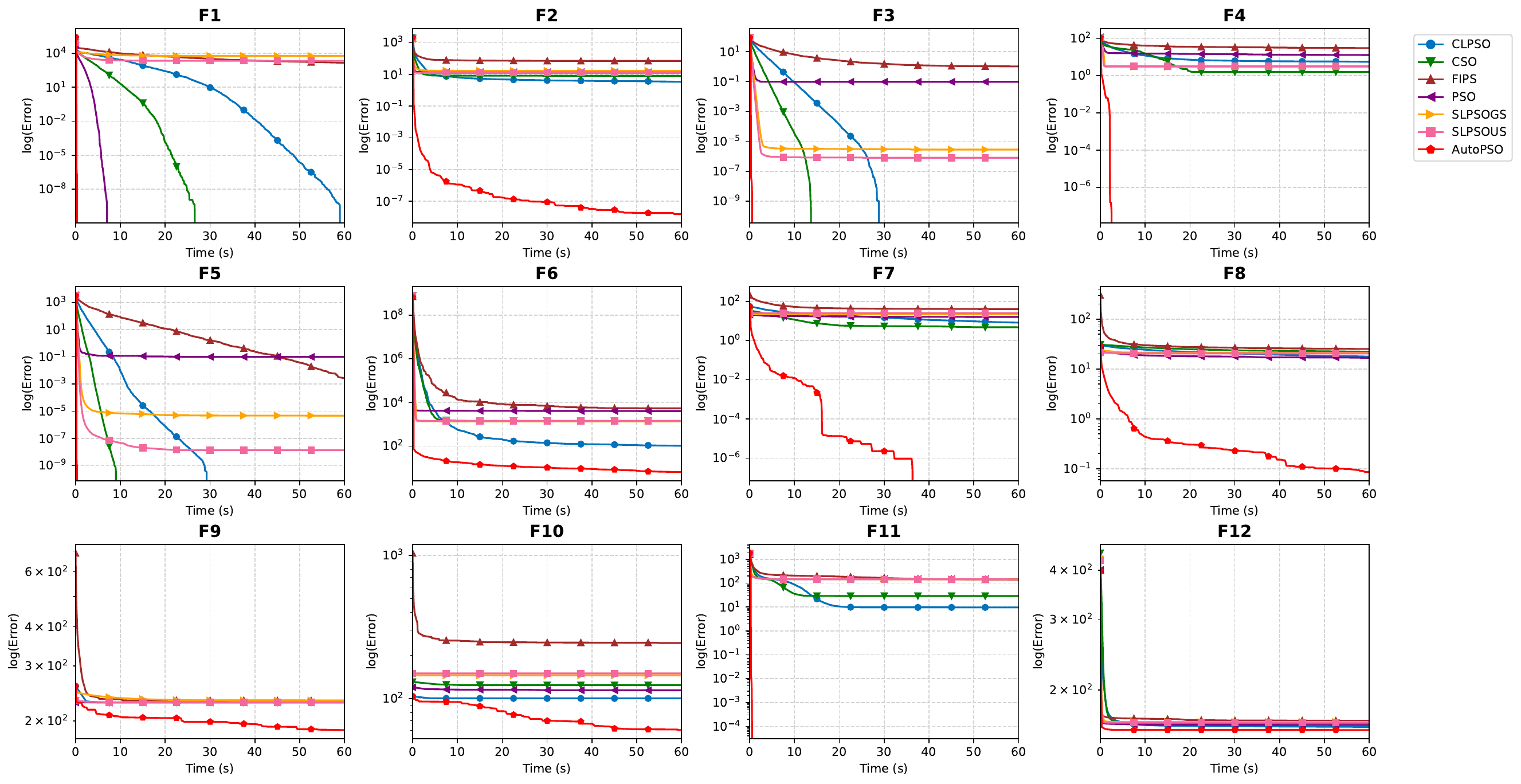}
    \caption{Convergence curves on 10D problems in CEC2022 benchmark suite.}
    \label{supp:fig:cec022-10d}
\end{figure*}
}
\newcommand{\suppFigCECTewD}{
\begin{figure*}[htbp]
    \centering
    \includegraphics[width=\textwidth]{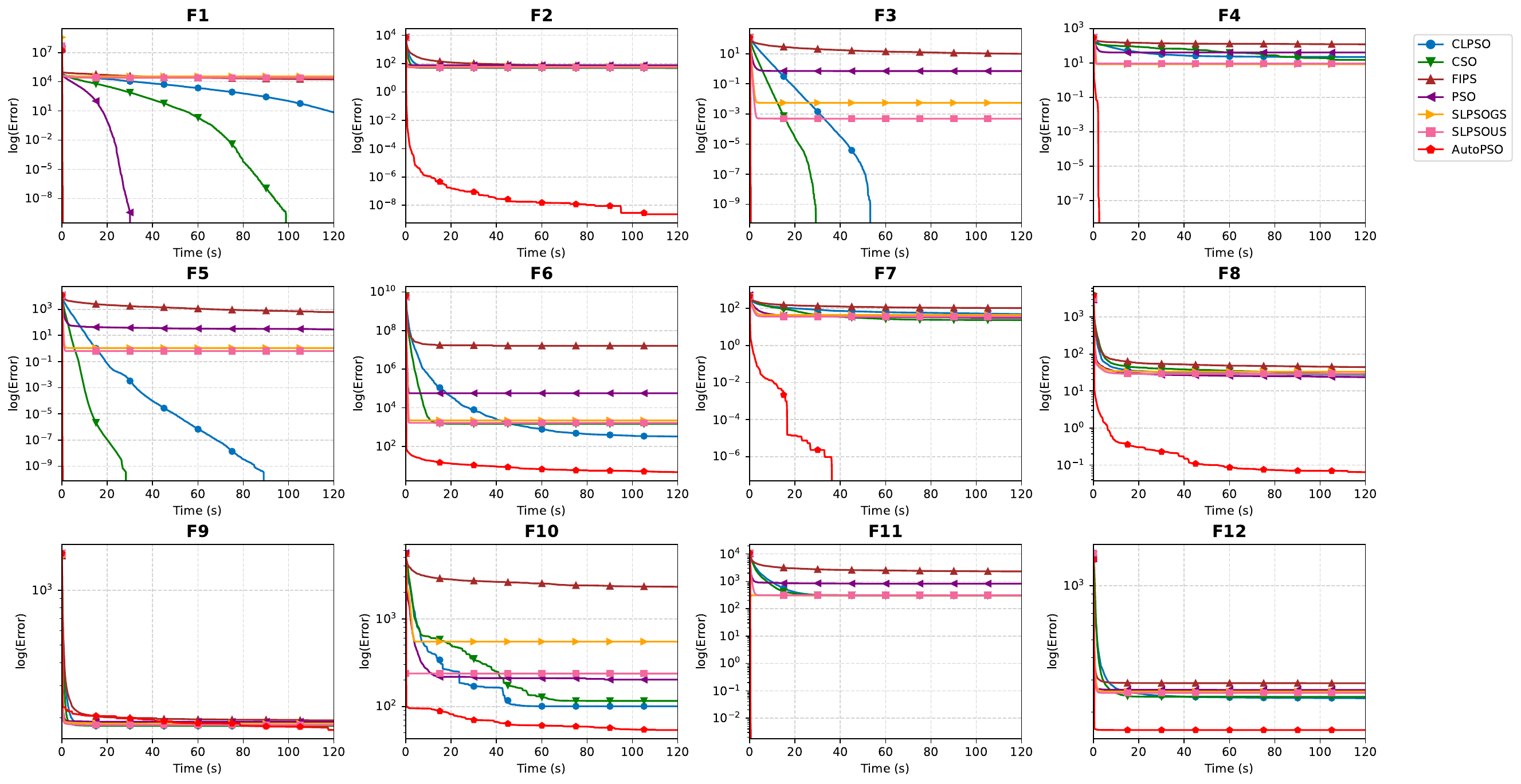}
    \caption{Convergence curves on 20D problems in CEC2022 benchmark suite.}
    \label{supp:fig:cec022-20d}
\end{figure*}
}

\newcommand{\suppTabCECFE}{
\begin{table*}[htbp]
\centering
\caption{Comparison between AutoPSO with other PSO variants under equal FEs.}
\label{supp:tab:CEC-FE}
\scalebox{0.75}{
\begin{threeparttable}

\begin{tabular}{p{0.4cm}cccccccc}
\toprule
Dim & Func &  \textbf{AutoPSO} & \textbf{PSO}     & \textbf{CSO}     & \textbf{CLPSO}   & \textbf{FIPS}    & \textbf{SLPSOGS} & \textbf{SLPSOUS} \\ \midrule

\multirow{12}{*}{10D}
& $F_1$  & \cellcolor{lightgray!30}\textbf{0.00E+00 (0.00E+00)} &\cellcolor{lightgray!30}\textbf{0.00E+00 (0.00E+00)}$\approx$ & \cellcolor{lightgray!30}\textbf{0.00E+00 (0.00E+00)}$\approx$ & \cellcolor{lightgray!30}\textbf{0.00E+00 (0.00E+00)}$\approx$ & 2.57E+03 (7.66E+03)$\approx$ & 7.79E+03 (2.87E+03)$-$ & 1.17E+03 (1.16E+03)$-$ \\
& $F_2$  & \cellcolor{lightgray!30}\textbf{2.36E-09 (6.30E-09)} & 9.39E+00 (3.21E+00)$-$ & 6.76E+00 (3.03E+00)$-$ & 1.47E+00 (2.96E+00)$\approx$ & 6.01E+01 (9.89E+01)$-$ & 1.51E+01 (1.58E+01)$-$ & 8.57E+00 (6.36E-01)$-$ \\
& $F_3$  & \cellcolor{lightgray!30}\textbf{0.00E+00 (0.00E+00)} & 1.82E-01 (3.94E-01)$\approx$ & \cellcolor{lightgray!30}\textbf{0.00E+00 (0.00E+00)}$\approx$ & \cellcolor{lightgray!30}\textbf{0.00E+00 (0.00E+00)}$\approx$ & 1.34E+00 (2.13E+00)$-$ & 1.10E-06 (1.42E-06)$-$ & 4.41E-07 (6.28E-07)$-$ \\
& $F_4$  & \cellcolor{lightgray!30}\textbf{0.00E+00 (0.00E+00)} & 6.15E+00 (1.89E+00)$-$ & 1.72E+00 (7.46E-01)$-$ & 1.63E+00 (8.77E-01)$-$ & 9.30E+00 (1.76E+00)$-$ & 3.26E+00 (1.48E+00)$-$ & 3.53E+00 (1.23E+00)$-$ \\
& $F_5$  & \cellcolor{lightgray!30}\textbf{0.00E+00 (0.00E+00)} & 1.31E-01 (3.05E-01)$-$ & \cellcolor{lightgray!30}\textbf{0.00E+00 (0.00E+00)}$\approx$ & \cellcolor{lightgray!30}\textbf{0.00E+00 (0.00E+00)}$\approx$ & \cellcolor{lightgray!30}\textbf{0.00E+00 (0.00E+00)}$\approx$ & 8.85E-02 (2.80E-01)$-$ & 1.05E-08 (2.28E-08)$\approx$ \\
& $F_6$  & \cellcolor{lightgray!30}\textbf{4.29E+00 (4.74E+00)} & 3.50E+03 (2.23E+03)$-$ & 6.14E+02 (7.40E+02)$-$ & 3.71E+01 (1.88E+01)$-$ & 2.48E+03 (2.56E+03)$-$ & 1.06E+03 (7.95E+02)$-$ & 1.86E+03 (1.31E+03)$-$ \\
& $F_7$  & \cellcolor{lightgray!30}\textbf{0.00E+00 (0.00E+00)} & 1.32E+01 (9.50E+00)$-$ & 2.47E+00 (5.59E+00)$-$ & 8.13E-03 (1.33E-02)$\approx$ & 3.65E+01 (4.80E+01)$-$ & 1.84E+01 (1.01E+01)$-$ & 1.87E+01 (8.25E+00)$-$ \\
& $F_8$  & \cellcolor{lightgray!30}\textbf{6.42E-02 (6.20E-02)} & 1.33E+01 (9.41E+00)$-$ & 1.54E+01 (8.85E+00)$-$ & 1.21E-01 (6.40E-02)$-$ & 1.99E+01 (4.35E+00)$-$ & 2.09E+01 (1.55E+00)$-$ & 2.11E+01 (6.03E-01)$-$ \\
& $F_9$  & \cellcolor{lightgray!30}\textbf{1.72E+02 (9.14E+01)} & 2.29E+02 (1.87E-05)$-$ & 2.29E+02 (1.13E-05)$\approx$ & 2.29E+02 (4.39E-06)$-$ & 2.29E+02 (9.41E-06)$-$ & 2.32E+02 (5.84E+00)$-$ & 2.29E+02 (1.79E-01)$-$ \\
& $F_{10}$ & \cellcolor{lightgray!30}\textbf{5.38E+01 (4.13E+01)} & 1.20E+02 (4.27E+01)$-$ & 1.20E+02 (4.10E+01)$-$ & 1.00E+02 (3.33E-02)$\approx$ & 2.07E+02 (2.23E+02)$-$ & 1.42E+02 (4.19E+01)$-$ & 1.20E+02 (3.76E+01)$-$ \\
& $F_{11}$ & \cellcolor{lightgray!30}\textbf{0.00E+00 (0.00E+00)} & 1.38E+02 (1.56E+02)$-$ & 5.45E+01 (7.21E+01)$\approx$ & 2.73E+01 (5.78E+01)$\approx$ & 1.19E+02 (9.72E+01)$-$ & 1.60E+02 (7.84E+01)$-$ & 1.07E+02 (1.16E+02)$-$ \\
& $F_{12}$ & \cellcolor{lightgray!30}\textbf{1.59E+02 (5.95E-04)} & 1.64E+02 (1.85E+00)$-$ & 1.65E+02 (1.26E+00)$-$ & 1.61E+02 (1.28E+00)$-$ & 1.71E+02 (7.44E+00)$-$ & 1.66E+02 (7.66E-01)$-$ & 1.65E+02 (6.97E-01)$-$ \\

\midrule
\multicolumn{2}{c}{$+/ \approx/ -$} & - & 0/2/10 & 0/5/7 & 0/7/5 & 0/2/10 & 0/0/12 & 0/1/11 \\

\midrule
\multirow{12}{*}{20D}
& $F_1$  & \cellcolor{lightgray!30}\textbf{0.00E+00 (0.00E+00)} & \cellcolor{lightgray!30}\textbf{0.00E+00 (0.00E+00)}$\approx$ & \cellcolor{lightgray!30}\textbf{0.00E+00 (0.00E+00)}$\approx$ & \cellcolor{lightgray!30}\textbf{0.00E+00 (0.00E+00)}$\approx$ & 1.08E+03 (2.59E+03)$-$ & 3.70E+04 (1.17E+04)$-$ & 2.53E+04 (8.63E+03)$-$ \\
& $F_2$  & \cellcolor{lightgray!30}\textbf{2.36E-09 (6.30E-09)} & 7.76E+01 (2.87E+01)$-$ & 4.91E+01 (6.95E-05)$-$ & 4.23E+01 (1.52E+01)$-$ & 5.63E+01 (1.61E+01)$-$ & 6.06E+01 (1.16E+01)$-$ & 4.94E+01 (3.84E-01)$-$ \\
& $F_3$  & \cellcolor{lightgray!30}\textbf{0.00E+00 (0.00E+00)} & 3.97E-01 (3.82E-01)$-$ & \cellcolor{lightgray!30}\textbf{0.00E+00 (0.00E+00)}$\approx$ & \cellcolor{lightgray!30}\textbf{0.00E+00 (0.00E+00)}$\approx$ & 3.12E+00 (2.44E+00)$-$ & 1.28E-02 (2.69E-02)$-$ & 6.09E-03 (1.92E-02)$-$ \\
& $F_4$  & \cellcolor{lightgray!30}\textbf{0.00E+00 (0.00E+00)} & 3.58E+01 (1.03E+01)$-$ & 5.70E+00 (1.21E+00)$-$ & 1.55E+01 (3.39E+00)$-$ & 7.54E+01 (5.75E+00)$-$ & 8.05E+00 (3.59E+00)$-$ & 8.59E+00 (1.91E+00)$-$ \\
& $F_5$  & \cellcolor{lightgray!30}\textbf{0.00E+00 (0.00E+00)} & 1.83E+01 (3.51E+01)$-$ & \cellcolor{lightgray!30}\textbf{0.00E+00 (0.00E+00)}$\approx$ & 8.14E-03 (2.57E-02)$\approx$ & 6.31E+01 (1.08E+02)$-$ & 1.79E+00 (3.70E+00)$-$ & 2.27E-01 (6.62E-01)$-$ \\
& $F_6$  & \cellcolor{lightgray!30}\textbf{4.57E+00 (5.27E+00)} & 7.10E+03 (7.99E+03)$-$ & 1.43E+03 (1.97E+03)$-$ & 1.85E+02 (1.32E+02)$-$ & 1.15E+07 (1.70E+07)$-$ & 8.56E+02 (8.33E+02)$-$ & 1.01E+03 (9.57E+02)$-$ \\
& $F_7$  & \cellcolor{lightgray!30}\textbf{0.00E+00 (0.00E+00)} & 2.21E+01 (2.53E+00)$-$ & 2.43E+01 (8.26E+00)$-$ & 1.58E+01 (8.05E+00)$-$ & 1.12E+02 (4.56E+01)$-$ & 4.57E+01 (1.35E+01)$-$ & 3.65E+01 (8.05E+00)$-$ \\
& $F_8$  & \cellcolor{lightgray!30}\textbf{6.42E-02 (6.20E-02)} & 2.12E+01 (4.40E-01)$-$ & 2.08E+01 (2.86E-01)$-$ & 2.11E+01 (4.40E-01)$-$ & 2.80E+01 (9.73E+00)$-$ & 3.98E+01 (3.68E+01)$-$ & 3.41E+01 (3.88E+01)$-$ \\
& $F_9$  & \cellcolor{lightgray!30}\textbf{1.72E+02 (9.16E+01)} & 1.91E+02 (1.61E+01)$-$ & 1.81E+02 (2.86E-02)$-$ & 1.81E+02 (8.83E-05)$-$ & 1.83E+02 (8.50E+00)$-$ & 1.86E+02 (1.51E+00)$-$ & 1.83E+02 (1.26E+00)$-$ \\
& $F_{10}$ & \cellcolor{lightgray!30}\textbf{5.38E+01 (4.13E+01)} & 1.74E+02 (1.52E+02)$-$ & 1.10E+02 (3.25E+01)$\approx$ & 1.00E+02 (3.33E-02)$-$ & 1.25E+03 (6.06E+02)$-$ & 4.54E+02 (4.03E+02)$-$ & 2.96E+02 (2.43E+02)$-$ \\
& $F_{11}$ & \cellcolor{lightgray!30}\textbf{0.00E+00 (0.00E+00)} & 7.63E+02 (3.60E+02)$-$ & 3.00E+02 (0.00E+00)$-$ & 2.18E+02 (1.34E+02)$-$ & 2.31E+03 (6.84E+02)$-$ & 3.16E+02 (5.12E+01)$-$ & 3.00E+02 (4.00E-05)$-$ \\
& $F_{12}$ & \cellcolor{lightgray!30}\textbf{1.59E+02 (5.93E-04)} & 2.61E+02 (1.58E+01)$-$ & 2.42E+02 (4.25E+00)$-$ & 2.37E+02 (2.78E+00)$-$ & 2.82E+02 (3.46E+01)$-$ & 2.58E+02 (7.12E+00)$-$ & 2.59E+02 (8.17E+00)$-$ \\

\midrule
\multicolumn{2}{c}{$+/ \approx/ -$} & - & 0/1/11 & 0/4/8 & 0/3/9 & 0/0/12 & 0/0/12 & 0/0/12 \\

\bottomrule
\end{tabular}

\begin{tablenotes}[flushleft]
\footnotesize
\item[*] The Wilcoxon rank-sum tests (significance level $\alpha=0.05$) were conducted between AutoPSO and each algorithm individually.
The final row displays the number of tasks where the corresponding algorithm performs statistically better ($+$), similar ($\approx$), or worse ($-$) compared to AutoPSO.
\end{tablenotes}

\end{threeparttable}
}
\end{table*}

}

\newcommand{\suppFigAckleyAll}{
\begin{figure*}[htbp]
  \centering

\begin{subfigure}[t]{\textwidth}
    \centering
    \begin{minipage}[t]{0.24\textwidth}\centering
      \includegraphics[width=\linewidth]{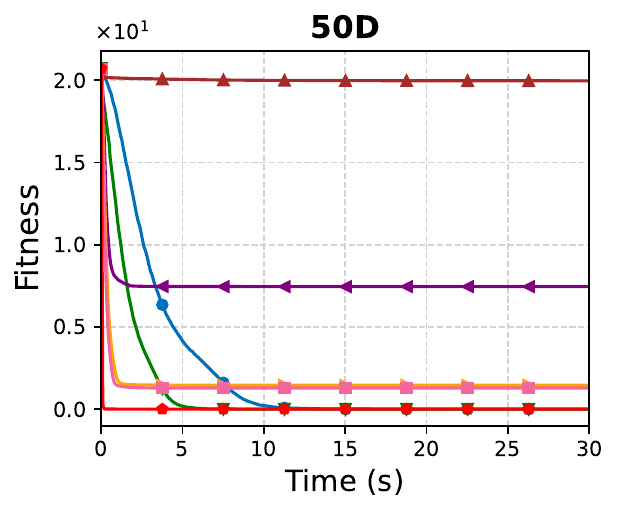}
    \end{minipage}\hspace{0.13cm}
    \begin{minipage}[t]{0.24\textwidth}\centering
      \includegraphics[width=\linewidth]{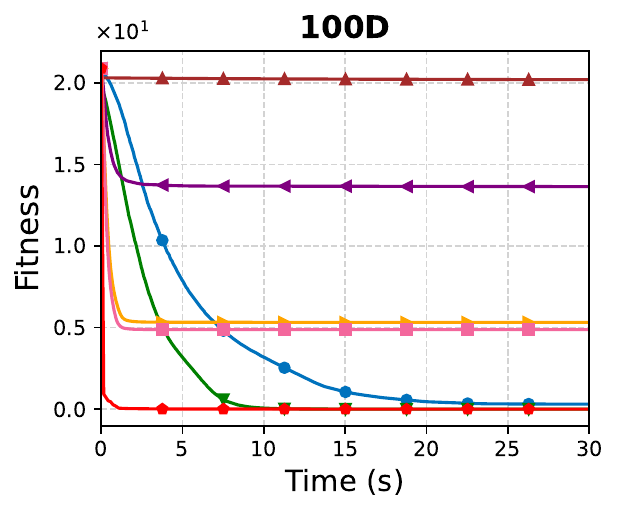}
    \end{minipage}\hspace{0.13cm}
    \begin{minipage}[t]{0.24\textwidth}\centering
      \includegraphics[width=\linewidth]{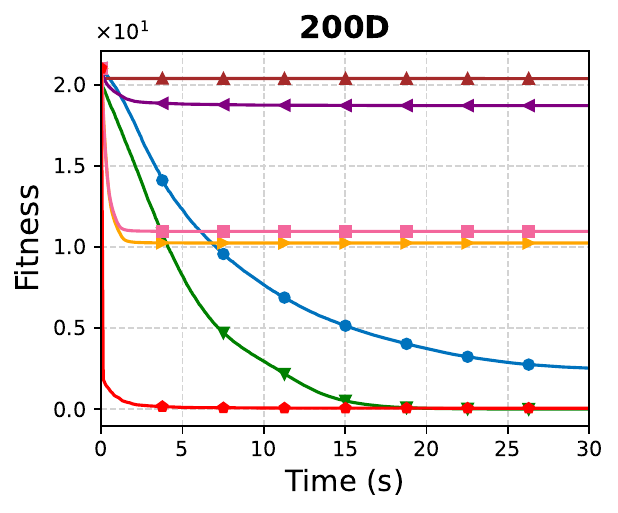}
    \end{minipage}\hspace{-0.6cm}
    \begin{minipage}[t]{0.24\textwidth}\centering
      \includegraphics[width=\linewidth]{fig/data/numerical/basic/label.pdf}
    \end{minipage}
    \vspace{-0.2cm}
    \caption{Ackley function}
    \label{supp:fig:ackley}
  \end{subfigure}

\vspace{-0.2cm}

\begin{subfigure}[t]{\textwidth}
    \centering
    \begin{minipage}[t]{0.24\textwidth}\centering
      \includegraphics[width=\linewidth]{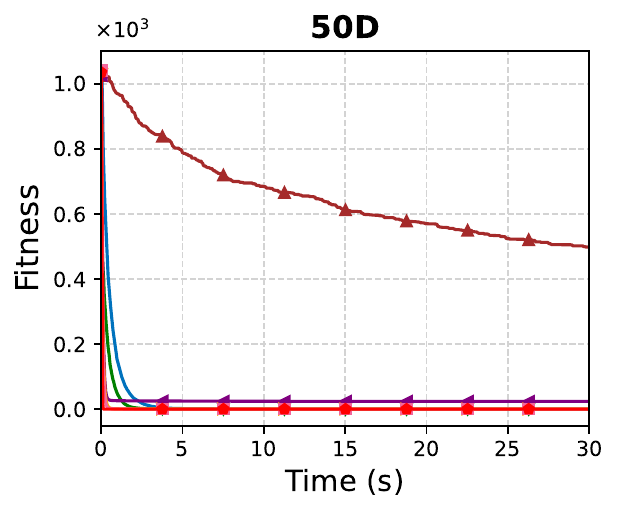}
    \end{minipage}\hspace{0.13cm}
    \begin{minipage}[t]{0.24\textwidth}\centering
      \includegraphics[width=\linewidth]{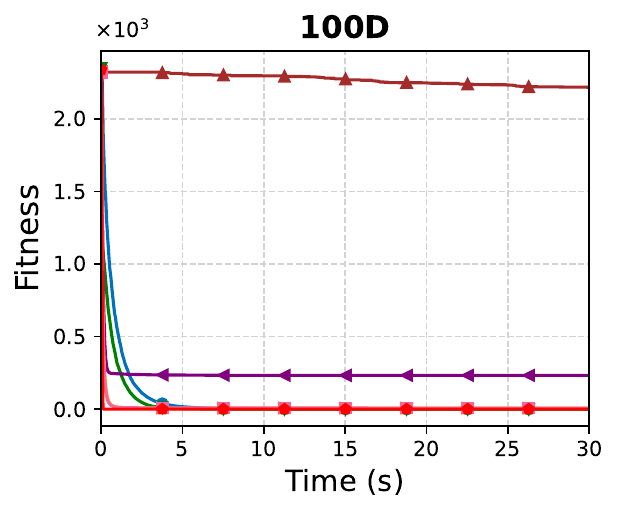}
    \end{minipage}\hspace{0.13cm}
    \begin{minipage}[t]{0.24\textwidth}\centering
      \includegraphics[width=\linewidth]{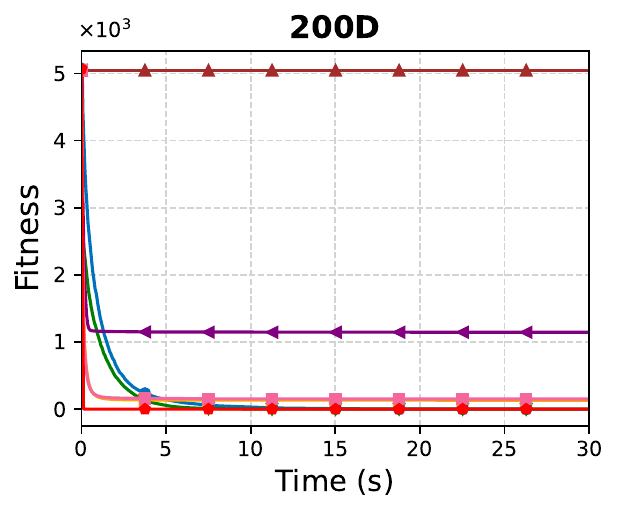}
    \end{minipage}\hspace{-0.6cm}
    \begin{minipage}[t]{0.24\textwidth}\centering
      \includegraphics[width=\linewidth]{fig/data/numerical/basic/label.pdf}
    \end{minipage}
    \vspace{-0.2cm}
    \caption{Griewank function}
    \label{supp:fig:griewank}
  \end{subfigure}

\vspace{-0.2cm}

\begin{subfigure}[t]{\textwidth}
    \centering
    \begin{minipage}[t]{0.24\textwidth}\centering
      \includegraphics[width=\linewidth]{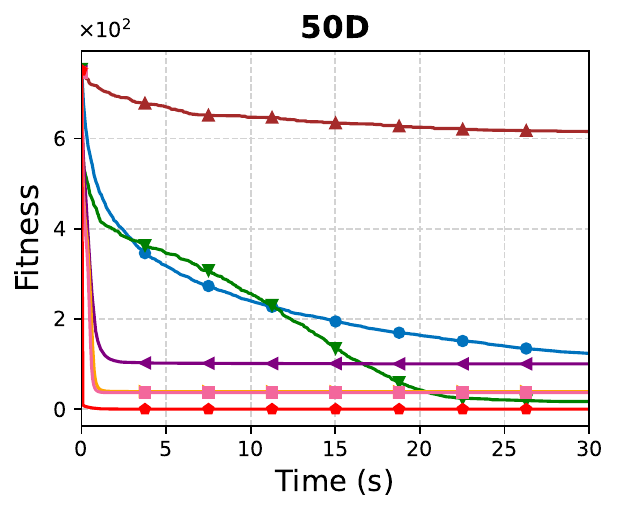}
    \end{minipage}\hspace{0.13cm}
    \begin{minipage}[t]{0.24\textwidth}\centering
      \includegraphics[width=\linewidth]{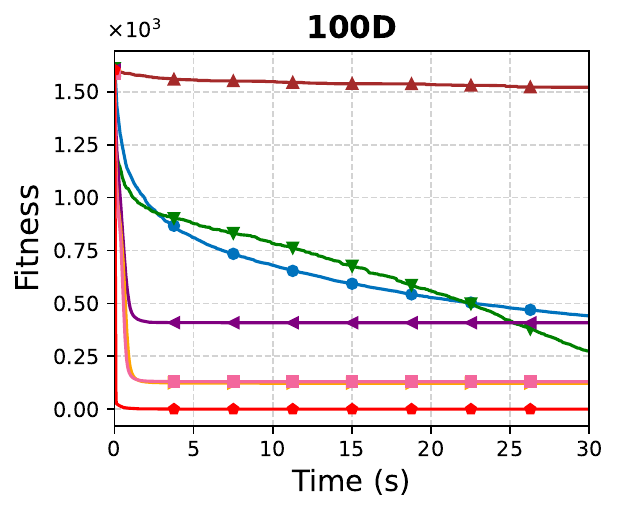}
    \end{minipage}\hspace{0.13cm}
    \begin{minipage}[t]{0.24\textwidth}\centering
      \includegraphics[width=\linewidth]{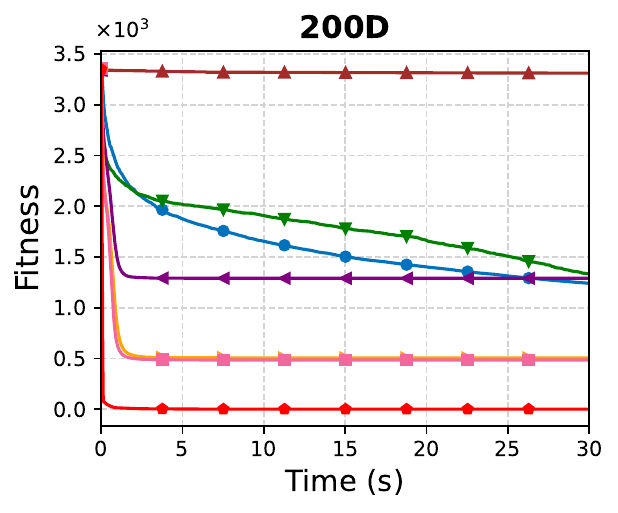}
    \end{minipage}\hspace{-0.6cm}
    \begin{minipage}[t]{0.24\textwidth}\centering
      \includegraphics[width=\linewidth]{fig/data/numerical/basic/label.pdf}
    \end{minipage}
    \vspace{-0.2cm}
    \caption{Rastrigin function}
    \label{supp:fig:rastrigin}
  \end{subfigure}

\vspace{-0.2cm}

\begin{subfigure}[t]{\textwidth}
    \centering
    \begin{minipage}[t]{0.24\textwidth}\centering
      \includegraphics[width=\linewidth]{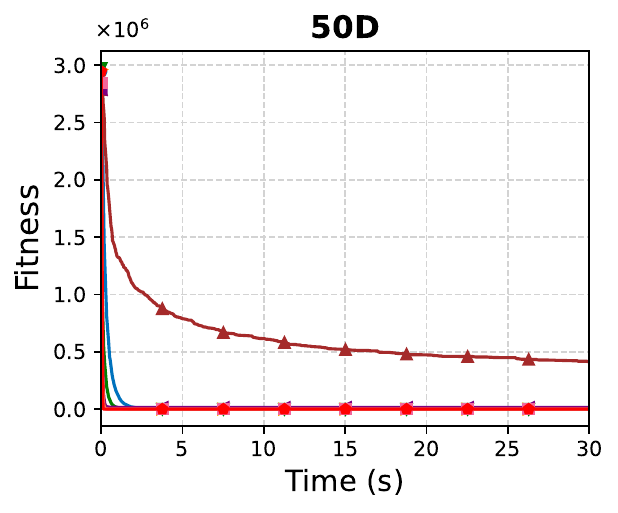}
    \end{minipage}\hspace{0.13cm}
    \begin{minipage}[t]{0.24\textwidth}\centering
      \includegraphics[width=\linewidth]{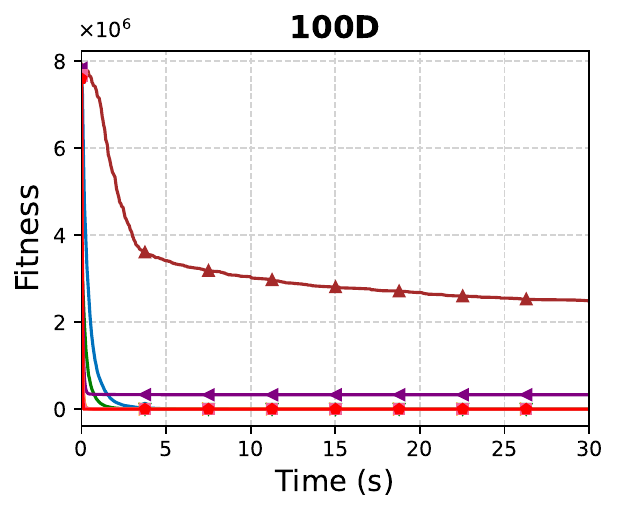}
    \end{minipage}\hspace{0.13cm}
    \begin{minipage}[t]{0.24\textwidth}\centering
      \includegraphics[width=\linewidth]{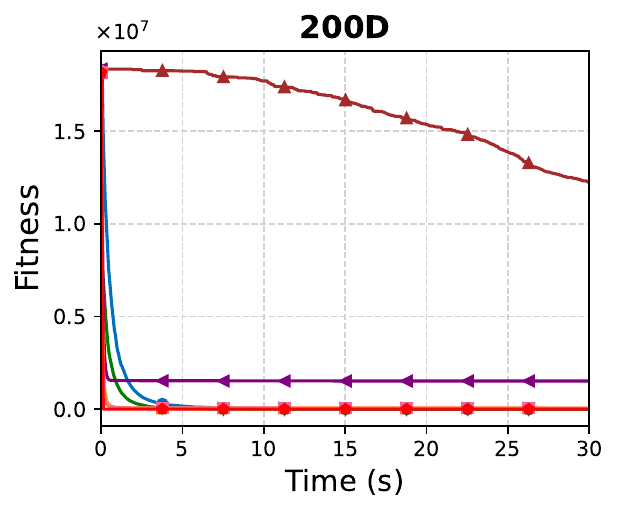}
    \end{minipage}\hspace{-0.6cm}
    \begin{minipage}[t]{0.24\textwidth}\centering
      \includegraphics[width=\linewidth]{fig/data/numerical/basic/label.pdf}
    \end{minipage}
    \vspace{-0.2cm}
    \caption{Rosenbrock function}
    \label{supp:fig:rosenbrock}
  \end{subfigure}

\vspace{-0.2cm}

\begin{subfigure}[t]{\textwidth}
    \centering
    \begin{minipage}[t]{0.24\textwidth}\centering
      \includegraphics[width=\linewidth]{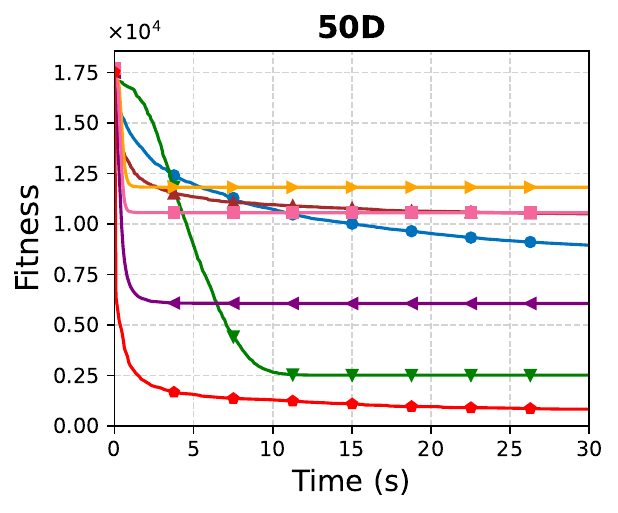}
    \end{minipage}\hspace{0.13cm}
    \begin{minipage}[t]{0.24\textwidth}\centering
      \includegraphics[width=\linewidth]{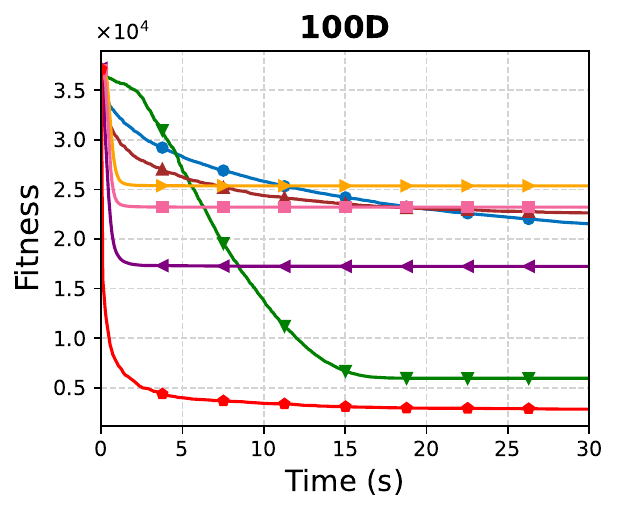}
    \end{minipage}\hspace{0.13cm}
    \begin{minipage}[t]{0.24\textwidth}\centering
      \includegraphics[width=\linewidth]{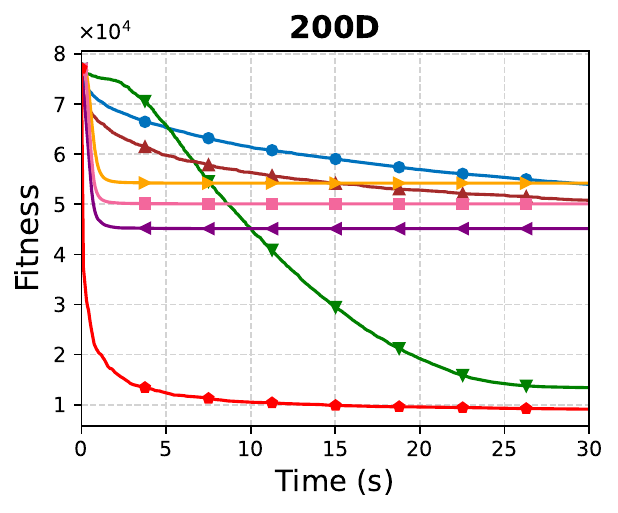}
    \end{minipage}\hspace{-0.6cm}
    \begin{minipage}[t]{0.24\textwidth}\centering
      \includegraphics[width=\linewidth]{fig/data/numerical/basic/label.pdf}
    \end{minipage}
    \vspace{-0.2cm}
    \caption{Schwefel function}
    \label{supp:fig:schwefel}
  \end{subfigure}

\caption{Convergence curves on six classic numerical optimization functions.}
  \label{supp:fig:classic-functions-part-one}
\end{figure*}

}
\newcommand{\suppFigAckleyAllT}{
\begin{figure*}[htbp]
  \centering

\begin{subfigure}[t]{\textwidth}
    \centering
    \begin{minipage}[t]{0.24\textwidth}\centering
      \includegraphics[width=\linewidth]{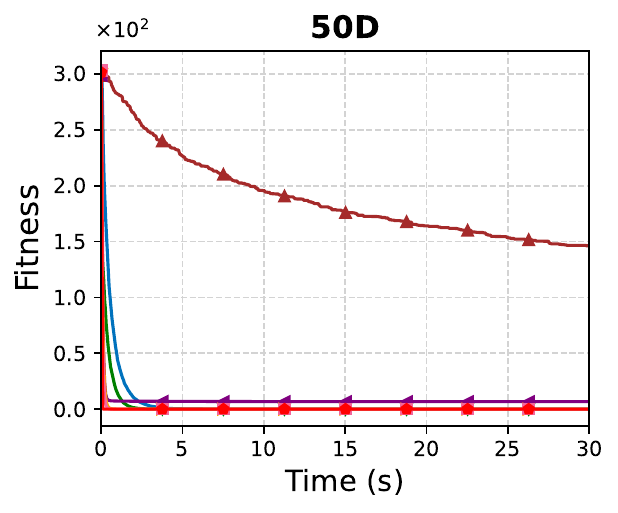}
    \end{minipage}\hspace{0.13cm}
    \begin{minipage}[t]{0.24\textwidth}\centering
      \includegraphics[width=\linewidth]{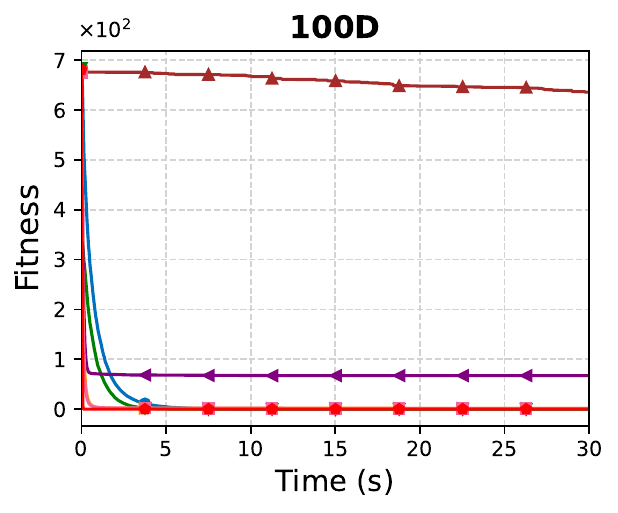}
    \end{minipage}\hspace{0.13cm}
    \begin{minipage}[t]{0.24\textwidth}\centering
      \includegraphics[width=\linewidth]{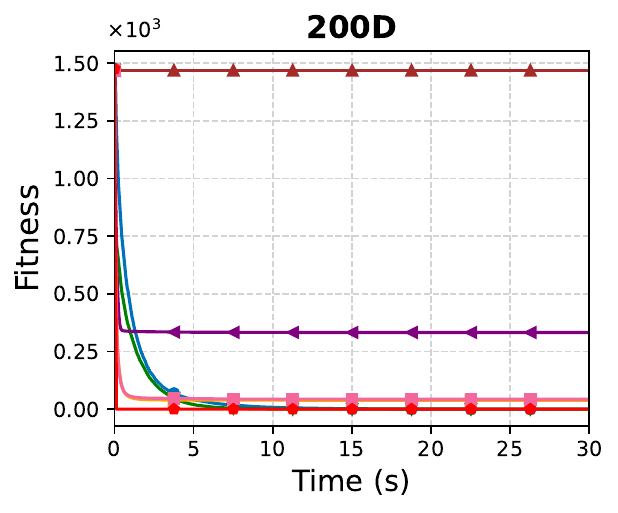}
    \end{minipage}\hspace{-0.6cm}
    \begin{minipage}[t]{0.24\textwidth}\centering
      \includegraphics[width=\linewidth]{fig/data/numerical/basic/label.pdf}
    \end{minipage}
    \vspace{-0.2cm}
    \caption{Sphere function}
    \label{supp:fig:sphere}
  \end{subfigure}

\caption{(Continued) Convergence curves on six classic numerical optimization functions.}
  \label{supp:fig:classic-functions-part-two}
\end{figure*}

}

\newcommand{\suppTabBasic}{
\begin{table*}[htbp]
\caption{Results on six classical numerical optimization functions with three dimensions.}
\centering
\label{supp:tab:basic}
\scalebox{0.74}{
\begin{threeparttable}

\begin{tabular}{ccccccccc}
\toprule
Func.                        & Dim  & \textbf{AutoPSO} & \textbf{PSO}     & \textbf{CSO}     & \textbf{CLPSO}   & \textbf{FIPS}    & \textbf{SLPSOGS} & \textbf{SLPSOUS} \\ \midrule
\multirow{3}{*}{Ackley} & 50D & \cellcolor{lightgray!30}\textbf{0.00E+00 (0.00E+00)} & 7.49E+00 (3.50E+00) - & 2.52E-01 (1.12E-02) - & 5.52E-01 (2.41E-02) - & 2.00E+01 (7.32E-03) - & 1.51E+00 (4.93E-01) - & 1.33E+00 (6.79E-01) - \\
 & 100D & \cellcolor{lightgray!30}\textbf{3.94E-06 (6.74E-07)} & 1.37E+01 (2.96E+00) - & 4.50E-01 (1.81E-02) - & 1.17E+00 (4.20E-01) - & 2.02E+01 (7.81E-02) - & 5.38E+00 (9.13E-01) - & 4.92E+00 (7.16E-01) - \\
 & 200D & \cellcolor{lightgray!30}\textbf{4.98E-02 (7.16E-02)} & 1.87E+01 (1.37E+00) - & 8.67E-01 (2.76E-02) - & 3.55E+00 (3.17E-01) - & 2.04E+01 (4.51E-02) - & 1.03E+01 (6.75E-01) - & 1.10E+01 (6.15E-01) - \\ \hline
 
\multirow{3}{*}{Griewank} & 50D & \cellcolor{lightgray!30}\textbf{0.00E+00 (0.00E+00)} & 2.49E+01 (1.63E+01) - & 1.98E+00 (1.03E-01) - & 4.61E+00 (2.84E-01) - & 4.62E+02 (1.94E+01) - & 1.18E+00 (3.13E-01) - & 7.38E-01 (1.34E-01) - \\
 & 100D & \cellcolor{lightgray!30}\textbf{0.00E+00 (0.00E+00)} & 2.35E+02 (6.80E+01) - & 7.97E+00 (4.56E-01) - & 1.43E+01 (6.57E-01) - & 2.11E+03 (5.38E+01) - & 8.53E+00 (2.15E+00) - & 8.83E+00 (3.68E+00) - \\
 & 200D & \cellcolor{lightgray!30}\textbf{5.96E-08 (4.79E-08)} & 1.15E+03 (1.43E+02) - & 2.69E+01 (9.72E-01) - & 4.30E+01 (1.71E+00) - & 5.05E+03 (1.71E+02) - & 1.38E+02 (2.38E+01) - & 1.56E+02 (4.23E+01) - \\ \hline
 
\multirow{3}{*}{Rastrigin} & 50D & \cellcolor{lightgray!30}\textbf{5.98E-03 (6.68E-03)} & 1.03E+02 (2.42E+01) - & 5.59E+01 (6.65E+00) - & 1.20E+02 (1.16E+01) - & 6.01E+02 (1.74E+01) - & 4.09E+01 (7.29E+00) - & 3.83E+01 (5.97E+00) - \\
 & 100D & \cellcolor{lightgray!30}\textbf{6.22E-02 (5.42E-02)} & 4.13E+02 (6.10E+01) - & 2.13E+02 (2.01E+01) - & 3.71E+02 (1.53E+01) - & 1.51E+03 (2.01E+01) - & 1.28E+02 (1.40E+01) - & 1.34E+02 (1.61E+01) - \\
 & 200D & \cellcolor{lightgray!30}\textbf{4.85E-01 (2.01E-01)} & 1.30E+03 (1.01E+02) - & 7.13E+02 (4.60E+01) - & 1.06E+03 (4.25E+01) - & 3.30E+03 (4.12E+01) - & 5.17E+02 (4.79E+01) - & 4.94E+02 (4.28E+01) - \\ \hline
 
\multirow{3}{*}{Rosenbrock} & 50D & \cellcolor{lightgray!30}\textbf{9.20E-05 (1.80E-04)} & 1.59E+04 (2.13E+04) - & 2.36E+03 (2.72E+02) - & 7.43E+03 (7.21E+02) - & 3.85E+05 (5.58E+04) - & 1.31E+03 (1.71E+02) - & 1.12E+03 (1.16E+02) - \\
 & 100D & \cellcolor{lightgray!30}\textbf{5.10E-03 (1.86E-02)} & 3.36E+05 (1.21E+05) - & 1.03E+04 (1.03E+03) - & 2.92E+04 (2.23E+03) - & 2.46E+06 (9.64E+04) - & 5.60E+03 (9.01E+02) - & 5.04E+03 (7.89E+02) - \\
 & 200D & \cellcolor{lightgray!30}\textbf{2.96E+01 (5.70E+01)} & 1.53E+06 (2.54E+05) - & 4.16E+04 (3.34E+03) - & 1.01E+05 (9.82E+03) - & 1.05E+07 (8.25E+05) - & 6.87E+04 (1.81E+04) - & 6.24E+04 (1.86E+04) - \\ \hline
 
\multirow{3}{*}{Schwefel} & 50D & \cellcolor{lightgray!30}\textbf{5.28E+02 (2.09E+02)} & 6.11E+03 (5.34E+02) - & 3.10E+03 (3.83E+02) - & 8.92E+03 (4.61E+02) - & 1.04E+04 (3.51E+02) - & 1.18E+04 (7.46E+02) - & 1.06E+04 (8.45E+02) - \\
 & 100D & \cellcolor{lightgray!30}\textbf{2.37E+03 (3.59E+02)} & 1.73E+04 (1.21E+03) - & 7.83E+03 (7.10E+02) - & 2.08E+04 (7.39E+02) - & 2.25E+04 (4.64E+02) - & 2.54E+04 (1.18E+03) - & 2.33E+04 (1.32E+03) - \\
 & 200D & \cellcolor{lightgray!30}\textbf{8.24E+03 (7.99E+02)} & 4.53E+04 (1.65E+03) - & 1.91E+04 (1.23E+03) - & 5.04E+04 (1.11E+03) - & 4.96E+04 (1.62E+03) - & 5.44E+04 (1.59E+03) - & 5.02E+04 (1.35E+03) - \\ \hline
 
\multirow{3}{*}{Sphere} & 50D & \cellcolor{lightgray!30}\textbf{0.00E+00 (0.00E+00)} & 6.96E+00 (4.75E+00) - & 5.78E-01 (3.38E-02) - & 1.36E+00 (1.24E-01) - & 1.32E+02 (6.14E+00) - & 2.23E-01 (1.06E-02) - & 1.79E-01 (1.29E-02) - \\
 & 100D & \cellcolor{lightgray!30}\textbf{5.20E-24 (2.74E-23)} & 6.79E+01 (1.95E+01) - & 2.22E+00 (1.15E-01) - & 4.11E+00 (1.89E-01) - & 6.08E+02 (1.42E+01) - & 2.17E+00 (6.01E-01) - & 2.04E+00 (9.74E-01) - \\
 & 200D & \cellcolor{lightgray!30}\textbf{1.30E-10 (1.00E-10)} & 3.34E+02 (4.22E+01) - & 8.36E+00 (3.81E-01) - & 1.25E+01 (5.10E-01) - & 1.47E+03 (4.99E+01) - & 3.96E+01 (6.96E+00) - & 4.44E+01 (1.23E+01) - \\ 
 \hline
\multicolumn{2}{c}{$+/ \approx/ -$} & -    &  0/0/18   &   0/0/18  &  0/0/18   &  0/0/18   &   0/0/18   &    0/0/18    \\
\bottomrule
\end{tabular}

\begin{tablenotes}[flushleft]
\footnotesize
\item[*] The Wilcoxon rank-sum tests (significance level $\alpha=0.05$) were conducted between AutoPSO and each individual algorithm. The final row indicates the number of tasks where the corresponding algorithm performs statistically better ($+$), similar ($\approx$), or worse ($-$) when compared to AutoPSO.
\end{tablenotes}

\end{threeparttable}
}

\end{table*}

}

\newcommand{\suppFigASR}{
\begin{figure}[htbp]
    \centering
    \includegraphics[width=0.22\textwidth]{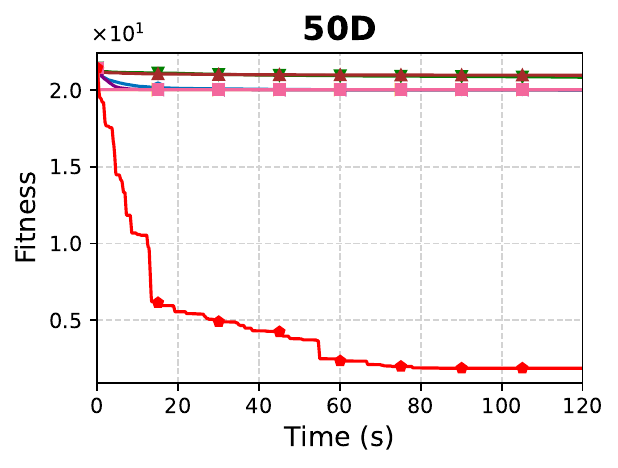}
    \hspace{-0.1cm}
    \includegraphics[width=0.22\textwidth]{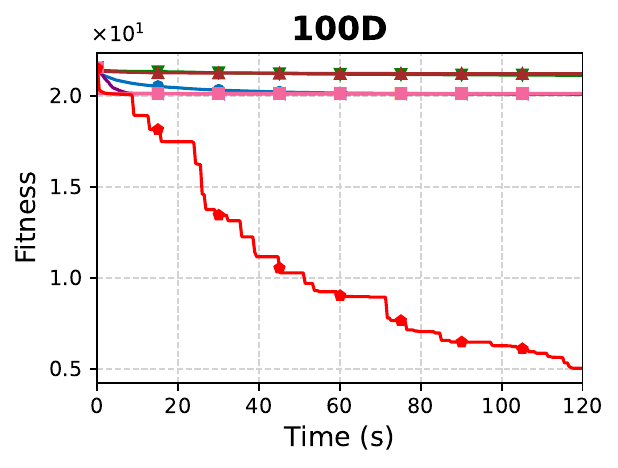}

    \vspace{-0.5cm}
    
    \includegraphics[width=0.22\textwidth]{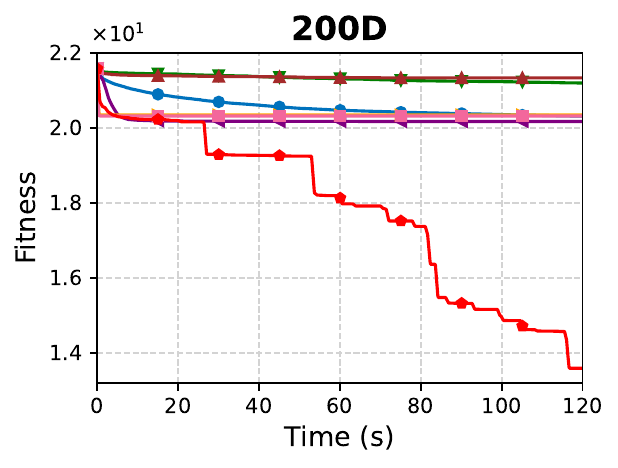}
    \hspace{-0.3cm}
    \includegraphics[width=0.22\textwidth]{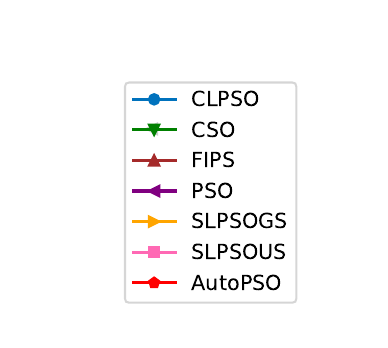}   
    \caption{Convergence curves on shifted and rotated Ackley function with three dimensions (50D, 100D and 200D).}
    \label{supp:fig:ackley-sr}
\end{figure}

}

\newcommand{\suppFigRSR}{
\begin{figure}[htbp]
    \centering
    \includegraphics[width=0.22\textwidth]{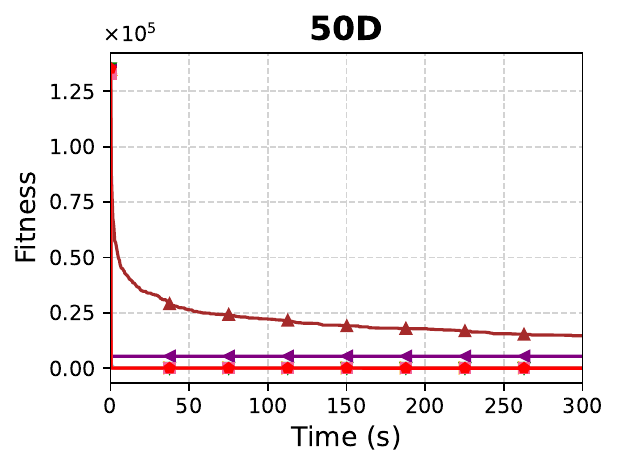}
    \hspace{-0.1cm}
    \includegraphics[width=0.22\textwidth]{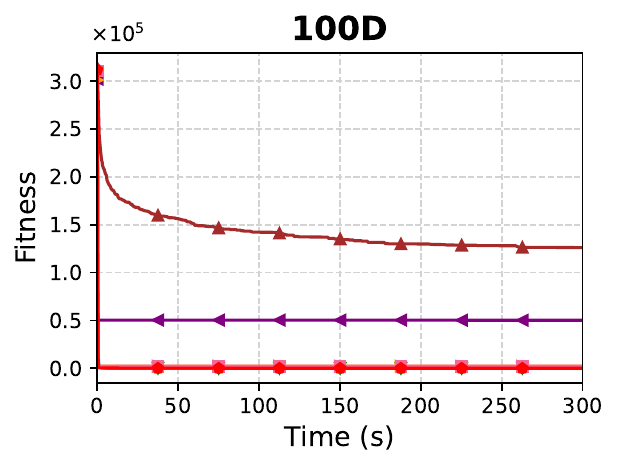}

    \vspace{-0.5cm}
    
    \includegraphics[width=0.22\textwidth]{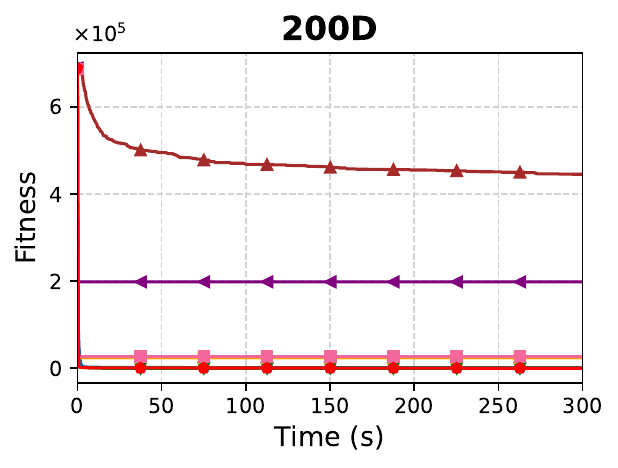}
    \hspace{-0.3cm}
    \includegraphics[width=0.22\textwidth]{fig/data/numerical/basic/label.pdf}   
    \caption{Convergence curves on shifted and rotated Rastrigin function with three dimensions (50D, 100D and 200D).}
    \label{supp:fig:rastrigin-sr}
\end{figure}
}

\newcommand{\suppFigAH}{
\begin{figure}[htbp]
    \centering
    \includegraphics[width=0.22\textwidth]{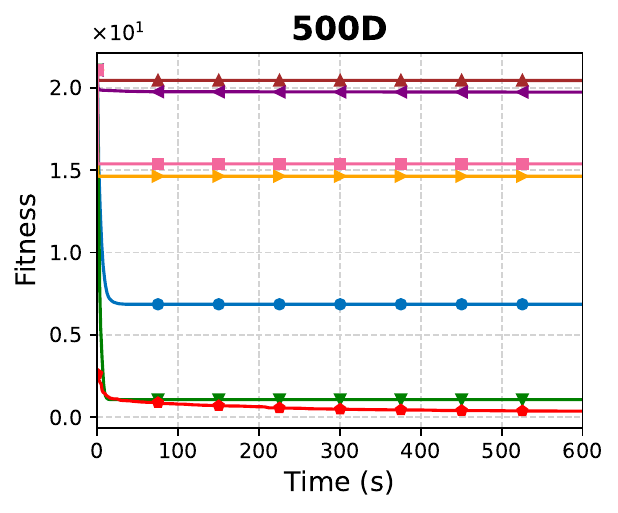}
    \hspace{-0.1cm}
    \includegraphics[width=0.22\textwidth]{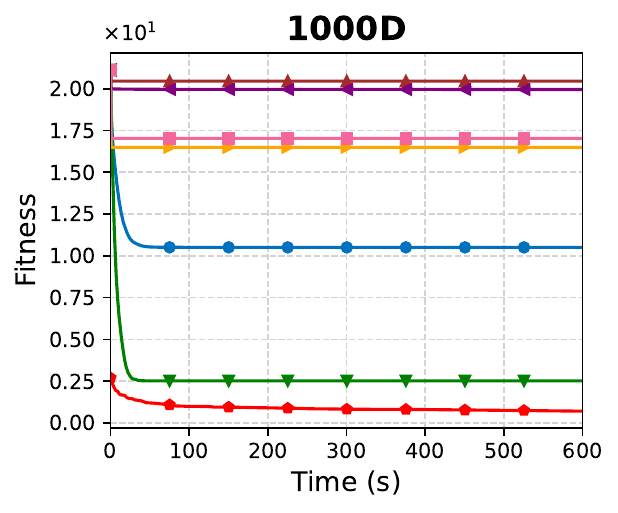}

    \vspace{-0.5cm}
    
    \includegraphics[width=0.22\textwidth]{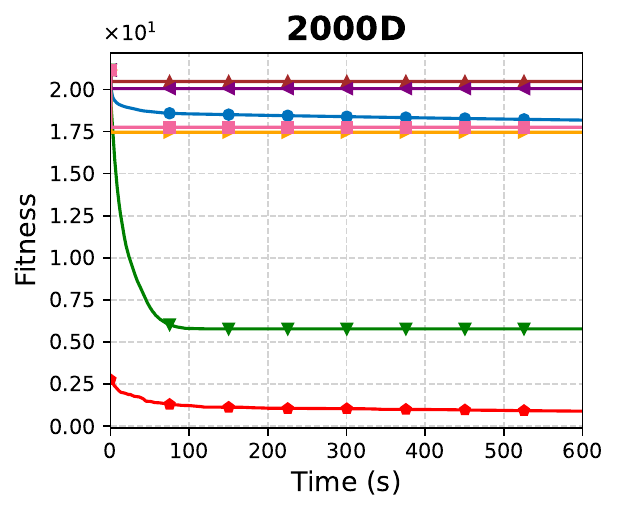}
    \hspace{-0.3cm}
    \includegraphics[width=0.22\textwidth]{fig/data/numerical/basic/label.pdf}   
    \caption{Convergence curves on very high dimensional Ackley function with three dimensions (500D, 1000D and 2000D).}
    \label{supp:fig:Ackley-high}
\end{figure}
}

\newcommand{\suppFigRH}{
\begin{figure}[htbp]
    \centering
    \includegraphics[width=0.22\textwidth]{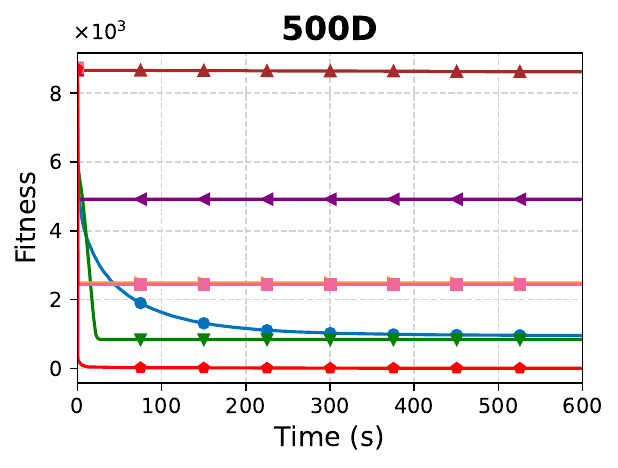}
    \hspace{-0.1cm}
    \includegraphics[width=0.22\textwidth]{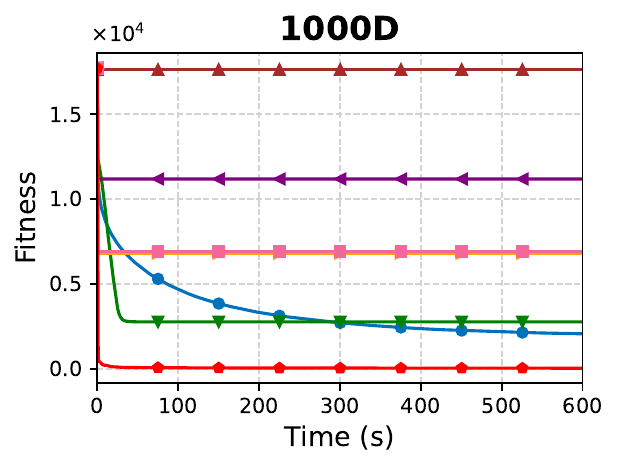}

    \vspace{-0.5cm}
    
    \includegraphics[width=0.22\textwidth]{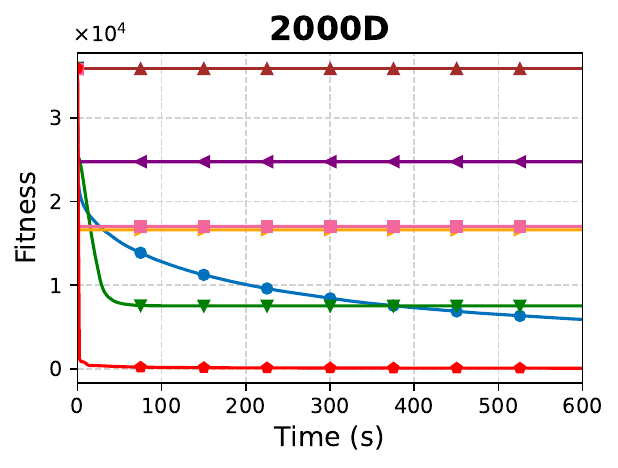}
    \hspace{-0.3cm}
    \includegraphics[width=0.22\textwidth]{fig/data/numerical/basic/label.pdf}   
    \caption{Convergence curves on vary high Rastrigin function with three dimensions (500D, 1000D and 2000D).}
    \label{supp:fig:Rastrigin-high}
\end{figure}
}

\newcommand{\suppFigBraxRe}{
\begin{figure}[htbp]
    \centering
    \includegraphics[width=0.24\textwidth]{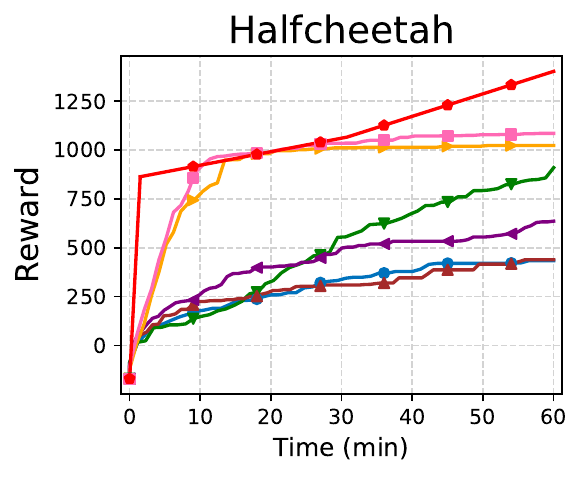}
    \hspace{-0.1cm}
    \includegraphics[width=0.24\textwidth]{fig/data/brax/new_hopper.pdf}
    
    \vspace{0.05cm}
    
    \includegraphics[width=0.24\textwidth]{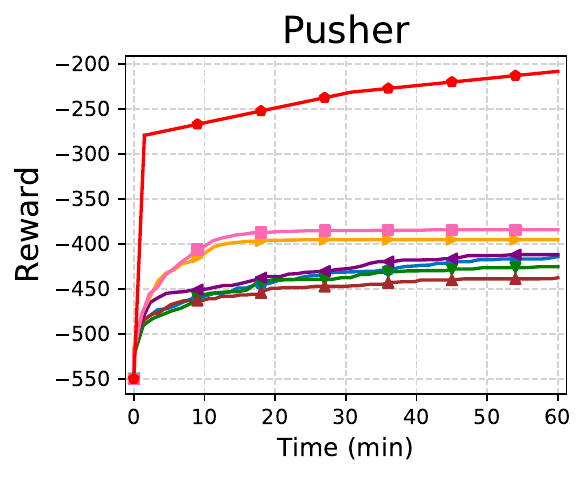}
    \hspace{-0.1cm}
    \includegraphics[width=0.24\textwidth]{fig/data/brax/new_reacher.pdf}

    \vspace{-0.3cm}

    \includegraphics[width=0.24\textwidth]{fig/data/brax/new_swimmer.pdf}
    \hspace{-0.3cm}
    \includegraphics[width=0.24\textwidth]{fig/data/brax/label.pdf}
    
    \caption{The reward curves achieved by AutoPSO and baseline PSO variants when applied to each robot control task.}
    \label{supp:fig:brax-rewards}
\end{figure}
}

\newcommand{\suppTabBraxRe}{
\begin{table*}[htbp]
\caption{Results on five robot control tasks based on Brax engine.}
\label{supp:tab:brax}
\centering
\scalebox{0.72}{
\begin{threeparttable}

\begin{tabular}{cccccccc}
\toprule
\textbf{Task}        & \textbf{AutoPSO} & \textbf{PSO}     & \textbf{CSO}     & \textbf{CLPSO}   & \textbf{FIPS}    & \textbf{SLPSOGS} & \textbf{SLPSOUS} \\
\midrule
Halfcheetah & \cellcolor{lightgray!30}\textbf{1.402E+03 (3.39E+02)} & 6.359E+02 (1.48E+02) - & 9.095E+02 (3.29E+02) - & 4.562E+02 (2.37E+02) - & 4.388E+02 (1.93E+02) - & 1.025E+03 (5.51E+02) $\approx$ & 1.085E+03 (4.54E+02) $\approx$ \\
Hopper & \cellcolor{lightgray!30}\textbf{1.263E+03 (1.23E+02)} & 1.197E+03 (2.59E+02) $\approx$ & 9.736E+02 (2.68E+02) - & 8.722E+02 (6.37E+01) - & 7.925E+02 (1.13E+02) - & 8.141E+02 (2.53E+02) - & 7.974E+02 (2.12E+02) - \\
Pusher & \cellcolor{lightgray!30}\textbf{-2.084E+02 (3.54E+01)} & -4.121E+02 (1.75E+01) - & -4.255E+02 (1.85E+01) - & -4.143E+02 (2.11E+01) - & -4.379E+02 (1.11E+01) - & -3.954E+02 (6.24E+01) - & -3.845E+02 (4.29E+01) - \\
Reacher & \cellcolor{lightgray!30}\textbf{-2.214E+01 (1.44E+01)} & -4.572E+02 (1.49E+02) - & -5.097E+02 (9.68E+01) - & -3.821E+02 (1.46E+02) - & -5.687E+02 (3.07E+01) - & -2.970E+02 (1.35E+02) - & -3.397E+02 (1.15E+02) - \\
Swimmer & \cellcolor{lightgray!30}\textbf{1.878E+02 (2.16E+00)} & 1.876E+02 (1.23E+00) $\approx$ & 1.831E+02 (1.39E+00) - & 1.779E+02 (3.85E+00) - & 1.721E+02 (5.19E+00) - & 1.853E+02 (1.33E+00) - & 1.853E+02 (9.66e-01) - \\ \hline
+ / $\approx$ / - &  - & 0/2/3 & 0/0/5 & 0/0/5 & 0/0/5 & 0/1/4 & 0/1/4 \\ 
\bottomrule
\end{tabular}

\begin{tablenotes}[flushleft]
\footnotesize
\item[*] The Wilcoxon rank-sum tests (significance level $\alpha=0.05$) were conducted between AutoPSO and each algorithm individually.
The final row displays the number of tasks where the corresponding algorithm performs statistically better ($+$), similar ($\approx$), or worse ($-$) compared to AutoPSO.
\end{tablenotes}

\end{threeparttable}}
\end{table*}
}

\newcommand{\suppFigCross}{
\begin{figure}[htbp]
   
    \centering
    \includegraphics[width=0.24\textwidth]{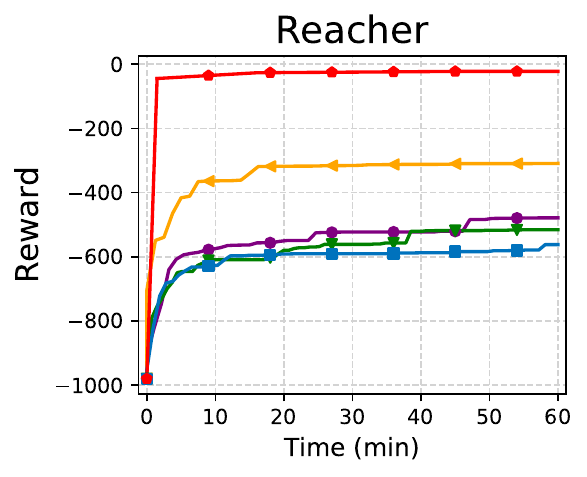}
    \hspace{-0.1cm}
    \includegraphics[width=0.24\textwidth]{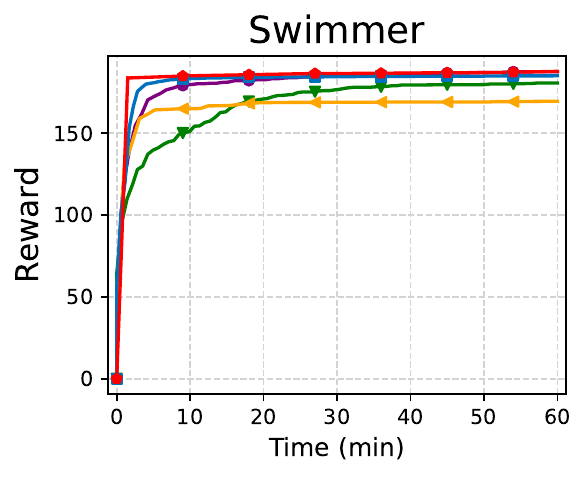}
    
    \vspace{-0.8cm}
    
   \includegraphics[width=0.23\textwidth]{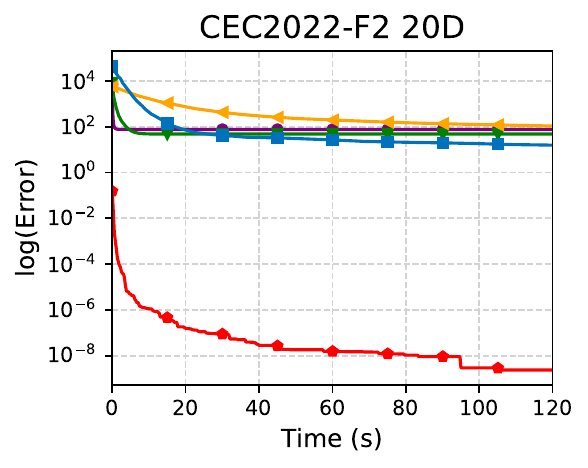}
    \hspace{-0.4cm}
    \includegraphics[width=0.24\textwidth]{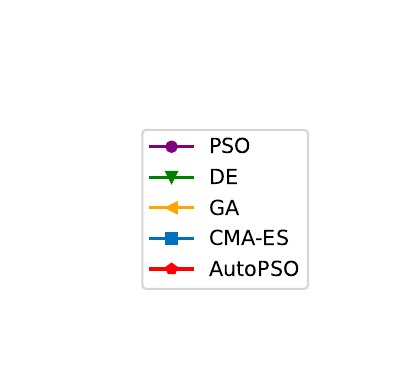}
    
    \caption{The convergence curves achieved by AutoPSO with four EAs.}
    \label{supp:fig:cross}
\end{figure}
}

\newcommand{\suppTabGen}{
\begin{table}[htbp]
\caption{Generalization performance of AutoPSO and variants on CEC2022-20D functions.}
\label{supp:tab:generalize}
\centering
\scalebox{0.54}{
\begin{threeparttable}
\begin{tabular}{cccccc}
\toprule
\multirow{2}{*}{\textbf{Func}} 
& \multirow{2}{*}{\textbf{AutoPSO}} 
& \multicolumn{4}{c}{\textbf{Variants find on}} \\
\cline{3-6}
&  & \textbf{-$F_6$} & \textbf{-$F_8$} & \textbf{-$F_{10}$} & \textbf{-$F_{12}$} \\
\midrule

$F_1$  & \cellcolor{lightgray!30}\textbf{0.00E+00 (0.00E+00)} & \cellcolor{lightgray!30}\textbf{0.00E+00 (0.00E+00)}$\approx$ & 2.40E+02 (3.66E+02)$-$ & \cellcolor{lightgray!30}\textbf{0.00E+00 (0.00E+00)}$\approx$ & \cellcolor{lightgray!30}\textbf{0.00E+00 (0.00E+00)}$\approx$ \\
$F_2$  & \cellcolor{lightgray!30}\textbf{2.36E-09 (6.30E-09)} & 3.98E+01 (1.88E+01)$-$ & 5.22E+01 (1.01E+01)$-$ & 6.88E+01 (2.78E+01)$-$ & 4.80E+01 (1.87E+00)$-$ \\
$F_3$  & \cellcolor{lightgray!30}\textbf{0.00E+00 (0.00E+00)} & 8.36E-03 (4.10E-03)$-$ & 3.42E-02 (1.08E-01)$-$ & 1.09E+00 (1.22E+00)$-$ & \cellcolor{lightgray!30}\textbf{0.00E+00 (0.00E+00)}$\approx$ \\
$F_4$  & \cellcolor{lightgray!30}\textbf{0.00E+00 (0.00E+00)} & 3.06E+01 (5.59E+00)$-$ & 1.93E+01 (6.66E+00)$-$ & 3.40E+01 (1.12E+01)$-$ & 6.02E+01 (6.67E+00)$-$ \\
$F_5$  & \cellcolor{lightgray!30}\textbf{0.00E+00 (0.00E+00)} & 9.78E+00 (1.31E+01)$-$ & {6.89E-01 (1.41E+00)}$\approx$ & 1.74E+01 (5.41E+01)$-$ & \cellcolor{lightgray!30}\textbf{0.00E+00 (0.00E+00)}$-$ \\
$F_6$  & \cellcolor{lightgray!30}\textbf{4.57E+00 (5.27E+00)} & \cellcolor{lightgray!30}\textbf{4.57E+00 (5.27E+00)}$\approx$ & 1.39E+04 (1.17E+04)$-$ & 1.43E+05 (4.08E+05)$-$ & 1.10E+04 (7.44E+03)$-$ \\
$F_7$  & \cellcolor{lightgray!30}\textbf{0.00E+00 (0.00E+00)} & 5.85E+01 (2.19E+01)$-$ & 3.25E+01 (5.35E+00)$-$ & 2.98E+01 (6.96E+00)$-$ & 2.28E+01 (1.76E+00)$-$ \\
$F_8$  & \cellcolor{lightgray!30}\textbf{6.42E-02 (6.20E-02)} & 4.27E+01 (3.40E+01)$-$ & \cellcolor{lightgray!30}\textbf{6.42E-02 (6.20E-02)}$\approx$ & 4.44E+01 (4.84E+01)$-$ & 2.53E+01 (7.17E+00)$-$ \\
$F_9$  & \cellcolor{lightgray!30}\textbf{1.72E+02 (9.16E+01)} & 1.81E+02 (1.44E-03)$-$ & 1.84E+02 (7.26E+00)$-$ & 1.87E+02 (1.96E+01)$\approx$ & 1.81E+02 (3.84E-04)$-$ \\
$F_{10}$ & \cellcolor{lightgray!30}\textbf{5.38E+01 (4.13E+01)} & 1.84E+03 (1.62E+03)$-$ & {5.16E+02 (4.62E+02)}$-$ & 9.69E+02 (7.53E+02)$-$ & {2.63E+02 (2.49E+02)}$-$ \\
$F_{11}$ & \cellcolor{lightgray!30}\textbf{0.00E+00 (0.00E+00)} & 3.18E+02 (3.86E+01)$-$ & 4.20E+02 (1.68E+02)$-$ & 6.09E+02 (3.79E+02)$-$ & 3.36E+02 (4.81E+01)$-$ \\
$F_{12}$ & \cellcolor{lightgray!30}\textbf{1.59E+02 (5.93E-04)} & 2.58E+02 (1.75E+01)$-$ & 2.58E+02 (1.10E+01)$-$ & 2.62E+02 (1.84E+01)$-$ & 2.43E+02 (6.58E+00)$-$ \\

\midrule
+ / $\approx$ / - &  - & 0/2/10 & 0/2/10 & 0/3/9 & 0/3/9 \\
\bottomrule
\end{tabular}

\begin{tablenotes}[flushleft]
\footnotesize
\item[*] The Wilcoxon rank-sum tests (significance level $\alpha=0.05$) were conducted between AutoPSO and each algorithm individually.
The final row displays the number of tasks where the corresponding algorithm performs statistically better ($+$), similar ($\approx$), or worse ($-$) compared to AutoPSO.
\end{tablenotes}

\end{threeparttable}}
\end{table}

}

\newcommand{\suppTabComponent}{
\begin{table*}[htbp]
\caption{Ablation study of AutoPSO on CEC2022-20D.}
\label{supp:tab:component}
\centering
\scalebox{0.63}{
\begin{threeparttable}
\begin{tabular}{cc|ccc|cccc|c}
\toprule
\textbf{Func} 
& \textbf{AutoPSO} 
& \textbf{AutoPSO-c} 
& \textbf{w/o. acc.} 
& \textbf{w/o. weight} 
& \textbf{AutoPSO-e} 
& \textbf{w/o. random} 
& \textbf{w/o. self} 
& \textbf{w/o. social} 
& \textbf{AutoPSO-m} \\
\midrule

$F_1$  & \cellcolor{lightgray!30}\textbf{0.00E+00 (0.00E+00)} & 1.61E-07 (5.42E-07)$-$ & \cellcolor{lightgray!30}\textbf{0.00E+00 (0.00E+00)}$\approx$ &\cellcolor{lightgray!30}\textbf{0.00E+00 (0.00E+00)}$\approx$ & \cellcolor{lightgray!30}\textbf{0.00E+00 (0.00E+00)}$\approx$ & \cellcolor{lightgray!30}\textbf{0.00E+00 (0.00E+00)}$\approx$ & \cellcolor{lightgray!30}\textbf{0.00E+00 (0.00E+00)}$\approx$ & 5.12E+00 (1.10E+01)$\approx$ & \cellcolor{lightgray!30}\textbf{0.00E+00 (0.00E+00)}$\approx$ \\
$F_2$  & \cellcolor{lightgray!30}\textbf{2.36E-09 (6.30E-09)} & 7.23E-01 (1.50E+00)$-$ & 4.79E-07 (1.07E-06)$\approx$ & 1.72E-04 (4.76E-04)$-$ & 6.14E-03 (3.03E-02)$-$ & 1.62E-05 (5.07E-05)$-$ & 1.33E-06 (2.00E-06)$-$ & 3.16E-02 (5.79E-02)$-$ & 2.26E-08 (3.58E-08)$-$ \\
$F_3$  & \cellcolor{lightgray!30}\textbf{0.00E+00 (0.00E+00)} & \cellcolor{lightgray!30}\textbf{0.00E+00 (0.00E+00)}$\approx$ & \cellcolor{lightgray!30}\textbf{0.00E+00 (0.00E+00)}$\approx$ & \cellcolor{lightgray!30}\textbf{0.00E+00 (0.00E+00)}$\approx$ & 1.26E-03 (1.04E-03)$-$ & 2.62E-05 (2.78E-05)$-$ & \cellcolor{lightgray!30}\textbf{0.00E+00 (0.00E+00)}$\approx$ & \cellcolor{lightgray!30}\textbf{0.00E+00 (0.00E+00)}$\approx$ & \cellcolor{lightgray!30}\textbf{0.00E+00 (0.00E+00)}$\approx$ \\
$F_4$  & \cellcolor{lightgray!30}\textbf{0.00E+00 (0.00E+00)} & 8.01E-01 (5.65E-01)$-$ & \cellcolor{lightgray!30}\textbf{0.00E+00 (0.00E+00)}$\approx$ & \cellcolor{lightgray!30}\textbf{0.00E+00 (0.00E+00)}$\approx$ & 6.33E+00 (1.25E+00)$-$ & 1.04E+00 (9.34E-01)$-$ & \cellcolor{lightgray!30}\textbf{0.00E+00 (0.00E+00)}$\approx$ & \cellcolor{lightgray!30}\textbf{0.00E+00 (0.00E+00)}$\approx$ & \cellcolor{lightgray!30}\textbf{0.00E+00 (0.00E+00)}$\approx$ \\
$F_5$  & \cellcolor{lightgray!30}\textbf{0.00E+00 (0.00E+00)} & \cellcolor{lightgray!30}\textbf{0.00E+00 (0.00E+00)}$\approx$ & \cellcolor{lightgray!30}\textbf{0.00E+00 (0.00E+00)}$\approx$ & \cellcolor{lightgray!30}\textbf{0.00E+00 (0.00E+00)}$\approx$ & 2.90E-03 (1.58E-02)$-$ & 1.06E-07 (2.16E-07)$\approx$ & \cellcolor{lightgray!30}\textbf{0.00E+00 (0.00E+00)}$\approx$ & \cellcolor{lightgray!30}\textbf{0.00E+00 (0.00E+00)}$\approx$ & \cellcolor{lightgray!30}\textbf{0.00E+00 (0.00E+00)}$\approx$ \\
$F_6$  & \cellcolor{lightgray!30}\textbf{4.57E+00 (5.27E+00)} & 5.61E+01 (2.42E+01)$-$ & 5.18E+00 (3.23E+00)$\approx$ & 1.59E+01 (1.07E+01)$-$ & 6.18E+01 (3.25E+01)$-$ & 2.60E+01 (1.19E+01)$-$ & 1.69E+01 (1.08E+01)$-$ & 1.40E+01 (8.76E+00)$-$ & 7.50E+00 (7.23E+00)$\approx$ \\
$F_7$  & \cellcolor{lightgray!30}\textbf{0.00E+00 (0.00E+00)} & 1.51E+00 (1.16E+00)$-$ & 2.14E-01 (2.42E-01)$-$ & 1.36E-02 (1.14E-02)$-$ & 4.89E+00 (3.74E+00)$-$ & 5.26E+00 (4.04E+00)$-$ & 1.04E-02 (7.38E-03)$-$ & 1.04E-02 (6.18E-03)$-$ & \cellcolor{lightgray!30}\textbf{0.00E+00 (0.00E+00)}$\approx$ \\
$F_8$  & \cellcolor{lightgray!30}\textbf{6.42E-02 (6.20E-02)} & 1.77E+01 (4.30E+00)$-$ & 8.46E+00 (8.24E+00)$-$ & 1.76E+00 (2.02E+00)$-$ & 6.70E+00 (7.36E+00)$-$ & 5.08E+00 (5.33E+00)$-$ & 1.21E+00 (8.11E-01)$-$ & 2.22E+00 (4.55E+00)$-$ & 1.28E-01 (4.51E-01)$\approx$ \\
$F_9$  & \cellcolor{lightgray!30}\textbf{1.72E+02 (9.16E+01)} & 1.81E+02 (8.11E-05)$-$ & 1.81E+02 (4.68E-05)$-$ & 1.81E+02 (1.61E-05)$-$ & 1.81E+02 (8.94E-05)$-$ & 1.81E+02 (6.65E-05)$-$ & 1.81E+02 (3.40E-05)$-$ & 1.81E+02 (1.27E-05)$-$ & 1.92E+02 (7.15E+01)$\approx$ \\
$F_{10}$ & {5.38E+01 (4.13E+01)} & 2.13E+01 (2.78E+01)$+$ & 3.05E+01 (4.29E+01)$\approx$ & 5.60E-01 (1.01E+00)$+$ & 5.13E+01 (4.32E+01)$\approx$ & 1.00E+02 (1.44E-02)$-$ & 1.97E+00 (4.86E+00)$+$ & \cellcolor{lightgray!30}\textbf{4.89E-01 (6.41E-01)}$+$ & 4.23E+01 (4.33E+01)$\approx$ \\
$F_{11}$ & \cellcolor{lightgray!30}\textbf{0.00E+00 (0.00E+00)} & 2.14E+02 (1.35E+02)$-$ & 1.91E+02 (1.44E+02)$-$ & 2.45E+02 (1.16E+02)$-$ & 1.52E+02 (1.58E+02)$-$ & 1.64E+02 (1.49E+02)$-$ & 2.45E+02 (1.16E+02)$-$ & 3.00E+02 (0.00E+00)$-$ & 2.90E+01 (8.87E+01)$\approx$ \\
$F_{12}$ & \cellcolor{lightgray!30}\textbf{1.59E+02 (5.93E-04)} & 2.30E+02 (1.04E+00)$-$ & 2.30E+02 (7.46E-01)$-$ & 2.28E+02 (7.13E-01)$-$ & 2.34E+02 (2.36E+00)$-$ & 2.29E+02 (1.79E+00)$-$ & 2.28E+02 (8.08E-01)$-$ & 2.28E+02 (4.90E-01)$-$ & \cellcolor{lightgray!30}\textbf{1.59E+02 (2.47E-03)}$-$ \\

\midrule
\textit{+/$\approx$/-} & - & 1/2/9 & 0/6/6 & 1/4/7 & 0/2/10 & 0/2/10 & 1/4/7 & 1/4/7 & 0/10/2 \\
\bottomrule
\end{tabular}
\begin{tablenotes}[flushleft]
\footnotesize
\item[*] The Wilcoxon rank-sum tests (significance level $\alpha=0.05$) were conducted between AutoPSO and each variant individually.
The final row displays the number of tasks where the corresponding algorithm performs statistically better ($+$), similar ($\approx$), or worse ($-$) compared to AutoPSO.
\end{tablenotes}

\end{threeparttable}
}
\end{table*}

}

\newcommand{\suppTabSub}{
\begin{table}[t]
\centering
\caption{Comparison of different swarm partition strategies on CEC2022-20D.}
\label{supp:tab:subswarm}
\scalebox{0.54}{
\begin{threeparttable}
\begin{tabular}{cccccc}
\toprule
\textbf{Func} & \textbf{\makecell{AutoPSO\\(2-sub, searched)}} & \textbf{1-swarm} & \textbf{\makecell{2-subswarm \\(fixed, 0.5/0.5)}} & \textbf{\makecell{3-subswarm\\(searched)}} & \textbf{\makecell{4-subswarm\\(searched)}} \\
\midrule

$F_1$ &
\cellcolor{lightgray!30}\textbf{0.00E+00 (0.00E+00)} &
\cellcolor{lightgray!30}\textbf{0.00E+00 (0.00E+00)}$\approx$ &
\cellcolor{lightgray!30}\textbf{0.00E+00 (0.00E+00)}$\approx$ &
\cellcolor{lightgray!30}\textbf{0.00E+00 (0.00E+00)}$\approx$ &
\cellcolor{lightgray!30}\textbf{0.00E+00 (0.00E+00)}$\approx$ \\

$F_2$ &
\cellcolor{lightgray!30}\textbf{2.36E-09 (6.30E-09)} &
2.26E-08 (3.58E-08)$-$ &
5.62E-05 (1.10E-04)$-$ &
1.98E-05 (4.26E-05)$-$ &
1.55E-05 (3.59E-05)$-$ \\

$F_3$ &
\cellcolor{lightgray!30}\textbf{0.00E+00 (0.00E+00)} &
0.00E+00 (0.00E+00)$\approx$ &
1.28E-08 (3.71E-08)$\approx$ &
\cellcolor{lightgray!30}\textbf{0.00E+00 (0.00E+00)}$\approx$ &
\cellcolor{lightgray!30}\textbf{0.00E+00 (0.00E+00)}$\approx$ \\

$F_4$ &
\cellcolor{lightgray!30}\textbf{0.00E+00 (0.00E+00)} &
\cellcolor{lightgray!30}\textbf{0.00E+00 (0.00E+00)}$\approx$ &
3.94E+00 (1.34E+00)$-$ &
\cellcolor{lightgray!30}\textbf{0.00E+00 (0.00E+00)}$\approx$ &
\cellcolor{lightgray!30}\textbf{0.00E+00 (0.00E+00)}$\approx$ \\

$F_5$ &
\cellcolor{lightgray!30}\textbf{0.00E+00 (0.00E+00)} &
\cellcolor{lightgray!30}\textbf{0.00E+00 (0.00E+00)}$\approx$ &
\cellcolor{lightgray!30}\textbf{0.00E+00 (0.00E+00)}$\approx$ &
\cellcolor{lightgray!30}\textbf{0.00E+00 (0.00E+00)}$\approx$ &
\cellcolor{lightgray!30}\textbf{0.00E+00 (0.00E+00)}$\approx$ \\

$F_6$ &
\cellcolor{lightgray!30}\textbf{4.57E+00 (5.27E+00)} &
7.50E+00 (7.23E+00)$\approx$ &
3.85E+01 (1.14E+01)$-$ &
1.90E+01 (1.29E+01)$-$ &
1.79E+01 (1.27E+01)$-$ \\

$F_7$ &
\cellcolor{lightgray!30}\textbf{0.00E+00 (0.00E+00)} &
\cellcolor{lightgray!30}\textbf{0.00E+00 (0.00E+00)}$\approx$ &
3.12E+00 (2.39E+00)$-$ &
1.30E-02 (1.09E-02)$-$ &
1.91E-02 (2.01E-02)$-$ \\

$F_8$ &
\cellcolor{lightgray!30}\textbf{6.42E-02 (6.20E-02)} &
1.28E-01 (4.51E-01)$\approx$ &
2.72E+00 (3.01E+00)$-$ &
6.49E-01 (3.77E-01)$-$ &
1.57E+00 (2.38E+00)$-$ \\

$F_9$ &
\cellcolor{lightgray!30}\textbf{1.72E+02 (9.16E+01)} &
1.92E+02 (7.15E+01)$\approx$ &
1.81E+02 (5.93E-05)$-$ &
1.81E+02 (3.49E-05)$-$ &
1.81E+02 (3.53E-05)$-$ \\

$F_{10}$ &
{5.38E+01 (4.13E+01)} &
4.23E+01 (4.33E+01)$\approx$ &
7.58E+01 (3.99E+01)$-$ &
1.07E+01 (2.84E+01)$+$ &
\cellcolor{lightgray!30}\textbf{1.04E+01 (2.84E+01)}$+$ \\

$F_{11}$ &
\cellcolor{lightgray!30}\textbf{0.00E+00 (0.00E+00)} &
\cellcolor{lightgray!30}\textbf{0.00E+00 (0.00E+00)}$\approx$ &
1.36E+02 (1.49E+02)$-$ &
1.21E+02 (1.39E+02)$-$ &
6.28E+01 (1.15E+02)$\approx$ \\

$F_{12}$ &
\cellcolor{lightgray!30}\textbf{1.59E+02 (5.93E-04)} &
1.59E+02 (2.47E-03)$-$ &
2.30E+02 (2.11E+00)$-$ &
2.28E+02 (8.60E-01)$-$ &
2.29E+02 (9.11E-01)$-$ \\

\midrule
\textit{+/$\approx$/-} & - & 0/10/2 & 0/3/9 & 1/4/7 & 1/5/6 \\
\bottomrule
\end{tabular}
\begin{tablenotes}[flushleft]
\footnotesize
\item[*] The Wilcoxon rank-sum tests (significance level $\alpha=0.05$) were conducted between AutoPSO and each algorithm individually.
The final row displays the number of tasks where the corresponding algorithm performs statistically better ($+$), similar ($\approx$), or worse ($-$) compared to AutoPSO.
\end{tablenotes}

\end{threeparttable}
}
\end{table}

}
\newcommand{\suppTabInd}{
\begin{table}[t]
\centering
\caption{Individual-level heterogeneity analysis on CEC2022-20D.}
\label{supp:tab:ind_analysis}
\scalebox{0.68}{
\begin{threeparttable}
\begin{tabular}{cccccc}
\hline
Func & AutoPSO & IndExemplar & IndParam & IndRandom \\
\hline
$F_{1}$  & \cellcolor{lightgray!30}\textbf{0.00E+00 (0.00E+00)} & \cellcolor{lightgray!30}\textbf{0.00E+00 (0.00E+00)}$\approx$ & \cellcolor{lightgray!30}\textbf{0.00E+00 (0.00E+00)}$\approx$ & 1.59E+01 (4.95E+01)$\approx$ \\
$F_{2}$  & \cellcolor{lightgray!30}\textbf{2.36E-09 (6.30E-09)} & 1.20E-09 (3.78E-09)$\approx$ & 4.13E-06 (6.47E-06)$-$ & 5.47E+01 (2.11E+01)$-$ \\
$F_{3}$  & \cellcolor{lightgray!30}\textbf{0.00E+00 (0.00E+00)} & \cellcolor{lightgray!30}\textbf{0.00E+00 (0.00E+00)}$\approx$ & \cellcolor{lightgray!30}\textbf{0.00E+00 (0.00E+00)}$\approx$ & 1.55E+01 (6.44E+00)$-$ \\
$F_{4}$  & \cellcolor{lightgray!30}\textbf{0.00E+00 (0.00E+00)} & \cellcolor{lightgray!30}\textbf{0.00E+00 (0.00E+00)}$\approx$ & \cellcolor{lightgray!30}\textbf{0.00E+00 (0.00E+00)}$\approx$ & 5.44E+01 (2.09E+01)$-$ \\
$F_{5}$  & \cellcolor{lightgray!30}\textbf{0.00E+00 (0.00E+00)} & \cellcolor{lightgray!30}\textbf{0.00E+00 (0.00E+00)}$\approx$ & \cellcolor{lightgray!30}\textbf{0.00E+00 (0.00E+00)}$\approx$ & 4.10E+02 (1.96E+02)$-$ \\
$F_{6}$  & 4.57E+00 (5.27E+00) & \cellcolor{lightgray!30}\textbf{7.12E-01 (7.59E-01)}$\approx$ & 2.39E+01 (9.29E+00)$-$ & 3.01E+03 (2.57E+03)$-$ \\
$F_{7}$  & \cellcolor{lightgray!30}\textbf{0.00E+00 (0.00E+00)} & \cellcolor{lightgray!30}\textbf{0.00E+00 (0.00E+00)}$\approx$ & 1.59E-02 (1.63E-02)$-$ & 8.24E+01 (2.59E+01)$-$ \\
$F_{8}$  & 6.42E-02 (6.20E-02) & \cellcolor{lightgray!30}\textbf{4.63E-02 (2.20E-02)}$\approx$ & 5.73E+00 (5.66E+00)$-$ & 2.74E+01 (4.67E+00)$-$ \\
$F_{9}$  & \cellcolor{lightgray!30}\textbf{1.72E+02 (9.16E+01)} & 1.79E+02 (7.92E+01)$\approx$ & 1.81E+02 (3.04E-05)$-$ & 1.81E+02 (5.93E-01)$-$ \\
$F_{10}$ & \cellcolor{lightgray!30}\textbf{5.38E+01 (4.13E+01)} & 8.21E+01 (3.24E+01)$-$ & 3.29E+00 (5.28E+00)$+$ & 1.01E+02 (4.23E-01)$-$ \\
$F_{11}$ & \cellcolor{lightgray!30}\textbf{0.00E+00 (0.00E+00)} & 2.73E+01 (8.62E+01)$\approx$ & 2.34E+02 (1.16E+02)$-$ & 3.09E+02 (2.90E+01)$-$ \\
$F_{12}$ & \cellcolor{lightgray!30}\textbf{1.59E+02 (5.93E-04)} & 1.59E+02 (5.01E-04)$+$ & 2.29E+02 (8.41E-01)$-$ & 2.82E+02 (2.94E+01)$-$ \\
\midrule
\textit{+/$\approx$/-} & - & 1/10/1 & 1/4/7 & 0/1/11 \\
\bottomrule
\end{tabular}
\begin{tablenotes}[flushleft]
\footnotesize
\item[*] The Wilcoxon rank-sum tests (significance level $\alpha=0.05$) were conducted between AutoPSO and each algorithm individually.
The final row displays the number of tasks where the corresponding algorithm performs statistically better ($+$), similar ($\approx$), or worse ($-$) compared to AutoPSO.
\end{tablenotes}

\end{threeparttable}
}
\end{table}

}
\newcommand{\suppTabAda}{
\begin{table}[t]
\centering
\caption{Comparison with adaptive PSO variants on CEC2022-20D.}
\label{supp:tab:adaptive}
\scalebox{0.7}{
\begin{threeparttable}
\begin{tabular}{cccccc}
\hline
\textbf{Func} & \textbf{InnerPSO} & \textbf{iwPSO} & \textbf{TVAC-PSO} \\

\hline
$F_{1}$  & \cellcolor{lightgray!30}\textbf{0.00E+00 (0.00E+00)} & 1.81E+04 (1.44E+04) $-$ & \cellcolor{lightgray!30}\textbf{0.00E+00 (0.00E+00)} $\approx$\\
$F_{2}$  & \cellcolor{lightgray!30}\textbf{4.83E+01 (1.68E+00)} & 3.39E+02 (1.71E+02) $-$& 5.83E+01 (1.77E+01) $-$\\
$F_{3}$  & 2.14E-02 (4.37E-02) & 2.66E+01 (1.27E+01) $-$ & \cellcolor{lightgray!30}\textbf{3.62E-05 (7.25E-05)} $\approx$\\
$F_{4}$  & \cellcolor{lightgray!30}\textbf{1.24E+01 (5.20E+00)} & 1.17E+02 (1.73E+01) $-$& 3.94E+01 (6.74E+00) $-$\\
$F_{5}$  & 2.33E+00 (1.95E+00) & 1.09E+03 (2.63E+02) $-$& \cellcolor{lightgray!30}\textbf{9.42E-01 (5.46E-01)} $+$\\
$F_{6}$  & \cellcolor{lightgray!30}\textbf{1.62E+05 (4.23E+05)} & 2.29E+07 (1.96E+07) $-$& 2.90E+05 (5.71E+05) $-$\\
$F_{7}$  & \cellcolor{lightgray!30}\textbf{1.87E+01 (9.05E+00)} & 6.85E+01 (3.33E+01) $-$& 1.90E+01 (6.45E+00) $-$\\
$F_{8}$  & 2.57E+01 (2.19E+00) & 4.11E+01 (6.71E+00) $-$& \cellcolor{lightgray!30}\textbf{2.12E+01 (2.15E-01)} $+$\\
$F_{9}$  & \cellcolor{lightgray!30}\textbf{1.82E+02 (1.91E+00)} & 2.67E+02 (6.68E+01) $-$& 1.87E+02 (1.18E+01) $-$\\
$F_{10}$ & 1.00E+03 (7.81E+02) & 1.26E+03 (1.34E+03) $-$& \cellcolor{lightgray!30}\textbf{1.11E+02 (2.16E+01)} $+$\\
$F_{11}$ & \cellcolor{lightgray!30}\textbf{3.00E+02 (0.00E+00)} & 2.85E+03 (1.12E+03) $-$& 5.98E+02 (2.25E+02) $-$\\
$F_{12}$ & \cellcolor{lightgray!30}\textbf{2.43E+02 (6.72E+00)} & 2.90E+02 (2.00E+01) $-$& 2.50E+02 (1.05E+01) $-$\\
\midrule
\textit{+/$\approx$/-} & - & 0/0/12 & 3/2/7  \\
\bottomrule
\end{tabular}
\begin{tablenotes}[flushleft]
\footnotesize
\item[*] The Wilcoxon rank-sum tests (significance level $\alpha=0.05$) were conducted between AutoPSO and each algorithm individually.
The final row displays the number of tasks where the corresponding algorithm performs statistically better ($+$), similar ($\approx$), or worse ($-$) compared to AutoPSO.
\end{tablenotes}

\end{threeparttable}
}
\end{table}

}
\newcommand{\suppFigOuter}{
\begin{figure}[htbp]
   
    \centering
    \includegraphics[width=0.24\textwidth]{fig/data/validation/outer/halfcheetah-120.pdf}
    \hspace{-0.1cm}
    \includegraphics[width=0.24\textwidth]{fig/data/validation/outer/hopper-120.pdf}
    
    \vspace{0.05cm}
    
    \includegraphics[width=0.24\textwidth]{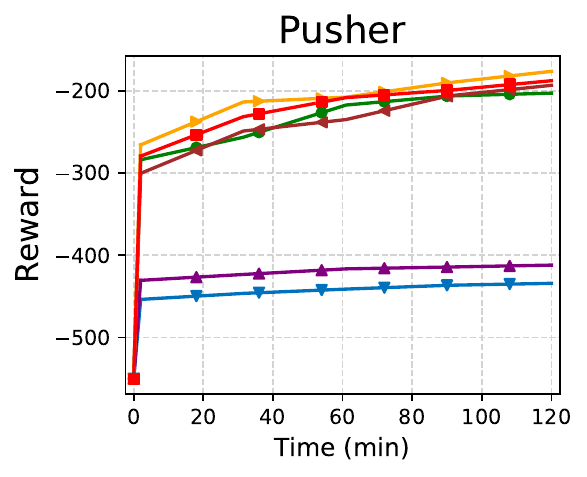}
    \hspace{-0.1cm}
    \includegraphics[width=0.24\textwidth]{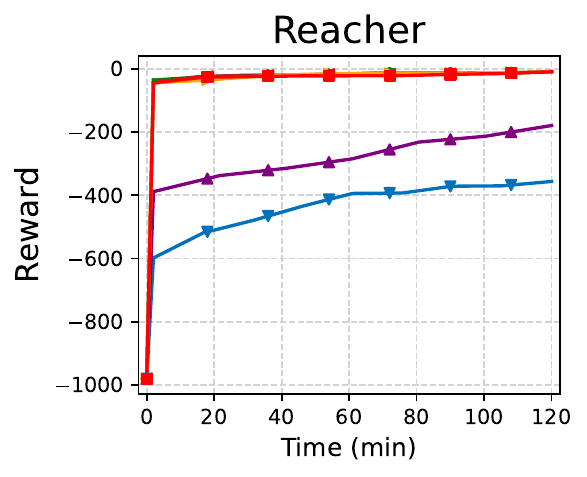}
    
    \vspace{0.05cm}
    
    \includegraphics[width=0.24\textwidth]{fig/data/validation/outer/swimmer-120.pdf}
    \hspace{-0.1cm}
    \includegraphics[width=0.24\textwidth]{fig/data/validation/outer/label.pdf}
    
    \caption{The reward curves achieved by AutoPSO with different outer optimizers when applied to each robot control task.}
    \label{supp:fig:outer}
\end{figure}

}
\newcommand{\suppTabOuter}{
\begin{table*}[htbp]
\centering
\scalebox{0.84}{
\begin{threeparttable}
\caption{Comparison of outer optimizers tested on robot control tasks}
\label{supp:tab:outer}

\begin{tabular}{ccccccc}
\toprule
\multicolumn{1}{c}{\textbf{Task}} & \multicolumn{1}{c}{\textbf{AutoPSO-PSO}} & \multicolumn{1}{c}{\textbf{AutoPSO-CMA-ES}} & \multicolumn{1}{c}{\textbf{AutoPSO-GA}} & \multicolumn{1}{c}{\textbf{AutoPSO-DE}} & \multicolumn{1}{c}{\textbf{AutoPSO-CLPSO}} & \multicolumn{1}{c}{\textbf{AutoPSO-SLPSOGS}} \\
\midrule
Halfcheetah & \cellcolor{lightgray!30}\textbf{1.503E+03 (2.60E+02)} & 6.850E+02 (4.25E+02) - & 6.267E+02 (2.08E+02) - & 1.387E+03 (1.86E+02) $\approx$ & 1.163E+03 (2.04E+02) - & 1.278E+03 (2.98E+02) $\approx$ \\
Hopper & 1.347E+03 (1.07E+02) & 9.067E+02 (1.17E+02) - & 9.006E+02 (1.83E+02) - & 1.382E+03 (1.30E+02) $\approx$ & 1.390E+03 (1.32E+02) $\approx$ & \cellcolor{lightgray!30}\textbf{1.424E+03 (1.32E+02) $\approx$} \\
Pusher & -1.878E+02 (3.52E+01) & -4.341E+02 (1.90E+01) - & -4.119E+02 (2.11E+01) - & -2.029E+02 (5.03E+01) $\approx$ & -1.934E+02 (4.70E+01) $\approx$ & \cellcolor{lightgray!30}\textbf{-1.764E+02 (3.22E+01) $\approx$} \\
Reacher & -9.441E+00 (4.40E+00) & -3.562E+02 (1.76E+02) - & -1.454E+02 (6.06E+01) - & -8.999E+00 (3.95E+00) $\approx$ & -7.967E+00 (2.99E+00) $\approx$ & \cellcolor{lightgray!30}\textbf{-7.310E+00 (2.08E+00) $\approx$} \\
Swimmer & 1.911E+02 (3.41E+00) & 1.810E+02 (6.82E+00) - & 1.747E+02 (4.09E+00) - & 1.904E+02 (1.44E+00) $\approx$ & 1.916E+02 (2.04E+00) $\approx$ & \cellcolor{lightgray!30}\textbf{1.926E+02 (2.62E+00) $\approx$} \\
\bottomrule
\end{tabular}

\begin{tablenotes}[flushleft]
\footnotesize
\item[*] The Wilcoxon rank-sum tests (significance level $\alpha=0.05$) were conducted between AutoPSO-PSO and each algorithm individually.
The final row displays the number of tasks where the corresponding algorithm performs statistically better ($+$), similar ($\approx$), or worse ($-$) compared to AutoPSO.
\end{tablenotes}
\end{threeparttable}
}
\end{table*}

}
\newcommand{\suppFigBasePop}{
\begin{figure}[htbp]
   
    \centering
    \includegraphics[width=0.24\textwidth]{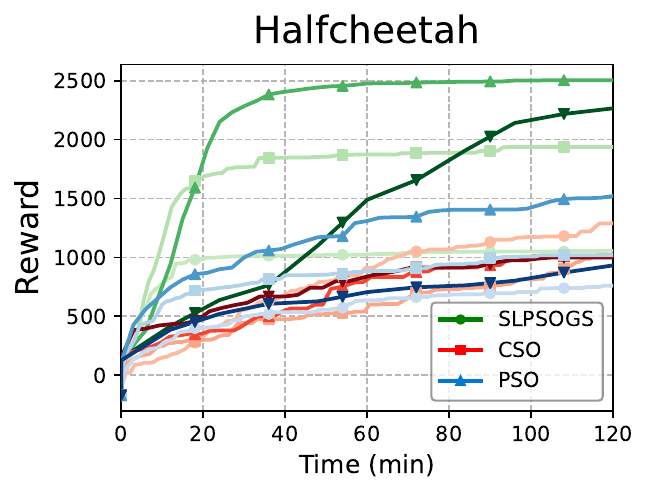}
    \hspace{-0.1cm}
    \includegraphics[width=0.24\textwidth]{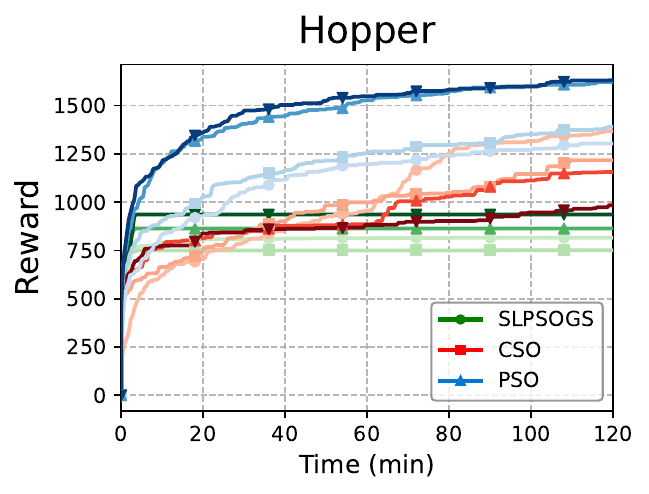}
    
    \vspace{0.05cm}
    
    \includegraphics[width=0.24\textwidth]{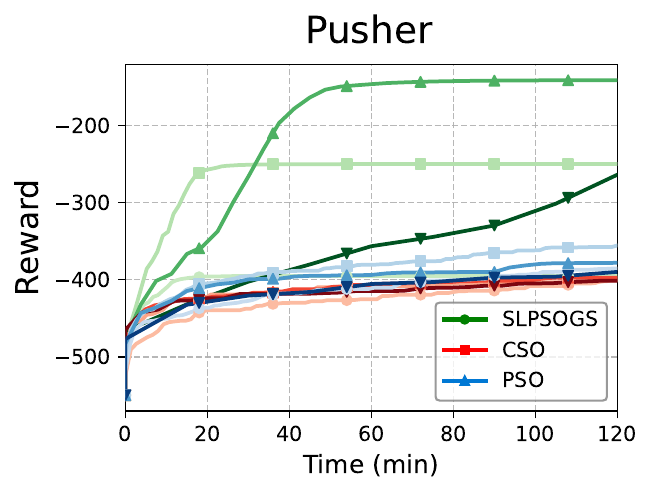}
    \hspace{-0.1cm}
    \includegraphics[width=0.24\textwidth]{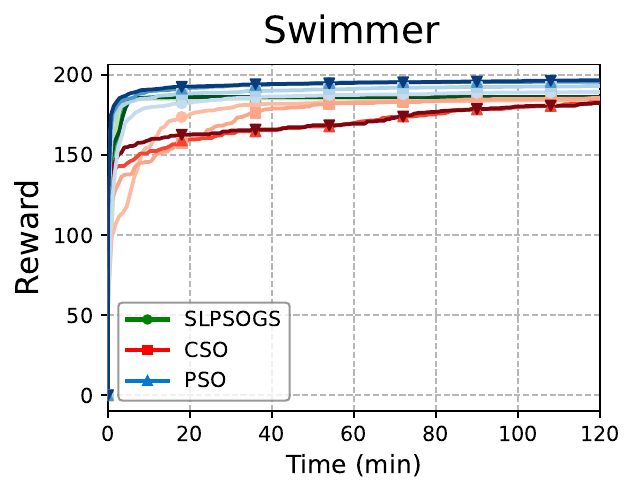}
    
    \caption{Reward curves of baseline algorithms under different population sizes on each robot control task. Each color denotes an algorithm, with lighter to darker shades indicating population sizes of 100, 1000, 5000, and 10000, respectively.}
    \label{supp:fig:base-pop}
\end{figure}

}

\begin{document}

\captionsetup{font={footnotesize}}
\captionsetup[table]{labelformat=simple, labelsep=newline, textfont=sc, justification=centering}

\title{AutoPSO: A Meta-Framework for Automated Particle Swarm Optimization
}

\author{%
    Xinmeng Yu,
    Jiaxin Gao,
    Jianguo Zhang,  
    Dongmei Jiang,
    and Ran Cheng,~\IEEEmembership{Senior Member,~IEEE}%
    \thanks{Xinmeng Yu and Jianguo Zhang are with the Department of Computer Science and Engineering, 
    Southern University of Science and Technology, Shenzhen 518055, China, 
    and also with the Pengcheng Laboratory, Shenzhen 518000, China 
    (e-mail: yxm981203@163.com; zhangjg@sustech.edu.cn).}
    \thanks{Jiaxin Gao is with the Department of Data Science and Artificial Intelligence, The Hong Kong Polytechnic University, Hong Kong SAR, China (e-mail: jiaxinn.gao@outlook.com))}
    \thanks{Dongmei Jiang is with the Pengcheng Laboratory, Shenzhen 518000, China 
    (e-mail: jiangdm@pcl.ac.cn).}
    \thanks{
    Ran Cheng is with the Department of Data Science and Artificial Intelligence, The Hong Kong Polytechnic University, Hong Kong SAR, China. He is also with The Hong Kong Polytechnic University Shenzhen Research Institute, Shenzhen, Guangdong Province, China; The Hong Kong Polytechnic University-Daya Bay Technology and Innovation Research Institute, Huizhou,Guangdong Province, China. (E-mail: ranchengcn@gmail.com) }
    \thanks{\emph{(Corresponding author: Ran Cheng.)}}%
}



\markboth{}{}\maketitle
\begin{abstract}
Particle swarm optimization (PSO) is a widely used metaheuristic, prized for its simplicity and small parameter set. Although decades of research have produced numerous PSO variants that improve performance by modifying key components (e.g., parameter schedules, swarm topologies, or updating rules), two fundamental challenges persist. First, most existing approaches are problem-specific and hand-crafted, leading to poor cross-task generalization and forcing practitioners to navigate an impractically large design space, which also hinders systematic reuse of prior effective mechanisms. Second, mainstream implementations remain CPU-bound, constraining scalability and substantially increasing computational cost in real-world applications.
To address these challenges, we propose \emph{AutoPSO}, a highly automated meta-framework for constructing customized PSO algorithms. AutoPSO formulates PSO-based optimization as a bi-level process: an outer search explores the joint space of effective PSO components, while an inner loop instantiates candidate variants to solve the target task and provide feedback. The outer search operates over a curated, open-design component pool, supporting flexible replacement of the component set and the outer optimizer. Crucially, by leveraging EvoX for population tensorization and batched evaluations, AutoPSO can efficiently assess thousands of particles within practical time budgets. Comprehensive experiments on numerical benchmarks and neuroevolution robotic control tasks demonstrate that AutoPSO consistently discovers novel PSO variants that significantly outperform strong baselines. Ablation and scalability studies further highlight the contribution of individual algorithmic components and confirm that AutoPSO achieves increasing performance gains with larger swarm sizes. 
Code is available at \url{https://github.com/EMI-Group/autopso}.
\end{abstract}
\IEEEpeerreviewmaketitle

\section{Introduction}
\IEEEPARstart{C}{omplex} optimization problems are pervasive in modern science and engineering. In many real-world scenarios, explicit gradients are unavailable or prohibitively expensive to compute, rendering gradient-based optimizers ineffective. Evolutionary algorithms (EAs)  provide a practical alternative by enabling derivative-free optimization through population-based search.
Among these techniques, PSO~\cite{Eberhart}, first proposed in 1995, has become one of the most widely adopted metaheuristics. Its conceptual simplicity and relatively few control parameters have facilitated extensive applications across diverse domains~\cite{Valle2008, Tran2019, Song2023}. Nevertheless, similar to many other evolutionary methods, canonical PSO is prone to premature convergence and may require a large number of iterations, resulting in considerable computational overhead.

Prior research has explored multiple directions to enhance the PSO algorithm. Parameter adaptation strategies, including inertia weights and constriction factors, have been introduced to improve the balance between exploration and exploitation~\cite{Shi1998, Clerc2002, Chaturvedi2009}. The diversity of learning exemplars has been increased through neighborhood topologies, swarm archives, or virtual leaders~\cite{Mendes2004, Liang2006, Ho2008}, while subpopulation partitioning has been introduced to preserve swarm diversity~\cite{Bird2006, Li2010, Xu2019}.
{These studies have provided a rich set of mechanisms for improving PSO, but their effectiveness is often problem-dependent. Mechanisms that promote fast convergence may work well on relatively smooth problems but can reduce diversity on multimodal or high-dimensional problems, whereas diversity-preserving strategies may enhance robustness but are not always the most efficient choice. Therefore, developing an effective PSO variant usually requires selecting and combining suitable mechanisms for the target problem or benchmark set.} 
{In existing studies, this design process is still largely conducted manually, relying on empirical comparison, expert knowledge, and repeated trial-and-error. Consequently, only a limited portion of possible mechanism combinations can be examined, while component interactions are difficult to evaluate systematically; once the target problem changes, the resulting variant may need to be reselected or redesigned.}

Automated algorithm design has emerged as an effective approach to overcome these limitations. Instead of relying on hand-crafted strategies, it enables algorithms to adapt automatically to the characteristics of the target problem. In evolutionary computation, this concept is often instantiated as the Meta-Black-Box Optimization (MetaBBO) framework, which formulates algorithm configuration and component composition as an outer-level optimization problem. Representative approaches employ configurators such as ParamILS~\cite{Hutter2009} and irace~\cite{LopezIbanez2016} to construct EAs, or leverage learning-based methods, including deep learning and reinforcement learning, to guide algorithm selection and component combination~\cite{Tan2021,Tan2022,Ma2024,Yin2022,Ma2023}. {For PSO, automated design is facilitated by its compact canonical update rule and the availability of reusable mechanisms developed in prior studies, which allow PSO variants to be represented within a modular and interpretable design space. Along this direction,} an early study employed genetic programming to evolve PSO velocity-update rules~\cite{Poli2005a}. Subsequent work introduced self-adaptive parameter controllers~\cite{Ziari2010} to enable dynamic adjustment across PSO instances. Grammar-based frameworks such as GEFPSO~\cite{Miranda2015} further supported the structured generation of PSO variants, while more recent component-based synthesis approaches~\cite{CamachoVillalon2022} construct PSO variants by assembling predefined operator libraries.

Despite these advances, many MetaBBO approaches rely on pretraining with large datasets and substantial computational resources. Their performance may also degrade when transferred to target tasks that differ from those seen during training, and the resulting optimizers are not always easy to interpret.
Similar challenges arise in automated PSO design. Some methods perform offline tuning, yielding fixed configurations with limited adaptability, while others optimize isolated components without considering their interactions. 
{Consequently, the resulting designs may favor specific problem types, limiting their robustness on heterogeneous benchmarks where different functions require different tradeoffs between convergence and diversity.}
Additionally, many EA implementations are CPU-bound, leading to high costs and poor scalability. These limitations motivate the development of a lightweight, online, interpretable, and adaptive framework for efficiently exploring the PSO design space, leveraging existing PSO insights to construct more flexible and effective algorithms.

To bridge these gaps, we propose {AutoPSO}, a meta-framework tailored to the PSO algorithm. Rather than pursuing a fully generalizable meta-optimizer, AutoPSO builds on established PSO research and leverages a generalized PSO skeleton to represent the design space in a transparent and structured manner. Specifically, each candidate variant is encoded as a standardized genome and instantiated through three core modules, i.e., parameter adaptation, exemplar selection, and sub-swarm partitioning. {AutoPSO then performs online meta-optimization in a two-level manner: the outer meta-optimizer searches over candidate PSO configurations, and the corresponding inner PSO instances solve the target problem to provide performance feedback for subsequent search.} This design enables both the reuse of existing PSO strategies and the construction of new variants. Furthermore, by exploiting tensorized evaluation on the GPU-accelerated EvoX platform, AutoPSO performs large-scale online meta-optimization efficiently, thereby reducing computational overhead while preserving interpretability and adaptive control over swarm behavior.

The main contributions of this work are threefold.
\begin{itemize}
\item We provide a systematic review and abstraction of recent PSO variants. Specifically, we extract a generalized PSO skeleton and consolidate key mechanisms into a reusable component library that supports modular algorithm construction.
\item We propose AutoPSO, a meta-framework that automatically assembles PSO variants by leveraging the generalized skeleton and modular component library. This approach enables both the reuse of existing strategies and the generation of novel configurations tailored to the target problem.

\item We develop a GPU-accelerated implementation built upon the EvoX platform, which enables efficient and large-scale online meta-optimization. We demonstrate its effectiveness through extensive experiments on numerical benchmarks and neuro-evolutionary robot-control tasks. In particular, we provide detailed analyses of the contributions of individual PSO components, the generalization ability of discovered variants, scalability with respect to problem size, and computational overhead.

 \end{itemize}
 
The remainder of this paper is organized as follows. Section II reviews prior research on PSO variants. Section III introduces the proposed AutoPSO method in detail. Section IV presents the experimental setup and analyzes the results. Finally, Section V concludes the paper and discusses future research directions.

\section{Related Work}

\subsection{The Canonical PSO}\label{sec:pso}

Originally proposed by Kennedy and Eberhart in 1995, PSO is a population-based optimization algorithm inspired by the collective behaviors of bird flocks and fish schools. Specifically, a swarm of $N$ particles searches in a $D$-dimensional space, where each particle maintains its position, velocity, and historical experience. In contrast to many evolutionary algorithms, PSO adopts a relatively simple search mechanism and does not rely on genetic operators such as crossover or mutation. Formally, for particle $i$ at iteration $t$, its state is represented by position $\mathbf{X}_i^t\in\mathbb{R}^D$ and velocity $\mathbf{V}_i^t\in\mathbb{R}^D$. The position update is defined as
\begin{equation}
\mathbf{X}_i^{t+1} = \mathbf{X}_i^t + \mathbf{V}_i^{t+1},
\label{eq:position}
\end{equation}
and the canonical velocity update is defined as follows
\begin{equation}
\mathbf{V}_{i}^{t+1} =
      w\mathbf{V}_{i}^{t}
      + c_{1} r_{1}^{t}\bigl(\mathbf{pbest}_{i}-\mathbf{X}_{i}^{t}\bigr)
      + c_{2} r_{2}^{t}\bigl(\mathbf{gbest}-\mathbf{X}_{i}^{t}\bigr),
\label{eq:pso-update1}
\end{equation}
where $w$ denotes the inertia weight, $c_1$ and $c_2$ are the cognitive and social acceleration coefficients, and $r_1^t,r_2^t\sim\mathcal{U}(0,1)$ are random variables. The terms $\mathbf{pbest}_i$ and $\mathbf{gbest}$ represent the personal best position of particle $i$ and the best position discovered by the swarm.

Although the canonical update equations are concise, they implicitly define key components governing PSO behavior. Control parameters $(w, c_1, c_2)$ balance exploration and exploitation, while learning exemplars (e.g., $\mathbf{pbest}$ and $\mathbf{gbest}$) shape information flow, influencing convergence and diversity. Additionally, exemplar selection depends on population structure, such as neighborhoods or sub-swarms. Together, these dimensions form the core PSO design space, providing a unified framework for constructing effective variants.

\subsection{The Development  of PSO Variants}\label{sec:adaptive-pso}

The development of PSO variants over the past decades can be understood as systematic efforts to refine the three core components of parameter control, exemplar selection, and population organization. Researchers have explored diverse strategies within each dimension, including static and adaptive parameter schemes, distinct exemplar definitions, and varied population structures such as neighborhoods and sub-swarms. These refinements aim to improve convergence speed, maintain population diversity, and enhance the ability of the algorithm to escape local optima. Consequently, PSO has evolved into a cohesive family of algorithms with broad search capabilities, thereby providing a rich foundation for systematic analysis, comparison, and further algorithmic innovation.

Specifically, one major research direction concerns {parameter control}. Early studies established that the performance of PSO is highly sensitive to the inertia and acceleration terms. The introduction of the inertia weight~\cite{Shi1998} and the constriction-factor formulation~\cite{Clerc2002} provided two influential parameterizations of the canonical update, while subsequent analyses offered recommended parameter settings from both empirical and dynamical-system viewpoints~\cite{Trelea2003}. Building on these foundations, later work moved from static parameter settings to time-dependent schedules, such as linearly decreasing inertia weights~\cite{Shi1999} and time-varying acceleration coefficients~\cite{Ratnaweera2004}. More sophisticated schemes further employed exponential, sinusoidal, stochastic, or chaotic modulation of parameters~\cite{Chen2018a,Tatsumi2015,Tian2019}, and recent studies have introduced adaptive or learning-based parameter control mechanisms driven by swarm diversity, stagnation signals, or reinforcement learning policies~\cite{Xue2020,Liu2019a,Yin2023}. Collectively, these studies confirm that regulating the strength of particle motion is a primary lever for controlling the exploration--exploitation tradeoff in PSO. At the same time, most of them remain confined to tuning a few coefficients within a pre-specified update template.

A second line of work focuses on {learning exemplars selection}.  
Classical PSO steers particle motion using the particle's historical personal best and the swarm's global best. While a dominant global leader can accelerate convergence, early analyses also cautioned that greedy guidance suppresses exploration and increases the risk of premature trapping~\cite{Shi1999}. To diversify information flow, population topologies were proposed, restricting social learning to neighborhood bests rather than a global best~\cite{Kennedy1999}. Beyond single leaders, multi-exemplar schemes aggregate peer information. For example, fully informed PSO ensures all neighbors contribute weighted influences to the velocity update~\cite{Mendes2004}. Furthermore, randomization strategies, such as stochastic leader selection~\cite{Mostaghim2003} or randomized neighborhoods~\cite{Kennedy2002}, have been used to maintain diversity.
Subsequent research introduced finer-grained exemplar designs and adaptive mechanisms. Comprehensive Learning PSO, for instance, forms a particle's search direction from dimension-wise exemplars randomly sampled from peers' $pbest$, effectively preserving diversity on multimodal landscapes~\cite{Liang2006}. Other population-statistic approaches include using a center particle that tracks the swarm centroid for guidance stabilization~\cite{Liu2007a}, and orthogonal learning, which synthesizes informative virtual reference positions via orthogonal experimental design~\cite{Ho2008}. Recognizing the distinct roles of leaders, adaptive mechanisms were further proposed to determine when and from whom a particle should learn. The aging-leader strategy periodically retires over-exploited leaders with challengers to balance exploration and convergence~\cite{Chen2013}. Separately, ensemble PSO adaptively integrates multiple PSO update strategies, thereby exploiting their complementary strengths~\cite{Lynn2017}. 

A third direction studies {swarm organization and hybrid search structures}. Multi-swarm and niching approaches were introduced to counteract the tendency of a single swarm to collapse prematurely around dominant leaders~\cite{Loevbjerg2001,Bird2006}. Hierarchical and adaptive multi-swarm models further refined this idea by allowing sub-swarms to explore at different scales, or to split and merge in response to online search states~\cite{Bergh2004,Janson2005,Yang2006}. In parallel, hybrid PSO frameworks incorporated local search, mutation, crossover, or external optimizers such as GA, DE, and CMA-ES to inject complementary search behaviors into the swarm~\cite{Liang2005a,Higashi2003,Andrews2006,Juang2004,Niu2008,Xu2019}. More recent studies have moved beyond fixed hybrid couplings toward adaptive combinations, in which the invoked strategies and their intensities are adjusted online according to search progress~\cite{Hu2013,Cao2019}. These developments implicitly treat PSO as a structured system whose performance depends on how the swarm is partitioned, coordinated, and coupled with auxiliary operators, rather than as a monolithic optimizer with a single universal form.

Taken together, the above literature suggests that effective PSO variants mainly manipulate three fundamental aspects: \emph{where particles learn from}, \emph{how strongly they are influenced}, and \emph{how the swarm is organized}. This observation is pivotal for the present work. Specifically, rather than viewing prior PSO variants as a fragmented collection of ad hoc modifications, we regard them as accumulated design knowledge regarding these three controllable dimensions.

\renewcommand{\arraystretch}{1.5}
\setlength{\tabcolsep}{3pt}

\subsection{Meta-Black-Box Optimization}
\label{sec:metabbo}
Despite decades of research on PSO and other black-box optimization methods, their performance remains highly dependent on manually designed heuristics and parameter tuning. As problem scale and complexity increase, manually configuring algorithm components and parameters becomes time-consuming and may fail to generalize to unseen tasks. To reduce reliance on expert knowledge and improve generalization, automated algorithm design (AAD) has emerged as a promising research direction.
AAD formalizes the search for effective algorithms by defining a structured design space and systematically exploring it. Early works employed exhaustive search strategies~\cite{Rice1976}, while algorithm portfolios and selection frameworks demonstrated the benefit of combining multiple candidate algorithms~\cite{Gomes2001}. Model-based and sequential optimization approaches further enhanced automated configuration~\cite{Hutter2009a,Hutter2011}, and hyper-heuristics or adaptive operator selection have been applied to tune algorithm components~\cite{Cowling2002,DaCosta2008}. Subsequent methods leverage automated parameter optimization and Bayesian strategies to efficiently explore the design space~\cite{Bergstra2011,Burke2013}. 

In the context of black-box optimization, this principle is realized in a Meta-Black-Box Optimization (MetaBBO) framework~\cite{Ma2025}. Here, the term “meta” refers to a higher-level strategy that governs the design of the lower-level optimizer. Specifically, MetaBBO formulates algorithm design as a bi-level problem: the lower-level optimizer performs the actual search and returns performance feedback, while a meta-level policy uses this feedback to make algorithmic decisions, such as parameter adjustments, operator selection, or algorithm combination, often leveraging information across multiple tasks~\cite{Mercer1978, Fogel1991,Stephenson2003a,Burke2013}. By treating algorithm configuration as a learnable meta-level problem, MetaBBO automatically discover optimizers that can adapt or generalize across different problem instances.

Recent MetaBBO methods exhibit diverse design strategies. One line leverages reinforcement learning (RL) to control the lower-level optimizer, including frameworks such as MetaBox~\cite{Ma2023}, RLEPSO~\cite{Yin2022}, and RLHPSDE~\cite{Tan2022}. These methods model the lower-level control as a Markov decision process, allowing RL agents to dynamically balance exploration and exploitation and adapt to different problem instances. Their advantage lies in online adaptability and improved generalization, but they incur high training costs, offer limited interpretability, and may face convergence challenges in high-dimensional or complex landscapes. Another approach employs neural networks or autoregressive models, such as DesignX~\cite{Guo2025} and the Autoregressive Metaheuristic Designer~\cite{Zhao2025a}, to generate optimizers as sequences, enabling discovery of nontrivial algorithmic patterns and end-to-end automation. While powerful, these methods require extensive pretraining and may not generalize reliably to extreme tasks.

Other studies focus on algorithm component composition and generation. Frameworks such as GSF~\cite{Yi2023} represent low-level operators uniformly and generate new algorithms via RL or neuroevolution, offering advantages for task-specific optimizer discovery but limited by predefined module libraries. Symbolic equation-based methods like SYMBOL~\cite{Chen2024} dynamically generate closed-form optimization rules with RL, achieving high interpretability and zero-shot generalization, but training becomes challenging in high-dimensional problems. Approaches like Auto-WEKA~\cite{Thornton2013} and Programming by Optimization (PbO)~\cite{Hoos2012} perform algorithm selection and hyperparameter search within a design space, suitable for rapid construction of task-adaptive optimizers.

The present work takes a complementary perspective by focusing on automated design within a single optimizer family. Rather than generating entirely new optimizers across heterogeneous paradigms, we concentrate on PSO and explicitly organize the accumulated design knowledge of PSO variants into a modular and searchable design space. By performing online configuration within this space, the proposed framework automatically constructs task-adapted PSO variants using feedback from the target problem itself. In this way, the method bridges classical PSO research and automated algorithm design: it retains the interpretability and structural insights of swarm-based optimization while leveraging systematic search to discover effective combinations of mechanisms. The resulting framework provides a practical pathway for transforming manually engineered PSO variants into an automatically configurable optimization system.

\section{Method}

This work is designed to achieve three main objectives: 
(i) organizing existing PSO mechanisms into a systematic and reusable library, 
(ii) developing an online method for automatically synthesizing high-performing PSO variants by exploring combinations of these mechanisms, thereby reducing manual design effort, and 
(iii) conducting a systematic analysis of the interactions among PSO components to better understand their impact on search behavior.
To this end, we first abstract a generalized PSO skeleton and augments it with mechanisms drawn from a curated library distilled from existing PSO studies. These mechanisms are organized into a modular configuration space, within which candidate PSO variants are automatically instantiated and evaluated by an outer-level optimization process. Instead of manually designing a new variant for each task, the proposed AutoPSO searches over this structured space of swarm update policies to identify effective combinations for the problem at hand while simultaneously solving it.
The remainder of this section introduces the generalized PSO template and the meta-level architecture of AutoPSO.

\subsection{Generalized PSO Skeleton and Design Choices}\label{sec:gPSO}
The diversity among PSO variants largely stems from differences in how information propagates through the swarm and how particles update their positions. Designing a PSO-based algorithm therefore requires consideration of three key aspects: which exemplars each particle learns from, how strongly particles react to selected exemplars, and how the swarm is organized (e.g., partitioned into subpopulations or update groups).
In this work, we formalize these aspects as three configurable modules: \emph{population splitting}, \emph{parameter control}, and \emph{exemplar assignment}. Together, they define a generalized PSO skeleton (gPSO) that captures a wide range of PSO-style search behaviors while remaining sufficiently compact to support automated configuration and systematic exploration of the algorithm design space.

\subsubsection{Population Splitting}
Maintaining diversity is crucial for avoiding premature convergence, especially on rugged or multimodal landscapes.

Instead of creating many fully isolated sub-swarms, AutoPSO adopts a dual-policy population split as a pragmatic compromise between behavioral expressiveness and configuration complexity.
At each iteration, AutoPSO uses a ratio \(r \in [0,1]\) to determine the proportion of particles updated by two different policies.
Specifically, approximately \(\lfloor rN \rfloor\) particles are updated by the first policy, while the remaining particles are updated by the second, where \(N\) is the population size.
Therefore, \(r\) controls the coverage of the two update rules within a shared population, rather than defining two fully independent swarms.

The purpose of the split is to introduce behavioral heterogeneity rather than to isolate information flow. All exemplar candidates, including global and local bests, are computed from the full population, ensuring that even small subgroups can access high-quality social information. This shared-exemplar design mitigates the inefficiency of small sub-swarms in limited inner-loop iterations, preserves reliable social guidance, and accelerates convergence. Meanwhile, the heterogeneous update policies induce diverse search behaviors across the population, maintaining exploratory coverage.
Representing the split via a single continuous ratio \(r\) keeps the configuration space compact and amenable to outer-loop optimization.
As a result, AutoPSO can flexibly regulate the relative prevalence of different search behaviors without introducing the additional complexity and instability often associated with fully independent multi-swarm structures.

\subsubsection{Parameter Control}
The motion of each particle in PSO is governed by an inertia term and two learning terms. In classical PSO, the inertia weight \(w\) regulates velocity persistence, while the acceleration coefficients control the strength of attraction toward reference positions. Early studies have shown that the inertia weight strongly influences the exploration–exploitation trade-off, with commonly used settings including fixed values around \(0.8\) or \(0.9\) or linearly decreasing schedules~\cite{Shi1998a, VandenBergh2006}. Stability analyses further indicate that large parameter values can lead to oscillatory or divergent particle trajectories, whereas moderate values are generally associated with more stable swarm behavior~\cite{Clerc2002,Trelea2003}.

In AutoPSO, each update policy uses one inertia weight and two learning coefficients.
Since two update policies are maintained simultaneously, the framework searches over one shared inertia term and four learning coefficients in total:
\((w, c_1, c_2, c_3, c_4)\).
Rather than optimizing over unconstrained coefficients, AutoPSO restricts all control parameters to bounded and practically stable ranges.
In the current implementation, the inertia term is searched within \([0,1]\), and all learning coefficients are also restricted to \([0,1]\).
These bounds are informed by classical PSO parameter selection and stability analyses, which indicate that moderate velocity multipliers and attraction strengths help avoid pathological updates and enhance swarm robustness~\cite{Shi1998a,Clerc2002}, as well as by insights gained from our empirical tuning. Chosen as conservative ranges, they reduce the risk of unstable updates during automated exploration and keep the outer-level optimization compact and numerically well-conditioned.

\subsubsection{Exemplar Assignment}
Exemplar assignment specifies the reference positions that guide particle updates. In canonical PSO, these references are typically limited to the particle's personal best and the global best. AutoPSO extends this mechanism by introducing a library of nine exemplar candidates, from which the outer-level optimizer selects configurations online. These candidates play different roles in balancing memory retention, social learning, and stochastic exploration, and can be organized into three categories:
\emph{self-based}, \emph{population-derived}, and \emph{randomized} exemplars, which based on the dominant source of information used to construct the exemplar.

\emph{Self-based exemplars} preserve individual experience and support memory-driven search:
\begin{itemize}
    \item \texttt{current}: sustains the particle's present position, serving as a null update that provides stability in dynamic or noisy environments.
    \item \texttt{pbest}: attracts the particle to its historical optima, preserving successful individual search directions.
\end{itemize}

\emph{Population-derived exemplars} capture collective intelligence to coordinate search: 
\begin{itemize}
    \item \texttt{gbest}: denotes the current global best position in the whole population, promoting strong convergence pressure;
    \item \texttt{center}: the centroid of the entire population, providing a distributed social signal that is less dependent on a single elite particle;
    
    \item \texttt{tournament-pbest}: a learned personal-best exemplar constructed by first selecting the better one from two randomly sampled personal-best candidates and then probabilistically mixing it with the particle's own personal best. This mechanism introduces selective information transfer while preserving individualized learning behavior.
\end{itemize}

\emph{Randomized exemplars} inject controlled randomness, preventing premature convergence and expanding the search frontier:
\begin{itemize}
    \item \texttt{new}: an exemplar given by a newly generated particle, sampled uniformly from the bounded search space rather than derived from perturbing an existing one.
    \item \texttt{random-x}: the current position of a uniformly sampled particle from the whole population. This neutral selection pressure helps maintain population diversity without directional bias;
    \item \texttt{random-pbest}: randomly selects from the whole set of personal bests, encouraging cross-fertilization of successful trajectories while maintaining stochasticity;
    \item \texttt{random-elite}: a position sampled from the top-performing fraction of the current population, adding quality-guided randomness that emphasizes promising regions.
\end{itemize}

At each iteration, AutoPSO assigns two exemplars to each of the two update groups. Specifically, denoting the selected exemplars as \((\mathbf{p}_1,\mathbf{p}_2,\mathbf{p}_3,\mathbf{p}_4)\), the first group uses \((\mathbf{p}_1,\mathbf{p}_2)\) with coefficients \((c_1,c_2)\), and the second group uses \((\mathbf{p}_3,\mathbf{p}_4)\) with \((c_3,c_4)\). By computing exemplars globally but applying them differently across groups, AutoPSO induces heterogeneous search behaviors from a shared information base.

As a result, the configuration space is nontrivial even before considering parameter values. Specifically, for a single update policy, selecting two exemplar channels from a library of nine candidates yields \(9^2=81\) ordered assignments. Furthermore, with two update policies, AutoPSO searches over two such assignments simultaneously, alongside the population split ratio and policy coefficients. This formulation creates a structured space of PSO variants, which enables the outer optimizer to discover effective combinations tailored to the optimization task.

\figAutoAlgo

\subsection{The Architecture of AutoPSO}

AutoPSO formulates PSO design as a black-box optimization problem over the configuration space induced by the generalized PSO template.
As illustrated in Fig.~\ref{fig:autoAlgo}, the framework separates {configuration discovery} from {problem solving} through a two-layer architecture.
The outer layer searches for effective PSO configurations, while the inner layer instantiates and executes the corresponding PSO variants on the target optimization task.
In our experiments, the outer optimizer is also implemented using PSO, although the framework itself is not restricted to a specific meta-optimizer.

Each individual in the outer population encodes a complete specification of one gPSO instance.
In the current implementation, this specification includes the population split ratio, the coefficients associated with the two update policies, and the exemplar assignments used by the four learning channels.
Once decoded, a candidate configuration defines a concrete inner PSO variant described in Section~\ref{sec:gPSO}.
During evaluation, each configuration is instantiated as an inner PSO solver and executed on the target problem for a fixed computational budget.
The best fitness value achieved by the inner run is then used as the quality indicator of that configuration.

Formally, let \(\theta \in \Theta\) denote a candidate configuration and let \(\mathcal{A}(\theta)\) denote the instantiated inner PSO induced by \(\theta\).
The outer-layer search can be written as
\[
\theta^\star = \arg\min_{\theta \in \Theta} \mathcal{L}(\mathcal{A}(\theta); f),
\]
where \(f\) is the target objective function and \(\mathcal{L}\) measures the optimization performance of the instantiated solver under a prescribed inner evaluation budget.
Since \(\mathcal{L}\) is generally nonconvex and non-differentiable, AutoPSO treats the configuration process as a black-box search problem and optimizes it using an outer optimizer.

This design provides a unified mechanism for exploring PSO variants without manual trial-and-error.
By recombining the modular components introduced in Section~\ref{sec:gPSO}, AutoPSO can recover or approximate a broad family of PSO-style search behaviors and, at the same time, discover novel configurations tailored to the target problem.
The configuration space is restricted to bounded and practically stable parameter domains, which improves the numerical robustness of the outer search and prevents pathological inner dynamics caused by extreme coefficient settings.
Although the current implementation focuses on static policy coefficients and a two-group update structure, the framework is extensible and can be generalized to incorporate richer parameter controllers and more sophisticated swarm organizations in future work.
Algorithm~\ref{alg:gPSO} and Algorithm~\ref{alg:autopso} summarize the gPSO template and the outer-layer optimization procedure, respectively.

\algoGPSO

\algoAutoPSO

\section{Experimental Study}

To evaluate the effectiveness of AutoPSO as an online optimization method, we compare it with six representative PSO variants: Competitive Swarm Optimizer (CSO)~\cite{Cheng2015}, Comprehensive Learning PSO (CLPSO)~\cite{Liang2006}, Fully Informed Particle Swarm (FIPS)~\cite{Mendes2004}, Social Learning PSO using Gaussian Sampling (SLPSOGS)~\cite{Cheng2015a}, Social Learning PSO using Uniform Sampling (SLPSOUS)~\cite{Cheng2015a}, and the canonical PSO. In addition, to assess the benefits of the meta-level optimization framework, we also compare AutoPSO with several recent MetaBBO approaches, including DEDQN~\cite{Tan2021}, SYMBOL~\cite{Chen2024}, GLEET~\cite{Ma2024}, RLHPSDE~\cite{Tan2022}, and RLEPSO~\cite{Yin2022}.
The benchmark suite comprises both numerical functions and practical robotic control tasks. Numerical experiments include the CEC2022 benchmark suite and several standard analytic functions such as Ackley and Rastrigin, tested across multiple dimensions with rotated and shifted versions to evaluate scalability. For practical scenarios, we consider five policy search tasks implemented in the Brax physics engine~\cite{Freeman2021}. These tasks feature high-dimensional state–action spaces and nonlinear stochastic dynamics, providing a realistic complement to the numerical benchmarks.
All algorithms are implemented using the EvoX library~\cite{Huang2024} and executed on a workstation equipped with an Intel Core i9-10900X CPU and an NVIDIA RTX 3090 GPU.

Most experiments in this work are conducted using a fixed wall-clock time budget. On modern GPUs, the FE budgets commonly used in prior studies (e.g., $10^5$--$10^6$) can often be exhausted within a very short time. As a result, strict FE limits may impose an overly restrictive evaluation budget and may not fully reflect algorithmic performance in a highly parallel setting. So we set a practical time window that allows all algorithms to perform a sufficient number of evaluations and fully utilize their search potential. Since all methods share the same implementation framework, GPU acceleration, and control flow within the \texttt{EvoX} toolkit, running each algorithm on identical hardware under the same time constraints ensures fair and consistent comparisons. For completeness, we also provide supplementary experiments under equal-FE settings to isolate optimization efficiency from execution speed.

\subsection{Performance on CEC2022 Benchmarks} \label{sec:numerical}
\subsubsection{Comparison under Equal Time}
We evaluate AutoPSO against six baseline PSO variants on the 10D and 20D functions from the CEC2022 benchmark suite, which comprises 12 problems spanning basic ($F_1$–$F_5$), hybrid ($F_6$–$F_8$), and composition ($F_9$–$F_{12}$) functions. Each run is limited to 60 seconds for 10D and 120 seconds for 20D functions, with 31 independent trials per problem.
Fig.~\ref{fig:cec022-20d} shows the convergence behavior for the 20D functions. Most baseline PSO variants exhibit rapid initial error reduction followed by prolonged stagnation, which is primarily caused by swarm contraction: as particle diversity decreases, velocity updates shrink, limiting exploration and causing the swarm to become trapped in local basins. Some variants, such as CSO and CLPSO, partially mitigate early stagnation through diversity-preserving mechanisms. CSO applies loser updates via pairwise competition, while CLPSO uses dimension-wise comprehensive learning to leverage multiple exemplars. These strategies improve performance on certain functions, however, as the swarm contracts, update magnitudes decrease, leading to marginal gains and stagnation on more rugged landscapes, such as $F_6$ and $F_{10}$.

In contrast, AutoPSO sustains progress throughout the runtime. Its two-level framework continuously adapts the configuration of the inner PSO instance based on feedback from the outer-level meta-optimizer, dynamically adjusting parameters, exemplar assignments, and swarm partitioning according to the current problem landscape. Conventional PSO variants effectively operate as static instances of a single update policy, where control parameters and exemplar selections are either fixed or adjusted only based on short-term local feedback from recent iterations. By coordinating adaptations across the outer and inner levels, AutoPSO mitigates contraction-driven stagnation, maintains particle diversity, and preserves effective exploration. 

\figCECfixTime

\subsubsection{Comparison under Equal FEs}~\label{sec:fes}
To separate optimization effectiveness from execution speed, we also compare all algorithms under an identical FE budget. For each benchmark, the FE limit is set to match the number of evaluations AutoPSO completes during the equal-time experiment. Each test was run 11 times to account for stochastic variability.
As reported in Table~\ref{tab:CEC-FE}, AutoPSO consistently achieves superior performance under the same FE budget, which corroborates the equal-time observations and suggests that its advantage is not merely due to faster execution.
A key observation is that additional evaluations do not necessarily yield better solutions for many PSO variants. Once stagnation occurs, further FEs are mainly spent on local refinements with minimal improvement, especially on rugged or multimodal landscapes.  
AutoPSO, however, allocates evaluations through its meta-framework. Part of the FE budget is used to optimize the outer-level configuration, while the inner-level PSO executes with a problem-specific setup. By tailoring the inner instance to the current landscape, each evaluation becomes more informative, enabling sustained progress under a fixed FE budget. This demonstrates that AutoPSO’s advantage stems from more effective evaluation use rather than faster execution.

\tabCECFE

\subsection{Performance on Large-scale Benchmarks}~\label{sec:large benchmark}

To evaluate scalability and robustness, we further examine on high-dimensional multimodal functions and their shifted and rotated variants. Table~\ref{tab:basic} reports results on the Ackley and Rastrigin functions with dimensions from 50 to 2000. The runtime budget is 30 seconds for 50--200D and 300 seconds for 500--2000D, with 31 independent runs per setting. 
As dimensionality increases, algorithms experience growing errors.  Variants with effective diversity-preserving mechanisms such as CLPSO and CSO maintain reasonable performance in moderate dimensions but degrade in extremely high-dimensional problems. In contrast, AutoPSO consistently achieves the lowest errors across all dimensions. In lower-dimensional settings, it converges rapidly to near-optimal values. At higher dimensions, the outer-level optimizer dynamically adjusts inner-level strategies to tailor the search to the problem landscape, preventing the performance degradation observed in fixed-strategy baselines. This enables AutoPSO to scale effectively with dimensionality while preserving high solution quality. 
Fig.~\ref{fig:ackley} presents the convergence curves on shifted and rotated Ackley functions for 50D, 100D, and 200D. AutoPSO exhibits faster and more stable convergence than all baselines, which often plateau or slow significantly after initial progress. Even as dimensionality increases to 200D, AutoPSO maintains a clear advantage, demonstrating its scalability and robustness on large-scale, transformed search spaces.

\tabBasic

\figAckley

\subsection{Performance on Robotic Control Tasks}

We further evaluate AutoPSO on robotic control tasks, where evolutionary algorithms are used to optimize neural network policy parameters. These problems are high-dimensional and computationally demanding, providing a challenging setting for assessing scalability in practical decision-making tasks. Experiments are conducted using the Brax simulator~\cite{Freeman2021} on five standard environments. All tasks share the same policy architecture, a three-layer fully connected neural network with task-specific input and output dimensions. Each experiment runs for 60 minutes and is repeated 11 times.
Fig.~\ref{fig:brax} shows the training curves across all tasks. AutoPSO consistently achieves faster reward improvement and higher final performance than the baseline methods. In particular, the curves of AutoPSO continue to improve in later stages, indicating sustained optimization capability rather than early stagnation.
A notable feature of AutoPSO is its rapid reward growth during the early stage of training. This behavior stems from its parallel search mechanism, where multiple PSO configurations are evaluated simultaneously in the outer layer. Promising configurations can therefore be identified early, enabling the inner optimization processes to quickly produce competitive policies.
In contrast, baseline methods rely on single-instance searches with fixed update rules, such strategies often suffer from search bias and stagnation once the swarm converges. By coordinating multiple PSO instances within a two-layer framework, AutoPSO introduces additional search diversity and exploration breadth, leading to faster early progress and improved final policy quality.

\figBrax

\subsection{Comparison with MetaBBO Methods}
To assess AutoPSO as a meta-optimizer, we compare it with recent MetaBBO approaches, including DEDQN~\cite{Tan2021}, SYMBOL~\cite{Chen2024}, GLEET~\cite{Ma2024}, RLHPSDE~\cite{Tan2022}, and RLEPSO~\cite{Yin2022}. DEDQN and RLHPSDE adapt DE parameters and mutation strategies via reinforcement learning, RLEPSO extends PSO with RL-guided social learning, GLEET learns exploration–exploitation trade-offs through deep RL, and SYMBOL generates candidate optimizers via symbolic equation learning. All methods were evaluated on 12 functions from CEC2022-20D benchmark under an equal FE budget of $2.48 \times 10^{7}$, with 11 independent runs per function. MetaBBO baselines used pretraining on the same problem set, whereas AutoPSO configures PSO components online without prior training.

Overall, AutoPSO demonstrates competitive or superior performance across most functions, including $F_7$, $F_{10}$, and $F_{11}$, where it achieves the lowest mean objective values. This advantage arises from its problem-specific online configuration, which adapts PSO components  to each function’s structural characteristics, enabling tailored exploration and exploitation. In contrast, MetaBBO methods rely on pre-trained strategies that generalize across functions. While this supports cross-task transfer, fixed learned policies may not fully match the landscape of individual complex functions, limiting performance.
Nonetheless, on certain functions, such as $F_1$ and $F_6$, RL-based MetaBBO methods like RLEPSO and RLHPSDE achieve higher optimization precision. These methods leverage deep RL (e.g. deep Q-learning) to dynamically balance exploration and exploitation, rapidly focusing search on promising regions. In contrast, AutoPSO introduces stochasticity in both outer- and inner-level swarm searches, which can result in slower convergence under a limited function evaluation budget.

Beyond raw performance, AutoPSO offers practical advantages. It requires no pre-collected training data or costly model pretraining, reducing computational overhead. Its modular design provides interpretability, allowing explicit analysis of the contributions of exemplars, sub-swarm splits, and parameter settings to search efficiency—contrasting with the black-box nature of many MetaBBO approaches. The main limitation lies in the restricted search space: AutoPSO explores only PSO-based structures, whereas some MetaBBO methods can generate a wider variety of optimizer forms and learn reusable strategies that generalize across multiple tasks. 
In summary, AutoPSO offers a lightweight, interpretable, and training-free approach that achieves strong task-specific performance within the PSO design space, providing a competitive and resource-efficient alternative to general-purpose MetaBBO methods.

\tabAad

\subsection{Generalization Ability of Discovered PSO Variants}
Since the outer-layer optimizer in AutoPSO constructs PSO variants automatically, we further examine the generalization ability of the discovered configurations through transfer experiments across different functions, dimensions, and tasks. Specifically, variants discovered on the CEC2022-20D functions $F_6$, $F_8$, $F_{10}$, and $F_{12}$ are applied to the remaining functions in the same benchmark. In addition, variants obtained from 20D $F_6$ and 200D Schwefel are transferred between the two problems, and these configurations are further evaluated on the \emph{Reacher} task. All experiments are repeated 11 times.
As reported in Table~\ref{tab:generalize}, variants discovered on one function generally remain competitive when applied to other functions with the same dimensionality, indicating that the learned update rules capture useful search behaviors. However, cross-dimensional transfer is substantially more challenging. As shown in Fig.~\ref{fig:generalize}, configurations optimized for 20D problems often stagnate when directly applied to 200D tasks. This observation is consistent with the No Free Lunch principle~\cite{Wolpert1997}, which states that no single configuration performs optimally across all problems.
Interestingly, the variant discovered on $F_6$ also performs well on the \emph{Reacher} task. As $F_6$ is a challenging problem in CEC2022, the configuration discovered on this problem likely learns robust search behaviors, which remain effective when optimizing neural network policies in the Reacher task.
In fact, the key strength of AutoPSO lies in its ability to discover problem-specific algorithm configurations dynamically. Since no single variant is universally optimal, tailoring the algorithm to each target problem is essential for achieving high performance across diverse tasks.

\figGeneralize

\tabGeneralize

\subsection{Mechanism Analysis and Ablation Study}

\subsection{Effect of Algorithm Components}

To evaluate the contribution of individual components, we conduct ablation studies on CEC2022-20D under the same setting as Section~\ref{sec:numerical}. Variants include disabling the multi-swarm partition (AutoPSO-m), fixing $w$, $c_1$, and $c_2$ (AutoPSO-c), restricting exemplar selection to personal-best and global-best only (AutoPSO-e), and finer-grained modifications such as removing acceleration coefficients (w/o. acc.), inertia weight (w/o. weight), or specific exemplar types (w/o. random, w/o. self, w/o. social). Results are summarized in Table~\ref{tab:component}.

Overall, the complete AutoPSO consistently achieves the best performance, while removing individual components leads to varying degrees of degradation. Among the ablations, disabling adaptive parameter control (AutoPSO-c) results in the largest performance drop, particularly on multimodal and composition functions such as $F_9$ and $F_{12}$. This demonstrates that dynamic adjustment of acceleration coefficients and inertia weight is crucial for regulating swarm behavior, maintaining an effective exploration–exploitation balance. Fixing either the acceleration coefficients or the inertia weight alone also reduces performance, indicating that these parameters jointly influence swarm dynamics, with acceleration coefficients exerting a slightly stronger effect.
The exemplar mechanism further enhances robustness. Restricting exemplars to only personal-best and global-best (AutoPSO-e) significantly reduces performance on multimodal functions ($F_2$, $F_6$, $F_{12}$), emphasizing the importance of diverse exemplars for escaping local optima. Analysis of finer-grained variants shows that removing random exemplars impairs exploration substantially, whereas removing only self or social exemplars has a milder impact, suggesting that random exemplars play a key role in maintaining population diversity.
Finally, removing the multi-swarm mechanism (AutoPSO-m) also degrades performance, confirming that swarm subdivision enhances population diversity and search effectiveness.
Despite a larger outer search space, the full AutoPSO sustains competitive performance under the same time budget, demonstrating the efficacy of its core components.

\tabComponent

\subsection{Impact of the Number of Sub-swarms}
We further examine the design of using two sub-swarms in AutoPSO by comparing several partition strategies on CEC2022-20D, including no partition, a fixed-ratio bipartition (0.5/0.5), and configurations with three or four sub-swarms. All experiments are executed within 120 seconds, with 11 independent runs each.
As shown in Table~\ref{tab:subswarm}, introducing sub-swarms generally improves performance over a single-swarm baseline by enabling heterogeneous update behaviors that enhance behavioral diversity. Allowing the partition ratio to be adapted via the outer optimizer further improves results, indicating that the optimal balance depends on problem characteristics: simpler landscapes favor centralized search, whereas multimodal or composite functions benefit from diversified subgroups.

In our design, exemplars are computed from the full population, ensuring all subgroups receive high-quality social guidance while following distinct update policies. This allows multiple sub-swarms, regardless of size, to increase diversity without constraining the search capability of individual groups.
However, increasing the number of sub-swarms beyond two does not consistently improve performance and can degrade results on some functions (e.g., $F_6$ and $F_{12}$). This is likely because additional sub-swarms expand the outer-level configuration space, making it more challenging for the meta-optimizer to identify effective setups within a fixed computational budget. More sub-swarms also reduce the number of particles per group, weakening the internal update mechanism and dispersing search trajectories. Moreover, parallel sub-swarms may generate conflicting directions or redundant exploration. Overall, a bipartite swarm provides a practical balance between search diversity and coordinated behavior.

\tabSubswarm

\subsubsection{Choice of the Outer Optimizer} 
Beyond the flexible design of the inner components, AutoPSO also allows the outer optimizer, which explores the space of PSO modules, to be replaced seamlessly. To evaluate the effect of this design choice, we instantiate AutoPSO with six different optimizers as the outer search engine and compare their performance. Each variant is run for 120 minutes on five robotic control benchmarks, with 11 independent trials per task. As shown in Fig.~\ref{fig:outer}, the choice of the outer optimizer has a clear impact on the optimization behavior. Among the tested methods, PSO-family optimizers generally achieve faster reward improvement in the early stage and remain among the best-performing methods throughout the optimization process. This advantage is particularly evident on \emph{HalfCheetah}. By contrast, GA and CMA-ES usually progress more slowly and remain less competitive overall. DE, in comparison, exhibits competitive performance on most tasks and is overall comparable to the PSO-based variants.

This difference is closely related to the structure of the outer optimization problem. In our framework, the outer search is conducted in a compact low-dimensional space (10D), while part of the decision vector, namely the exemplar indices, is discretized by rounding during evaluation, which induces a non-smooth objective landscape. Such a setting favors optimizers that can search effectively in low-dimensional spaces while remaining robust to discontinuities. PSO-based methods are particularly well suited to this setting. Their search is guided by the cooperation between personal-best and global-best information, which enables effective exploration and exploitation without relying on local smoothness. This makes them robust to the non-smoothness introduced by rounding. CMA-ES, by contrast, depends more heavily on smooth local fitness variations for covariance adaptation and is thus less effective when discontinuities are present. GA relies mainly on crossover and mutation, and therefore provides weaker local refinement in this compact search space. DE remains competitive, as its differential search balances exploration and exploitation and tolerates discretization.

\figOuter

\subsection{Computational Cost and Scalability Analysis}
\subsubsection{Scaling with Population Size}
Population-based algorithms are naturally suited to parallel hardware, since individuals can be processed concurrently. In AutoPSO, this property is further strengthened by the hierarchical design: the outer level evolves independent PSO instances, while each outer individual maintains an inner swarm. This structure enables simultaneous scaling at both levels. To evaluate its effect, we test different inner and outer population configurations on the 200D functions using an RTX 3090 GPU. Fig.~\ref{fig:scaling-popsize} reports the runtime and fitness over 10 iterations, averaged over 11 independent runs.
As shown in the left panel, runtime increases with population size, but the growth is markedly sublinear. In particular, a 100$\times$ increase in the total population results in about a 3$\times$ increase in runtime. This trend indicates that the enlarged workload is effectively absorbed by GPU parallelism, so the additional computational cost is substantially amortized. More importantly, larger populations consistently yield better solution quality. For a fixed total population, assigning more individuals to the outer level generally leads to better performance on Ackley and Sphere.

\figScalingPopsize

We further examine population scaling on robotic control tasks under a 120-minute evaluation budget. Fig.~\ref{fig:popsize} shows the reward trajectories of seven population configurations over 11 independent runs. A consistent pattern can be observed: larger populations achieve faster reward improvement and higher final performance within the same wall-clock budget.
Increasing the outer population produces a pronounced gains in the early stage, as more concurrent PSO instances improve exploration while largely preserving iteration speed. In contrast, enlarging the inner swarm mainly enhances within-instance diversity, but also increases the per-step computational cost and may therefore slow early progress.
By coupling instance-level and within-instance parallelism, AutoPSO converts hardware parallelism into tangible optimization gains while maintaining favorable scalability.

\figPopsize

\subsubsection{Scaling with Problem Dimension}
Section~\ref{sec:large benchmark} has demonstrated the optimization performance of AutoPSO on high-dimensional problems. Here, we further evaluate its computational scalability with respect to problem dimension. Specifically, we measure runtime while increasing the dimension from 16 to 8192 on the Ackley, Griewank, and Sphere functions. The outer and inner population sizes are both set to 100, with each configuration run for 10 iterations and repeated 11 times.
The left panel of Fig.~\ref{fig:scaling} shows runtime trends on an RTX 3090 GPU. All three functions are standard numerical benchmarks with comparable evaluation costs, resulting in similar runtime curves. As expected, computational cost increases with dimensionality, since both particle updates and fitness evaluations scale with the number of decision variables. Even at 8192 dimensions, the runtime remains around 250 s, demonstrating practical feasibility for large-scale optimization.
This efficiency stems from AutoPSO’s hierarchical design implemented in EvoX, which allows full utilization of GPU parallelism. The two-level structure maps population-based operations effectively onto parallel hardware, ensuring high device occupancy and minimizing overhead. The right panel of Fig.~\ref{fig:scaling} compares GPU and CPU (Intel Core i9-10900X) runtimes. While runtime increases with dimensionality on both platforms, the CPU exhibits rapid growth due to limited concurrency, whereas the GPU maintains moderate growth by exploiting massive parallelism. At the highest dimensions, the GPU achieves orders-of-magnitude speedup, highlighting the practical advantage of mapping AutoPSO’s design to highly parallel hardware and improving its applicability to large-scale optimization.

\figScaling

\section{Conclusion}

This paper proposed AutoPSO, a meta-framework for automated PSO design based on a generalized PSO skeleton and a modular component library. By organizing representative PSO mechanisms into three configurable modules, AutoPSO provides an interpretable design space for assembling task-adaptive PSO variants and searches this space through online meta-optimization.
Extensive experiments on numerical benchmarks and neuro-evolutionary control tasks demonstrate that AutoPSO consistently outperforms strong PSO variants and recent MetaBBO methods, often with statistically significant improvements. 
{AutoPSO is particularly useful for optimization scenarios in which PSO is a reasonable candidate solver, but a single manually selected variant may not provide robust performance across the target problems. Since it introduces an additional outer-level search, its benefit is more likely to justify the cost when sufficient function-evaluation budget or parallel computing resources are available. For simple problems or severely restricted budgets, directly applying a standard PSO variant may remain more practical.}
AutoPSO also provides a flexible basis for future research. Its modular formulation can be extended to support finer-grained update components, more diverse outer-loop optimizers, and hybridization with mechanisms from other evolutionary paradigms. {Future work will further improve the efficiency of the meta-search process and extend the framework to broader optimization scenarios while preserving its modularity and interpretability.}

\bibliographystyle{IEEEtran}
\bibliography{refs}
\clearpage
\twocolumn[
\begin{@twocolumnfalse}
\begin{center}
    \fontsize{24}{29}\selectfont AutoPSO: A Meta-Framework for Automated Particle Swarm Optimization\\
    (Supplementary Document)
    \vspace{1em}
\end{center}
\end{@twocolumnfalse}
]

\begingroup

\setcounter{algorithm}{0}
\setcounter{table}{0}
\setcounter{figure}{0}
\setcounter{section}{0}
\setcounter{equation}{0}
\renewcommand\thealgorithm{S\arabic{algorithm}}
\renewcommand\thetable{S.\Roman{table}}
\renewcommand\thefigure{S\arabic{figure}}
\renewcommand\thesection{S.\Roman{section}}
\renewcommand\theequation{S.\arabic{equation}}

\providecommand{\theHalgorithm}{}
\providecommand{\theHtable}{}
\providecommand{\theHfigure}{}
\providecommand{\theHsection}{}
\providecommand{\theHsubsection}{}
\providecommand{\theHsubsubsection}{}
\providecommand{\theHequation}{}
\renewcommand{\theHalgorithm}{supp.algorithm.\arabic{algorithm}}
\renewcommand{\theHtable}{supp.table.\arabic{table}}
\renewcommand{\theHfigure}{supp.figure.\arabic{figure}}
\renewcommand{\theHsection}{supp.section.\arabic{section}}
\renewcommand{\theHsubsection}{\theHsection.\arabic{subsection}}
\renewcommand{\theHsubsubsection}{\theHsubsection.\arabic{subsubsection}}
\renewcommand{\theHequation}{supp.equation.\arabic{equation}}
\section{Introduction}
This document serves as the supplementary material for the paper titled \emph{AutoPSO: A Meta-framework for Automated Particle Swarm Optimization}. It is organized as follows: Section II details the downloadable resources, including the testing platforms, software libraries, and test problems utilized in our experiments. Section III details the benchmarks employed. Section IV reports the complete results of the comparative studies, providing visualizations and statistical summaries that illustrate the performance of tested algorithms.

\section{Downloadable Material}
All experiments in this work utilize the \emph{EvoX} platform \cite{Huang2024}. This comprehensive framework is tailored to evolutionary computation research and streamlines the execution and management of algorithms across diverse hardware configurations. Our evaluation spans both single-objective and multi-objective settings, covering two primary families of tasks: numerical optimization and neuroevolution. Specifically, the numerical optimization problems are drawn largely from the CEC 2022~\cite{Abdelatti2021} benchmark suites, while the neuroevolution tasks employ environments based on the Brax engine. These benchmark functions are widely used in the literature to provide standardized, diverse, and progressively challenging testbeds for optimization algorithms. Detailed descriptions and datasets are available via the links provided below.

\begin{itemize}
     \item \textbf{EvoX}: \url{https://github.com/EMI-Group/evox}

\item \textbf{CEC2022 Benchmark}: \url{https://github.com/P-N-Suganthan/2022-SO-BO}

\item \textbf{Robotic control tasks based on Brax engine}: \url{https://github.com/google/brax}

\textbf{MetaBox}: \url{https://github.com/MetaEvo/MetaBox/tree/v2.0.0}
\end{itemize}

\section{Benchmarks}
\subsection{Numerical Optimization Functions}
Table~\ref{supp:tab:benchmark} summarizes the numerical benchmarks utilized in this study. The primary benchmark set is the CEC2022 suite, which comprises 12 functions divided into three categories: basic ($F_{1}$--$F_{5}$), hybrid ($F_{6}$--$F_{8}$), and composition ($F_{9}$--$F_{12}$). These functions collectively cover fundamental optimization characteristics, including multimodality, separability, ill-conditioning, and non-convexity, providing a comprehensive assessment of algorithm performance.

In addition to CEC2022, we incorporate several canonical test functions—Ackley, Griewank, Rastrigin, Rosenbrock, Schwefel, and Sphere—commonly employed in optimization research. These functions enable comparison with prior studies and facilitate evaluation across both simple unimodal landscapes (e.g., Sphere) and highly multimodal or deceptive landscapes (e.g., Rastrigin and Schwefel).

\suppTabBenchmark

\subsection{Robot Control Task based on Brax Engine}

Brax provides a suite of physics-based robotic control environments that are widely used in reinforcement learning (RL) and evolutionary algorithm research. These environments simulate articulated agents performing continuous control tasks in 3D, allowing rigorous evaluation of policy optimization algorithms.

Figure~\ref{supp:fig:brax-environments} presents representative Brax environments. Each environment leverages Brax’s differentiable physics engine, implemented in JAX, enabling highly parallelized simulation on GPUs and TPUs. This accelerates gradient-based optimization and supports large-scale experimentation. Consequently, Brax environments are compatible with various RL and evolutionary algorithms, including Proximal Policy Optimization (PPO), Augmented Random Search (ARS), and Evolutionary Strategies (ES), making them versatile benchmarks for both policy-gradient and population-based methods.

\suppFigBrax

\section{Experimental Setting}
Key hyperparameters for the baseline algorithms were set following the recommendations in their respective original papers. Table~\ref{supp:tab:params} summarizes the configurations employed in our experiments. All algorithms share a population size of 100 unless otherwise specified, ensuring fair comparison across optimization strategies.

\suppTabParams

\section{Exemplar Pool in AutoPSO}
Table~\ref{supp:tab:exemplar} lists the nine exemplar candidates used in AutoPSO. Exemplar assignment strategies are categorized into three types: self-based, population-based, and randomized. The table also shows the corresponding code references, facilitating reproducibility and clarity in implementation.

\suppTabExemplar

\section{Overall Optimization Performance}
\subsection{Performance on CEC2022 Benchmark Suite}

\subsubsection{Comparison under Equal Time}

The experiments compare the proposed AutoPSO against six classical PSO variants using the CEC2022 benchmark suite. The complete numerical results are summarized in Table~\ref{supp:tab:CEC2022}. Furthermore, Figs.~\ref{supp:fig:cec022-10d} and~\ref{supp:fig:cec022-20d} illustrate the corresponding convergence curves for 10- and 20-dimensional problems, respectively. To ensure statistical reliability, all results are averaged over 31 independent runs.

\suppFigCECTenD

\suppFigCECTewD

\suppTabCEC

\subsubsection{Comparison under Equal FEs}

We compare the performance of several algorithms, including AutoPSO, PSO, CSO, CLPSO, FIPS, SLPSOGS, SLPSOUS, iwPSO, and TVAC-PSO, under the same number of FEs. All experiments were conducted on the CEC2022 benchmark functions with dimensions 10D and 20D, using a fixed total of 248,000,000 FEs for each algorithm. Each configuration was repeated 11 independent runs to ensure statistical reliability. The detailed results of this comparison are presented in Table~\ref{supp:tab:CEC-FE}.

\suppTabCECFE

\subsection{Performance on Large-scale and Shifted and Rotated Benchmarks}
To evaluate the scalability and robustness of AutoPSO, we conducted extensive experiments on classical and modified numerical optimization functions across a wide range of dimensionalities. 
For each numerical function, we evaluated performance in multiple dimensions to assess scalability. Specifically:
\begin{itemize}
\item \textbf{Classical functions:} Ackley, Griewank, Rastrigin, Rosenbrock, Schwefel, and Sphere. Each function was tested in 50D, 100D, and 200D. The maximum runtime per run was set to 30 seconds, and results were averaged over 31 independent runs to ensure statistical reliability.
\item \textbf{Shifted and rotated functions:} To investigate robustness to transformations, we additionally evaluated shifted and rotated versions of Ackley and Rastrigin, with 50D, 100D, and 200D dimensions over 31 runs. Each run was allowed up to 300 seconds. 
\item \textbf{Very high-dimensional functions:} Further experiments were conducted on extremely high-dimensional Ackley and Rastrigin functions (500D, 1000D, and 2000D) to test scalability, with a runtime limit of 300 seconds per run. Each function run for 31 times.
\end{itemize}

\subsubsection{Results on Classical Benchmarks}

Table~\ref{supp:tab:basic} summarizes the experimental results on the six classical numerical functions. AutoPSO consistently achieved the best or comparable performance across all dimensions. Figs.~\ref{supp:fig:ackley}–\ref{supp:fig:sphere} show the convergence curves for each function in 50D, 100D, and 200D. 
The experimental results on six classical numerical optimization functions are summarized in Table~\ref{supp:tab:basic}. To complement the CEC2022 benchmarks and provide a comprehensive assessment, evaluations for each function were conducted across three dimensions: 50, 100, and 200. Figs.~\ref{supp:fig:ackley} to~\ref{supp:fig:sphere} illustrate the corresponding convergence curves. All reported results represent the averages over 31 independent runs, thereby ensuring statistical reliability.
\suppFigAckleyAll

\suppFigAckleyAllT

\suppTabBasic

\subsubsection{Performance on Shifted and Rotated Functions}

To evaluate robustness under coordinate transformations, we tested shifted and rotated Ackley and Rastrigin functions. Figs.~\ref{supp:fig:ackley-sr} and~\ref{supp:fig:rastrigin-sr} illustrate the convergence curves across 50D, 100D, and 200D. AutoPSO maintained superior performance in terms of convergence speed and solution quality, demonstrating its invariance to shifts and rotations in the search space.

\suppFigASR
\suppFigRSR

\subsubsection{Performance on Large-dimensional Benchmarks}

To test scalability, we extended the experiments to very high-dimensional Ackley and Rastrigin functions (500D, 1000D, and 2000D). Figs.~\ref{supp:fig:Ackley-high} and~\ref{supp:fig:Rastrigin-high} show convergence curves. AutoPSO consistently outperformed baseline algorithms in both convergence rate and final solution quality, confirming its effectiveness in large-scale optimization scenarios.

\suppFigAH

\suppFigRH

\subsection{Performance on Neuroevolution Robot Control Tasks}
We evaluate the tested algorithms on a set of robot control tasks implemented in the Brax engine.
To assess the capability of AutoPSO, we selected five representative environments: HalfCheetah, Hopper, Pusher, Reacher, and Swimmer.
The overall results are summarized in Table~\ref{supp:tab:brax}, and the corresponding convergence curves are illustrated in Fig.~\ref{supp:fig:brax-rewards}.
To ensure statistical reliability, all reported values are averaged over 11 independent runs.

\suppFigBraxRe

\suppTabBraxRe

\subsection{Comparison with Other EAs}
To further assess the generality and robustness of AutoPSO, we compare it with representative algorithms from three additional EA families: DE, GA, and CMA-ES, along with the canonical PSO baseline. These comparisons are conducted under the identical experimental settings described previously. As shown in Fig.~\ref{supp:fig:cross}, AutoPSO consistently delivers superior performance across all cases. 
In the Reacher and Swimmer environments, it converges substantially faster and attains higher rewards than competing methods, highlighting its ability to adaptively regulate swarm dynamics for efficient policy search.
Furthermore, on the 20-dimensional CEC2022 $F_2$, AutoPSO reduces error by several orders of magnitude, whereas other algorithms quickly stagnate.
While DE typically excels on classical CEC benchmarks and CMA-ES is renowned for its robust covariance adaptation in continuous optimization, both approaches rely on manual hyperparameter tuning and remain sensitive to task-specific conditions.
By contrast, AutoPSO unifies and extends the strengths of various PSO variants via meta-level self-tuning, thereby enabling automatic adjustment to heterogeneous problem landscapes.
This cohesive design allows AutoPSO not only to overcome the intra-family limitations of conventional PSO but also to surpass the performance boundaries observed across distinct EA families.

\suppFigCross

\subsection{Generalization Ability of Discovered PSO Variants}
The variants discovered on the CEC2022-20D benchmark functions $F_6$, $F_8$, $F_{10}$, and $F_{12}$ are applied to the remaining functions in the same benchmark to evaluate their generalization performance. 
Each value represents the mean performance over 11 independent runs, and the minimum mean in each row is highlighted in gray and bold for clarity.

\suppTabGen

\section{Mechanism Analysis and Ablation Study}

\subsubsection{Effect of Algorithm Components}

conduct ablation studies on CEC2022-20D 120s 31 runs. The tested variants include removing the multi-swarm partition mechanism (AutoPSO-m), disabling adaptive parameter control by fixing $w$, $c_1$, and $c_2$ to 0.6, 2.5, and 0.8 (AutoPSO-c), and restricting exemplar selection to personal-best and global-best solutions only (AutoPSO-e). We further introduce finer-grained variants to isolate specific elements, including removing acceleration coefficients (w/o acc.), removing inertia weight (w/o weight), and excluding exemplar categories (w/o random, w/o self, and w/o social). The results are summarized in Table~\ref{supp:tab:component}.

\suppTabComponent

\subsubsection{Impact of the Number of Sub-swarms}
We examine the design of using two sub-swarms in AutoPSO by comparing several partition strategies on CEC2022-20D, including no partition, a fixed-ratio bipartition (0.5/0.5), and configurations with three or four sub-swarms. The results were summarized in Table~\ref{supp:tab:subswarm}.

\suppTabSub

\subsection{Individual-Level Heterogeneity Analysis}
In AutoPSO, parameter configurations and exemplar update rules are assigned at the sub-swarm level rather than at the individual particle level. This design introduces behavioral diversity across sub-swarms while maintaining coherent swarm dynamics within each group. Moreover, using a unified configuration within each sub-swarm reduces the search space of the outer-level optimizer and improves computational efficiency for large-scale parallel evaluation.

To investigate whether particle-level heterogeneity could further benefit AutoPSO, we design three variants incorporating individual-level mechanisms: (i) {IndParam}: introduces particle-level parameter perturbation. For each particle, the parameters $w$, $c_1$, and $c_2$ are slightly perturbed around the sub-swarm configuration discovered by AutoPSO, resulting in heterogeneous parameters across particles.
(ii) {IndExemplar}: introduces particle-level exemplar selection. Each particle independently samples learning targets from the two exemplar sources used in AutoPSO, introducing individual-level diversity in exemplar selection while keeping sub-swarm parameters shared.
(iii) {IndRandom}: each particle independently selects a complete update strategy (parameters and exemplar sources) from the configuration pool, effectively removing sub-swarm structure and introducing maximal particle-level heterogeneity.
All variants are evaluated on the 20D CEC2022 benchmark within 120s over 11 runs. Results are summarized in Table~\ref{supp:tab:ind_analysis}.

\suppTabInd

The experimental results reveal several important insights. AutoPSO demonstrates that sub-swarm-level configurations, combined with stochasticity and exemplar sampling, already generate sufficient heterogeneity, leading to consistently strong performance across most benchmark functions. Introducing particle-level exemplar selection through {IndExemplar} provides modest improvements on certain complex functions, suggesting that allowing individual particles to sample independently from exemplar sources can enhance exploration in challenging landscapes. In contrast, {IndParam}, which perturbs parameters at the particle level, shows limited additional benefit, indicating that small parameter variations alone do not substantially increase effective behavioral diversity. Notably, {IndRandom} exhibits markedly poorer performance. In this variant, each particle independently selects a full update strategy from the configuration pool, which breaks the cohesion within sub-swarms and disrupts the collaborative information flow that underlies effective swarm dynamics. As a result, particle trajectories become highly inconsistent, impairing convergence and leading to larger variance in fitness outcomes.

These results support the design choice of sub-swarm-level configuration in AutoPSO as a practical trade-off: it maintains heterogeneous dynamics while keeping the outer-level search space tractable and the meta-optimization process computationally feasible for large-scale problems. Extending AutoPSO to fully individual-level adaptive configurations remains a promising direction for future work.

\subsection{Comparison with Adaptive Parameter-Control Methods} 
In the inner solver of AutoPSO, the PSO variants use fixed parameter values discovered by the outer-level meta-search. To examine whether this design reduces algorithmic flexibility, we conducted experiments on the CEC2022 20D benchmark suite, using a 120-second evaluation budget per run and 11 independent trials per function. We compared AutoPSO with two representative adaptive PSO variants: iw-PSO, which linearly decreases the inertia weight over iterations, and TVAC-PSO, which gradually decreases the cognitive coefficient while increasing the social coefficient. Each inner-level PSO variant corresponds to the best configuration discovered by AutoPSO for that specific problem, including the optimized inertia weight, acceleration coefficients, exemplar assignments, and population split ratio. 

The results are summarized in Table~\ref{supp:tab:adaptive}. AutoPSO outperforms both iw-PSO and TVAC-PSO across most benchmark functions, with notable advantages on challenging functions such as $F_2$ and $F_6$. This performance stems from the ability of AutoPSO to discover problem-specific parameter regimes that align with the landscape characteristics, rather than relying on generic, hand-designed schedules shared across problems. 
Furthermore, while adaptive schemes provide short-term responsiveness, frequent adjustments can sometimes introduce oscillations, over-contraction, or instability in complex or high-dimensional landscapes. In contrast, the fixed parameters identified by AutoPSO maintain consistent step sizes and social influence throughout the search, which can help preserve stable swarm dynamics.

\suppTabAda

\subsection{Choice of Outer Optimizer}
We instantiated four widely used population-based methods as outer optimizers: CMA-ES, DE, GA, and PSO.  
Each variant was evaluated on five robotic control benchmarks, with 11 independent runs per task.  
Fig.~\ref{supp:fig:outer} illustrates the mean reward convergence curves, and the final average returns across runs are summarized in Table~\ref{supp:tab:outer}.

\suppFigOuter
\suppTabOuter

\subsection{Effect of Population Size for EAs}
Since the proposed AutoPSO algorithm exhibits improved performance with larger swarm sizes, we further evaluate the baseline methods under varying population scales. Fig.~\ref{supp:fig:base-pop} presents the results of three representative algorithms using population sizes of 100, 1000, 5000, and 10000. The results indicate that not all algorithms consistently benefit from increasing the number of particles. This scalability characteristic makes AutoPSO particularly well suited for parallel computing environments, as it can effectively leverage GPU hardware to achieve substantial performance gains in practical applications.

\suppFigBasePop

\endgroup

\end{document}